\documentclass[10pt,twocolumn,letterpaper]{article}

\usepackage[pagenumbers]{cvpr} 

\usepackage[dvipsnames]{xcolor}
\newcommand{\red}{\color{red}}
\usepackage[ruled]{algorithm2e} 
\usepackage{booktabs} 
\usepackage{abstract}
\usepackage[tracking=true]{microtype}
\usepackage{titling}
\usepackage{enumitem}
\usepackage{amsfonts}
\usepackage{amsmath}
\usepackage{amssymb}
\usepackage{bm}
\usepackage{tikz}
\usepackage{pifont}
\usepackage{listings}
\usepackage{bbding} 
\usepackage{makecell}
\usepackage{multirow}
\usepackage{multicol}
\definecolor{cvprblue}{rgb}{0.21,0.49,0.74}
\definecolor{darkgreen}{rgb}{0,0.5,0}
\newcommand{\cmark}{\textcolor{darkgreen}{\ding{51}}}
\newcommand{\xmark}{\textcolor{red}{\ding{55}}}%

\newcommand{\btheta}{\bm{\theta}}

\definecolor{cvprblue}{rgb}{0.21,0.49,0.74}
\usepackage[pagebackref,breaklinks,colorlinks,citecolor=cvprblue]{hyperref}

\def\paperID{85} 
\def\confName{3DV\xspace}
\def\confYear{2025\xspace}

\newcommand{\shenlongdone}[1]{}

\newcommand{\green}{\color{darkgreen}}

\begin{document}
\twocolumn[{%
\renewcommand\twocolumn[1][]{#1}%

\title{Plenoptic PNG: Real-Time Neural Radiance Fields in 150 KB}


\author{Jae Yong Lee \thanks{Currently at Apple} \hspace{2pt}$^1$\\
\and
Yuqun Wu$^1$\\
\and
Chuhang Zou$^2$\\
\and
Derek Hoiem$^1$ \\
\and
Shenlong Wang$^1$
\and
$^1${\small University of Illinois at Urbana-Champaign}\\
{\tt\small \{lee896, yuqunwu2, dhoiem, shenlong\}@illinois.edu}
\and
$^2${\small Amazon Inc.}\\
{\tt\small zouchuha@amazon.com}
}

\date{}

\maketitle
\begin{center}
    \vspace{-1em}
    \centering
    \scriptsize
    \captionsetup{type=figure}
    \includegraphics[width=0.9\textwidth]{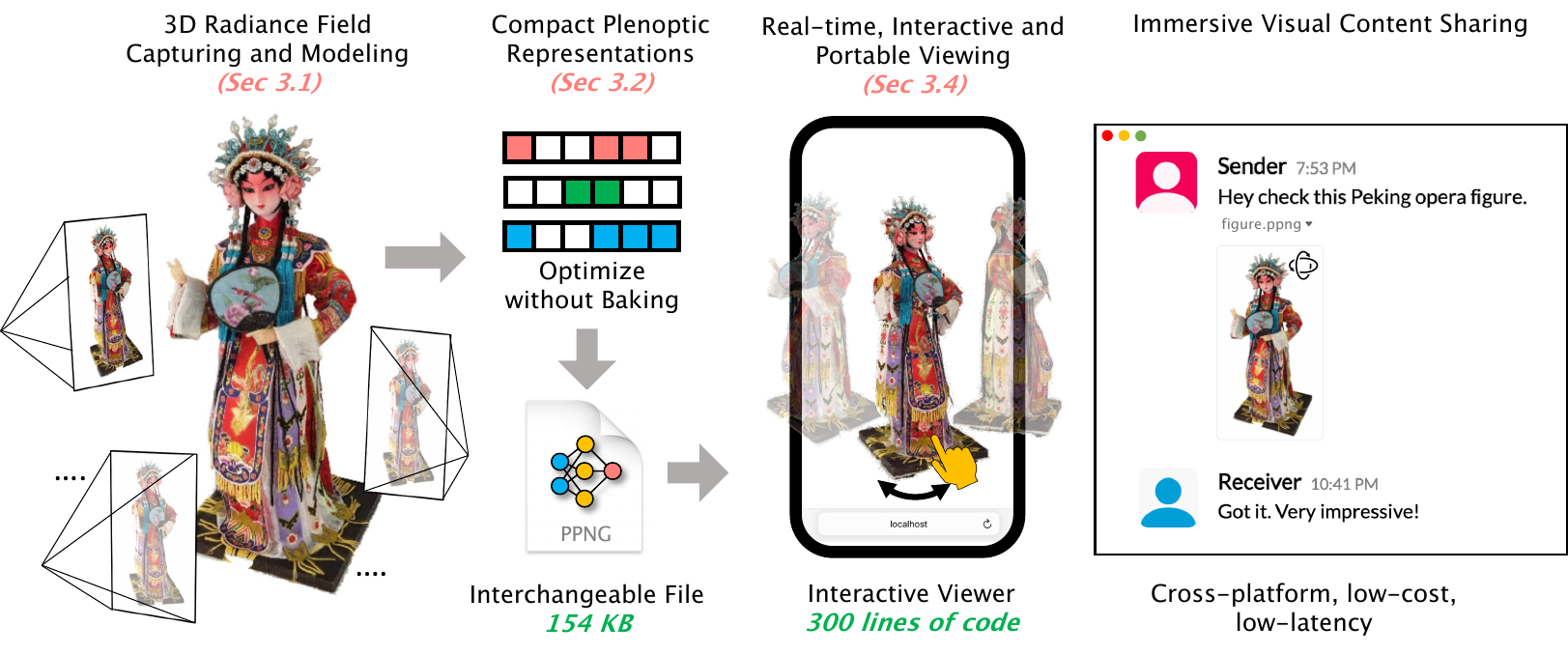}
    \vspace{-1em}
    \caption{Given multi-view images as input, Plenoptic PNG (PPNG) encodes the parameters of the free-viewpoint radiance field into a compact, interchangeable file as small as 154 KB in just a few minutes. On the user side, the PPNG file can be decoded into a WebGL-compatible shader and 3D texture within 100 ms, and renders on lightweight devices in real-time. PPNG opens up potential new applications, allowing for easier capture and sharing of photo-realistic visuals across platforms.}
\label{fig:teaser}

\end{center}%
}]
\saythanks
\begin{abstract}
The goal of this paper is to encode a 3D scene into an {\it extremely compact} representation from 2D images and to enable its transmittance, decoding and rendering in real-time across various platforms.
Despite the progress in NeRFs and Gaussian Splats, their large model size and specialized renderers make it challenging to distribute free-viewpoint 3D content as easily as images. 
To address this, we have designed a novel 3D representation that encodes the plenoptic function into sinusoidal function indexed dense volumes. This approach facilitates feature sharing across different locations, improving compactness over traditional spatial voxels. 
The memory footprint of the dense 3D feature grid can be further reduced using spatial decomposition techniques. This design combines the strengths of spatial hashing functions and voxel decomposition, resulting in a model size as small as 150 KB for each 3D scene. 
Moreover, PPNG features a lightweight rendering pipeline with only 300 lines of code that decodes its representation into standard GL textures and fragment shaders. This enables real-time rendering using the traditional GL pipeline, ensuring universal compatibility and efficiency across various platforms without additional dependencies.
Our results are available at: \url{https://jyl.kr/ppng}
\end{abstract}

\section{Introduction}

Capturing and viewing immersive content has become easier than ever. 
Recent progress in approaches like Neural Radiance Fields (NeRF) and Gaussian Splatting have enabled users to capture 3D content from mobile devices. The commercialization of XR devices, such as the Apple Vision Pro and Oculus, has enhanced the viewing experiences of photorealistic immersive visual content. However, challenges persist in efficiently storing, transmitting, and browsing this content across various devices. 
In this work, we pursue a novel approach to model and encode free-viewpoint 3D content into a file as small as a PNG photo, making viewing and interacting with it as easy as browsing videos on various devices like mobile phones, laptops, or AR/VR glasses.

There are three key desiderata to democratize immersive, photorealistic 3D content. First, the model size must be small to avoid degrading user experience in instant messaging and web browsing. Second, viewing and interaction should be universal, not relying on specialized dependencies or hardware. 
Third, rendering and interaction must be smooth and real-time. Despite significant progress in this field, current approaches fail to meet all these criteria. For instance, NeRFs and their variants may be compact, but many are not real-time and depend on specialized packages such as CUDA-based neural volume renderers. 
Explicit methods like Gaussian Splats and NeRF baking are fast and versatile, but explicit geometry requires substantial storage. 
Table~\ref{tab:related_work} summarizes these methods and their alignment with the desired criteria. 

To achieve this, we propose Plenoptic Portable Neural Graphics, a novel framework that is highly compact and fast to render and train, providing a free-view portable network graphics object.
At the core of our framework is a novel, compact scene representation and a real-time, cross-platform GL-compatible render. 
Unlike coordinate-based MLPs~\cite{Mildenhall2020NeRFRS, Barron2021MipNeRFAM, Barron2023ZipNeRFAG} or spatial voxel grids~\cite{Yu2021PlenoxelsRF, Chen2022TensoRFTR, Mller2022InstantNG}, our method leverages an explicit 3D voxel feature grid 
indexed by the sinusoidal encoding of the spatial coordinate. 
This new spectrally indexed volume enables feature sharing across different spatial locations, improving compactness over spatial voxels. The model size of this dense 3D voxel feature grid can be further reduced through tensor-rank decomposition. Consequently, this representation design combines the best aspects of spatial hashing function~\cite{Mller2022InstantNG} and voxel decomposition~\cite{Chen2022TensoRFTR} approaches, resulting in a size as small as 154 KB. Additionally, we develop a novel, lightweight, and real-time rendering pipeline that can decode Plenoptic PNG representation instantly into standard GL textures and shaders, and render with OpenGL pipeline, making it universally viewable on any platform without additional dependencies. 

Our experiments demonstrate that Plenoptic PNG surpasses baselines with a significantly reduced model size, as small as 154 KB — 100 times smaller than previous memory-efficient methods. We also show that Plenoptic PNG achieves the best balance between training speed, rendering quality, and model size among all real-time Web-ready NeRF methods, producing a widely accessible, realistic, and efficient interchangeable file format for immersive 3D media. We invite the reader to view our project page in the supplementary file, where we render 8 neural scenes simultaneously in real-time on a webpage at 1.2 MB in total. Our key contributions are:
\begin{itemize}[leftmargin=*]
\item We present a novel neural scene model that encodes multiple views into an extremely compact tensor representation indexed by Fourier encoding, showing significant model size reduction compared to prior work.
\item We develop a lightweight rendering pipeline that can instantly decode the Plenoptic PNG representation into standard GL textures and shaders, and render it in real-time in WebGL, making it viewable and interactable on any platform.
\vspace{-2mm}
\end{itemize}

\section{Related works}
\begin{table}[t]
\caption{{\bf Comparison of various NeRF methods.} Previous implicit neural representations tend to have a smaller memory footprint but suffer from relatively lower speed and are incompatible with web renderers. Explicit approaches enjoy real-time speed and GL compatibility but require a large model size. Our method is the first of its kind to achieve a kilobyte-level model size and satisfies all the criteria.
}
\vspace{-3mm}
\resizebox{\linewidth}{!}{
\begin{tabular}{l|c|c|c|c}
\toprule
Method       & 
\makecell{Real-time \\ ({\green FPS: $>$ 30 Hz})} & 
\makecell{Web-Ready \\ ({\green GL native})} & 
\makecell {Memory-Efficient \\ ({\green Model Size: $<$ 5 MB})} & 
\makecell{Fast Training \\ ({\green{Time:} $<$ 15 min})} \\ 
\midrule
NeRF        & \red      \xmark  & \xmark  & \green{5 MB}    & \red      hours \\ 
Plenoxel    & \green    \cmark  & \cmark  & \red 778 MB     & \green    11.4 min \\ 
Plenoctree    & \green    \cmark  & \cmark  & \red 1976 MB     & \red    hours \\
DIVeR    & \green    \cmark  & \xmark  & \red 67.8 MB     & \red    hours \\
SNeRG       & \green    \cmark  & \cmark  & \red 86.8 MB    & \red      hours\\
TensoRF(CP) & \red      \xmark  & \xmark  & \green 3.9 MB   & \red 25.2 min\\ 
Wavelet-NeRF (4-DWT) & \red     \xmark  & \cmark  & \green 710 KB     & \red 23 min  \\
Instant NGP & \green      \cmark  & \xmark  & \red 32 MB      & \green    5 min \\
Compact-NGP & \green    \cmark  & \cmark  & \green 357 KB     & \green     12 min \\
VQRF        & \red      \xmark  & \xmark  & \green 1.4 MB   & \green    8 min \\
MobileNeRF  & \green    \cmark  & \cmark  & \red 125.8 MB   & \red      hours \\ 
BakedSDF    & \green    \cmark  & \cmark  & \red 382 MB     & \red      hours \\
MERF        & \green    \cmark  & \cmark  & \red 120 MB     & \red      hours \\
Re-Rend     & \green    \cmark  & \cmark  & \red 199 MB     & \red      hours \\
Gauss. Spl. & \green    \cmark  & \cmark  & \red  67.3 MB   & \green    7.6 min \\
\midrule
PPNG-3 (Ours)      & \green    \cmark  & \cmark  & \red 32.7 MB      & \green    5 min \\ 
PPNG-2 (Ours)      & \green    \cmark  & \cmark  & \green 2.5MB    & \green    10 min \\ 
PPNG-1 (Ours)      & \green    \cmark  & \cmark  & \green 151 KB   & \green    13 min \\ 
\bottomrule
\end{tabular}
}
\label{tab:related_work}
\end{table}

Our goal is to encode multiple 2D images of a  3D scene into an extremely compact representation that can be rendered from custom viewpoints in real-time across various platforms. Our method relates most closely to real-time neural radiance field methods, and we draw inspiration from 3D and neural compression. 

\begin{figure*}[t]
    \centering
    \includegraphics[width=0.99\textwidth]{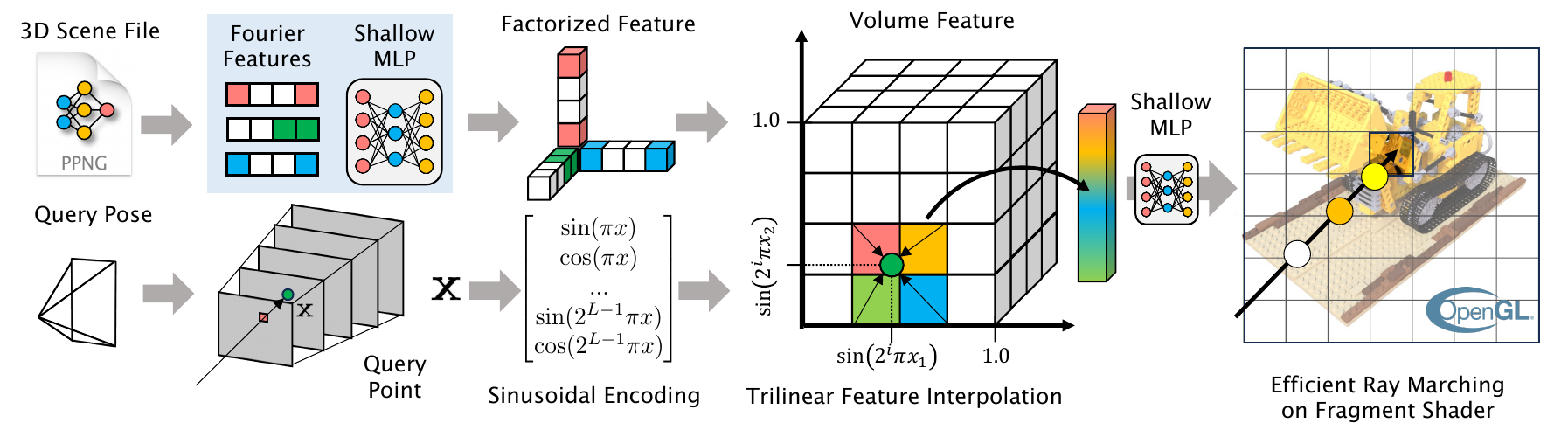}
    \vspace{-5mm}
    \caption{{\bf Overview of our PPNG-1 Rendering Procedure:} For a given PPNG file of a 3D scene, we first extract the factorized Fourier features and the shallow MLP weights (top-left). The factorized Fourier features are then composed to construct a dense Fourier-indexed feature grid (middle). In the rendering stage, for each query point we compute the sinusoidal positional encoding to extract the corresponding feature from the Fourier-indexed voxel grid. The feature vectors, spanning across the spectrum for both sine and cosine at each frequency, are then concatenated. These features are subsequently passed into the fragment shader, which employs a shallow MLP for inferring color and density and applies ray matching to determine the final pixel color.}
    \label{fig:rendering}
        \vspace{-3mm}
\end{figure*}
\noindent\textbf{Real-time Neural Radiance Field (NeRF)}.~
NeRF~\cite{Mildenhall2020NeRFRS} has emerged as one of the most promising and widely adopted novel view synthesis methods. NeRF represents the 3D scene with coordinate-based multi-layer perceptrons (MLPs) and achieves high-quality rendering through volume rendering. Despite its compactness, the original NeRF suffers from slow training and rendering. Feature volume-based approaches~\cite{Yu2021PlenoxelsRF, Chen2022TensoRFTR, Mller2022InstantNG, Kerbl20233DGS} encode the scene with a dense feature grid, which leads to faster rendering and training. 
Rendering speed can be further accelerated using sparse volumetric data structures~\cite{Liu2020NeuralSV, Sun2021DirectVG, Garbin2021FastNeRFHN, Yu2021PlenOctreesFR, Wu2021DIVeRRA, Material2023PlenVDBME}. Additionally, methods like those in ~\cite{Kurz2022AdaNeRFAS, Wang2023AdaptiveSF} optimize the volumetric sampling process to increase rendering speed during inference. However, most neural volume rendering methods still require a high-capacity GPU and specialized volume renderer, which limits their applicability.

An appealing alternative is jointly learning explicit geometry, such as points and meshes, and baking appearance features like opacity and view-dependent color onto the geometric surfaces~\cite{Hedman2021BakingNR, Chen2022MobileNeRFET, Reiser2023MERFMR, Yariv2023BakedSDFMN, Rojas2023ReReNDRR}. Such approaches often align with real-time graphics pipelines like OpenGL, making them accessible across various devices. However, most explicit geometry-based neural rendering methods suffer from memory inefficiency and slow training speeds. Very recently, Gaussian splatting~\cite{Kerbl20233DGS} has emerged as an exception, striking the best balance between speed, quality, and training speed. Nevertheless, the file sizes for Gaussian Splatting scenes still range from tens to hundreds of MBs.




\noindent\textbf{Volume Compression}.~
Compressing volumetric 3D data has long been a challenge in graphics. The key lies in designing both a compact 3D representation and an expressive encoding. Numerous approaches explore efficient data structures like octrees coupled with entropy coding~\cite{Huang2008AGS, Garcia2018IntraFrameCO, schnabel2006octree, Huang2020OctSqueezeOE} to reduce redundancies in 3D data. Block-based coding schemes, commonly used in volumetric compression~\cite{Balsa2014StateoftheArtIC, Crassin2009GigaVoxelsRS}, can be further optimized through data filtering~\cite{Heitz2015TheSM, Heitz2012RepresentingAA}. Recent works also leverage various tools, such as the Karhunen-Loève transform~\cite{Tang2018RealtimeCA}, auto-encoders~\cite{wang2019learned, Tang2020DeepIV, biswas2020muscle}, and wavelet transform~\cite{de2016compression}, to compress nodes in tree structures. While these compressed explicit representations suit traditional graphics, they fall short in rendering photorealistic images from free viewpoints. 
With Gaussian Splatting~\cite{Kerbl20233DGS} emerging as an alternative to photorealistic representation, some of the most recent works explore anchor / hash-grid based compression~\cite{chen2024hac} and distributing smaller number of samples ~\cite{fang2024mini}. These Gaussian Splatting variants show impressive compression ratio, yet still use an order of magnitude larger model size compared to the implicit models. 

\noindent\textbf{Neural Field Compression}.~ 
As implicit representations~\cite{Park2019DeepSDFLC, Mildenhall2020NeRFRS} show promise in graphics, many works aim to reduce memory footprint while maintaining high accuracy~\cite{Bird20213DSC, Lu2021CompressiveNR, Gordon2022OnQI}. Inspired by pioneering light field work~\cite{Levoy1996LightFR}, real-time light-field compression approaches distills a compact representation from NeRF and achieved reduced memory footage and real-time rendering on mobile devices~\cite{Cao2022RealTimeNL, Gupta2023lightspeed}. However, this process incurs high training costs and cannot yet achieve KB-level compression. 
Drawing on its widespread use in graphics, NGLoD~\cite{Takikawa2021NeuralGL} learns a sparse octree with continuous levels of detail (LOD). Various approaches using learning and handcrafted codebooks have also been proposed to compress neural fields, such as wavelets~\cite{Rho2022MaskedWR}, vector-quantized feature codebooks~\cite{Lu2021CompressiveNR, Gordon2022OnQI, Li2022CompressingVR}, learning-based feature indexing mapping~\cite{Takikawa2022VariableBN, Takikawa2023CompactNG}, multi-scale look-up tables~\cite{Datta2023EfficientGR}, adaptable rank~\cite{yuan2023slimmerf} binarization~\cite{shin2023binary}, and neural image compression~\cite{li2024nerfcodec}.
While showing promising results, they often require additional memory for storing codebooks, use extra decoding time or takes additional time to optimize / render. 
Moreover, despite their small size, such compressed neural field representations usually need specialized CUDA-dependent renderers and decoders~\cite{Takikawa2023CompactNG,Mller2022InstantNG} or decodes into large size~\cite{Rho2022MaskedWR,Chen2022TensoRFTR} (i.e, when fully composed to dense voxel grid of size~$O(500^3)$ to avoid feature composition for faster rendering) which makes it hard to run on light-weight devices such as mobile phones. 
Instead, our representation can render in generic graphics library such as WebGL, and decodes into reasonably small size of 32.7 MB per scene.
Our scene representation can be seen as a special instance of recent Dictionary Fields~\cite{Chen2023DictionaryFL}, which unify various representations, including vanilla NeRF, NGP, and TensoRF, by encompassing permutations of coordinate, basis/coefficient, and activation representation. 
Unlike Dictionary Fields' broad unification, our work focuses on solving real-time rendering with minimal data transfer. We achieve this with a dual representation design choice: 1) encoding NeRF into a compact 1D/2D factorized tensor for efficient data transfer, and 2) decoding it into a GL-compatible 3D feature grid for real-time web rendering.


To summarize, Table~\ref{tab:related_work} presents the capabilities of current NeRF-based approaches in terms of model size, training speed, rendering speed, and web compatibility. While each has its strengths, our approach meets all requirements for wide application in daily use.

\section{Plenoptic Portable Neural Graphics}
The goal of PPNG is to encode a set of images with known poses into a small model that can be efficiently transmitted and rendered on ubiquitous devices.
To achieve this, we first present a novel, compact neural representation that uses a sinusoidal function encoded feature volume (Section~\ref{ssec:ppng}). 
Unlike spatial coordinate indexed volumes, this approach allows for feature sharing by design, thus requiring fewer parameters to achieve the same capacity. 
We demonstrate that our Fourier-indexed volume can be factorized into low-rank tensor approximations to further reduce the model size (Section~\ref{ssec:ppng-12}). 
We implement a fast training scheme in \textit{tiny-cuda-nn}, utilizing a voxel-based density cache~\cite{Mller2022InstantNG} (Section~\ref{ssec:training}), and design a new GLSL-based renderer to decode the parameters for real-time rendering (Section~\ref{ssec:rendering}).
Figure~\ref{fig:rendering} shows an overview of the representation and rendering pipeline.


\begin{figure}[t]
    \centering
    \begin{tabular}{cc}
        \includegraphics[width=0.4\linewidth]{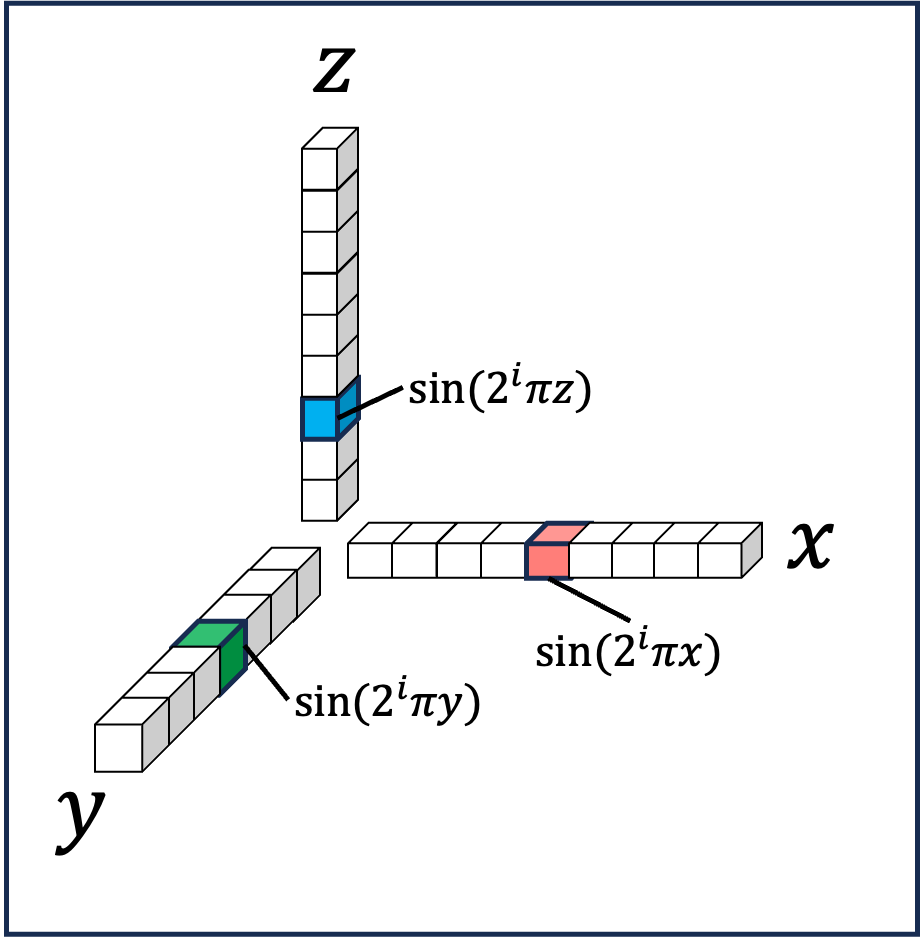} &
        \includegraphics[width=0.4\linewidth]{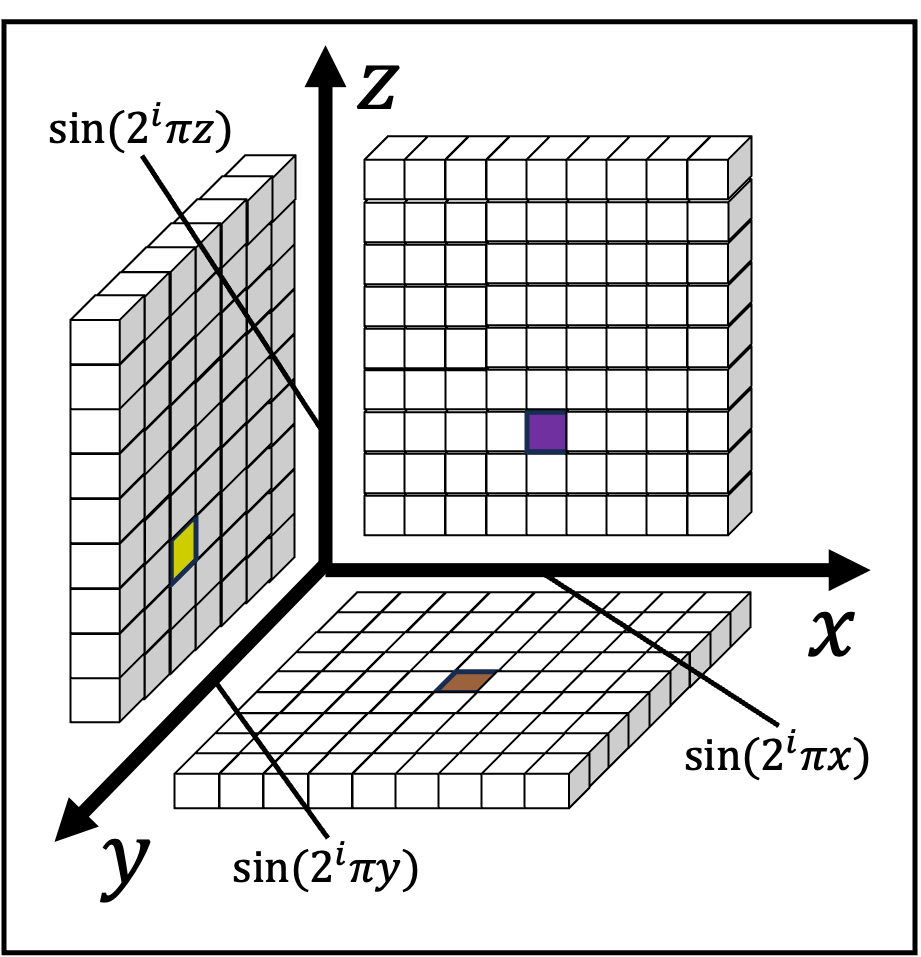} \\
         PPNG 1 & PPNG 2
    \end{tabular}
    \vspace{-3mm}
    \caption{{\bf Visualization of Two Factorized Plenoptic PNG Representations:} PPNG-1 (Equation~\ref{eq:ppng_1_to_3}) utilizes tensor-rank decomposition (left), while PPNG-2 (Equation~\ref{eq:ppng_2_to_3}) employs tri-plane decomposition (right).}
    \label{fig:ppng_1_2}
    \vspace{-3mm}
\end{figure}
\subsection{Plenoptic Portable Neural Graphics}
\label{ssec:ppng}
Our core contribution lies in a novel volumetric neural feature representation referred to as plenoptic portable neural graphics. The goal of this learnable representation is to approximate the mapping from coordinates $\mathbf{p} \in \mathbb{R}^3$ to color and opacity values $(\mathbf{c}, \sigma) \in \mathbb{R}^{3+1}$.
We develop an efficient and compact representation 
that leverages: (1) positional encoding to convert the spatial coordinate into a multi-scale, multi-dimensional Fourier embedding; and (2) volume-based feature queries to enable fast inference. 

Given the input query point $\mathbf{p}=(x, y, z)$, positional encoding~\cite{Mildenhall2020NeRFRS, tancik2020_fourier} is applied to map the Euclidean coordinates input to sinusoidal activations across $L$ different frequency levels for each axis: 

\vspace{-3mm}
\begin{equation}
\label{eq:pos_enc}
    \gamma(\mathbf{p}) = \text{concat}([\gamma(x), \gamma(y), \gamma(z)]) \in \mathbb{R}^{3\times L \times 2},
\end{equation}
 where activation for each axis is 
\begin{align*}
\label{eq:pos_enc}
 \gamma(w) &= [(\sin(f_i\pi w), \cos(f_i\pi w))]_{i=0}^{L-1} \in \mathbb{R}^{L\times 2}, \\
\end{align*}
This resulting encoding is a continuous, multi-scale, periodic representation of $\mathbf{p}$ along each coordinate.  

We maintain $L\times 2$ feature volume cubes, $\{\mathbf{V}^\mathrm{sin}_{i}, \mathbf{V}^\mathrm{cos}_{i} \in \mathbb{R}^{Q^3 \times D}\}_i^{L\times 2}$, each with a resolution of $Q^3$ and $D$-dimensional features per entry. 
These features are indexed by a 3D slice of the positional encoding $\gamma_i^\mathrm{sin} = [\sin(f_i\pi x),$ $ \sin(f_i\pi y), \sin(f_i\pi z)]$ (or cosine embedding $\gamma_i^\mathrm{cos}$) at corresponding frequency.

We query the feature vector $\mathbf{z}^\mathrm{sin}_i, \mathbf{z}^\mathrm{cos}_i$ from each volume across each frequency as: 
\begin{equation}
\label{eq:query}
\forall i:  \text{\ } \mathbf{z}^\mathrm{sin}_i =  \pi_\mathrm{tri}(\gamma_i^\mathrm{sin}, \mathbf{V}_i^\mathrm{sin}), \text{\ }
  \mathbf{z}^\mathrm{cos}_i = \pi_\mathrm{tri}(\gamma_i^\mathrm{cos}, \mathbf{V}_i^\mathrm{cos})
\end{equation}
where $\pi_\mathrm{tri}$ is tri-linear interpolation. 
We then concat all the features $\mathbf{z}(\mathbf{p}) = \mathrm{concat}[...\mathbf{z}^\mathrm{sin}_i, \mathbf{z}^\mathrm{cos}_i...]_{i=0}^L$, which result in a feature vector of length $F = 2 \times L \times D$. Finally, this feature vector is passed into a shallow MLP $g_{\btheta}$ with encoded viewing direction $\mathbf{d}$ to regress opacity and view-dependent color: 
\begin{equation}
    (\mathbf{c}, \sigma) = g_{\btheta}(\mathbf{z}(\mathbf{p}), \mathbf{d})
\end{equation}

Despite its simplicity, our design offers multiple benefits. {\it Efficiency:} Similar to volume-based neural fields, our approach is extremely efficient, enabling real-time rendering. {\it Feature sharing:} Features at each voxel location in the Fourier domain are shared and accessed simultaneously by multiple spatial locations. This design allows us to extract redundancies over space (compared to spatial volumes) while maintaining smoothness (unlike spatial hashing). {\it Compactness:} The total model size is $L \times 2 \times Q^3 \times D + |\btheta|$, where $|\btheta|$ represents the number of parameters for the shallow MLP. In practice, given the redundancies and the multi-scale nature of each volume, the size of each volume $Q$ can be significantly smaller than the typical size for spatial volumes $S$ (e.g. 80 vs 512), resulting in a smaller overall model size.

\begin{figure}[t]
\vspace{-3mm}
    \centering
    \includegraphics[width=.95\linewidth]{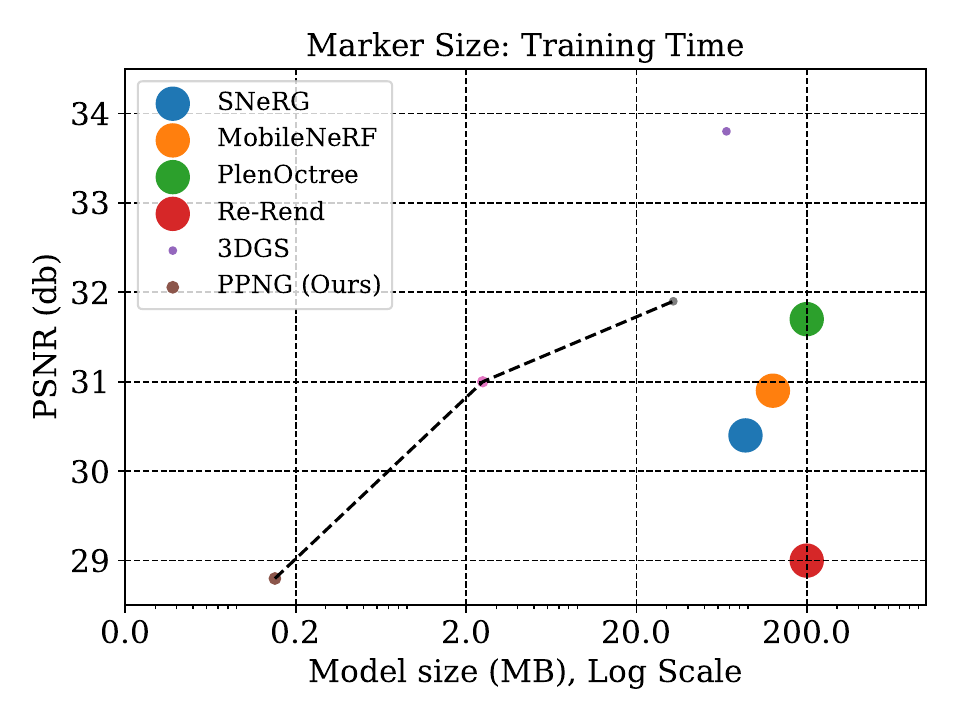}
    \vspace{-5mm}
    \caption{{\bf Quantitative Comparison with Real-Time, Web-Compatible NeRF Models on NeRF Synthetic dataset}. Our approaches are {\bf 2-3 orders of magnitude} smaller than baselines in terms of model size (x-axis) and over {\bf 10x-100x} faster in training speed (marker size), while maintaining competitive PSNR (y-axis).}
\label{fig:quantitative}
    \vspace{-3mm}
\end{figure}

\subsection{Factorized Plenoptic PNG}
\label{ssec:ppng-12}

The vanilla Plenoptic PNG (denoted as PPNG-3) stores a dense 3D volume $\mathbf{V}_i^\mathrm{sin} \in \mathbb{R}^{Q^3 \times D}$ for each frequency, with memory complexity being $Q^3 \times D$. We note that this Fourier-indexed feature representation is axis-aligned and smooth in its index coordinates (sinusoidal space). 
Inspired by the success of tensorial factorization in spatial fields~\cite{Chen2022TensoRFTR}, we  propose leveraging low-rank tensor decomposition techniques to further compress Plenoptic PNG.

Our first approach, PPNG-1, inspired by CP-decomposition~\cite{Chen2022TensoRFTR}, decomposes 3D volumes into a set of triplets of 1D vectors $(\mathbf{v}^r_x, \mathbf{v}^r_z, \mathbf{v}^r_y)$. Specifically, for each $\mathbf{V}_i$ (omitting $\mathrm{sin}$ and $\mathrm{cos}$ superscripts for simplicity), we approximate them using the following equation, instead of directly storing the 3D volume:
\begin{equation}
    \mathbf{V}_i = \sum_r^R
    \mathbf{v}_{i, x}^r \otimes \mathbf{v}_{i, y}^r \otimes  \mathbf{v}_{i, z}^r
    \label{eq:ppng_1_to_3} \text{\ \ (PPNG-1)},
\end{equation}
where $\otimes$ is the outer product, $R$ is the total number of triplet components. PPNG-1 has a memory complexity being $Q\times 3\times R \times D$ for each volume and offers the most compressed representation among all variants with some trade-off in training speed and quality.

Our second approach, PPNG-2, incorporates tri-plane decomposition to approximate 3D volumes into a set of triplets of 2D feature planes $(\mathbf{v}_{xy}, \mathbf{v}_{xz}, \mathbf{v}_{yz})$:
\begin{equation}
\label{eq:ppng_2_to_3} 
\mathbf{V}_i =
\sum_r^R \mathbf{v}_{i, xy}^r \otimes \mathbf{v}_{i, xz}^r \otimes \mathbf{v}_{i, yz}^r
\text{\ \ (PPNG-2)},
\end{equation}
PPNG-2's memory complexity is $Q^2 \times 3 \times R \times D$ for each volume. It offers a good balance between quality and model size, positioned between PPNG-1 and PPNG-3.

During inference, decoding PPNG-1 and PPNG-2 into the PPNG-3 tensor and loading them into the renderer takes $\mathbb{O}(Q^3 \times R)$ time and can be easily parallelized, making the decoding efficient. Figure~\ref{fig:ppng_1_2} depicts the two representations for one single volume.

\subsection{Encoding and Implementation Details}
\label{ssec:training}

During the training/encoding stage, given a collection of posed images, we jointly train our Fourier feature volume $\mathbf{V}$ 
and our shallow MLP network $g_\theta$ with volume rendering. \shenlongdone{f is not mentioned before, should be g?}
We minimize Huber-loss between the volume-rendered pixel colors and observed pixel colors for its robustness to outliers. 

We set volume quantization size $Q=80$, \# of factorized components $R=8$ for PPNG-1, \# of factorized components $R=2$ for PPNG-2, \# of frequency levels $L=4$ and feature dimension $D=4$ throughout all implementations. 
We implement PPNG-1, 2 and 3 in \textit{tiny-cuda-nn}~\cite{Mller2022tinycudann}, which supports voxel-based density caching~\cite{Mller2022InstantNG} for accelerated ray integration.

The input to our MLP is a feature vector of size $F= 32 = 2 \times L \times C$. A single linear layer produces a density value and a 15-length feature vector.  These values and the degree three spherical harmonics (length 16) of the viewing direction are input to a 2-layer MLP that outputs RGB color and contains a size 16 hidden layer.  In total, our MLP contains 1,072 parameters.  Our experiments ablate using different sized networks.

With the chosen parameters, we have parameter size of $125 KB$ for PPNG-1, $2.45 MB$ for PPNG-2 and $32.7 MB$ for PPNG-3 using half-precision floating points (including shallow MLP weights). 
We encode voxel-based density cache using run-length encoding~(RLE). This typically results in additional $50 KB$ to $150 KB$ \shenlongdone{result in 50-150 KB size drop?} depending on the complexity of the scene. 
We use CBOR~\cite{Bormann2013ConciseBO} to aggregate the Fourier feature parameters, shallow MLP weights, and the voxel-based density caches into a single binary file. We emphasize that all three approaches are directly encoded end-to-end and there are {no additional processes} (a.k.a baking) in converting the optimized PPNG into a binary encoding. Both RLE and CBOR are very efficient to decode; for PPNG, we observe a decoding time of less than 20 ms for both RLE and CBOR decodings combined.

\subsection{Real-time, Interactive and Portable Viewing}
\label{ssec:rendering}
What sets PPNG apart from other spatial volume feature-based NeRF methods~\cite{Chen2022TensoRFTR, Mller2022InstantNG} is its significantly more compact Fourier feature volume ($80^3$ vs $512^3$). This enables us to store and render it directly as a GL 3D texture on low-cost GPUs with limited memory, such as those in mobile devices. Inspired by this, we implement 
PPNG representations in real-time by porting the volume rendering of PPNG in a traditional GL pipeline using WebGL2 with GLSL. Figure~\ref{fig:rendering} illustrates this process. 
Given a binary PPNG file, our decoder first checks which PPNG type (among PPNG-1, 2, and 3) that the binary file contains. If the given binary is PPNG-1 or PPNG-2, we efficiently convert it to PPNG-3 using Eq~\ref{eq:ppng_1_to_3} or Eq~\ref{eq:ppng_2_to_3} respectively. The conversion is parallelized with GLSL-based code. 

Given PPNG-3, we load each volumetric Fourier feature $V$ as a 3D texture image. Since we set $D = 4$, we can use a texture format set as RGBA to load each volume into a single 3D texture image. We set the texture filtering parameter set to linear, which enables tri-linear interpolation for texture sampling the loaded volumes in GLSL.  
We then load RLE encoded voxel-based density cache for empty-space skipping, by decoding it as occupancy grid. 
Similar to the volumetric Fourier feature volumes, we load the occupancy grid as another single-channel 3D texture image. 
Finally, we load shallow MLP by chunking the MLP into set of 4x4 matrices (\texttt{Mat4}) for faster inference. 

At render time, we use a fragment shader to perform volume rendering. From the camera origin, we cast a ray to each pixel in world space, and densely sample along the ray. For each sample, we check if it is occupied with the occupancy grid. If occupied, we query density and color at the sampled point using the method described in Section~\ref{ssec:ppng}. The sampled color and density are integrated with volume rendering equation~\cite{Mildenhall2020NeRFRS} and are terminated if accumulated transmittance falls below a threshold. 
We include a GLSL implementation in the supplementary material. 
\begin{table*}[h]
\caption{{\bf Quantitative Evaluation of Our Method on Various Datasets:} Implicit methods typically feature a smaller model size and better performance but require specialized renderers or hardware (e.g., CUDA GPUs). In contrast, web-ready methods are not memory-efficient. Our approach achieves a good balance, particularly excelling in model size. It tends to yield better rendering quality in object-centric scenes than in unbounded scenes. $^\ast$ indicates results from us.
}
\vspace{-3mm}
\resizebox{\textwidth}{!}{
\begin{tabular}{lcc|ccccccccccccccccccc}
\toprule
\multirow{2}{*}{Model Type} &&
\multirow{2}{*}{Real-time Web} &&
\multicolumn{6}{c}{Synthetic NeRF} &&
\multicolumn{3}{c}{Blended MVS} &&
\multicolumn{3}{c}{Tanks and Temples} \\
&&&&
PSNR~$\uparrow$ & SSIM~$\uparrow$ & LPIPS~$\downarrow$ & Size & FPS & Training Time && 
PSNR~$\uparrow$ & SSIM~$\uparrow$ & LPIPS~$\downarrow$ && 
PSNR~$\uparrow$ & SSIM~$\uparrow$ & LPIPS~$\downarrow$ \\
\midrule
NeRF~\cite{Mildenhall2020NeRFRS}  && \xmark &&
31.01 & 0.947 & 0.081 & 5 MB & - & 35 hrs &&
24.15 & 0.828 & 0.192 &&
25.78 & 0.864 & 0.198 \\
NSVF~\cite{Liu2020NeuralSV} && \xmark &&
31.74 & 0.953 & 0.047 & -  & - & 48 hrs (8 GPU) &&
26.90 & 0.898 & 0.113 &&
28.40 & 0.900 & 0.153 \\
Diver~\cite{Wu2021DIVeRRA} && \xmark &&
32.32 & 0.960 & 0.032 & 67.8 MB & - & 50 hrs &&
27.25 & 0.910 & 0.073 &&
28.18 & 0.912 & 0.116 \\
TensoRF-CP~\cite{Chen2022TensoRFTR} && \xmark &&
31.56 & 0.949 & 0.041 & 3.9 MB & - & 25.2 min &&
- & - & - &&
27.59 & 0.897 & 0.144 \\
TensoRF-VM~\cite{Chen2022TensoRFTR} && \xmark &&
33.14 & 0.963 & 0.027 & 71.8 MB & - & 17.4 min &&
- & - & - &&
28.56 & 0.920 & 0.125 \\
InstantNGP$^\ast$~\cite{Mller2022InstantNG} && \xmark &&
32.82 & 0.960 & 0.037 & 11.6 MB & - & 5 min &&
28.70 & 0.943 & 0.037 &&
28.36 & 0.930 & 0.099 \\
Dictionary Field~\cite{Chen2023DictionaryFL} && \xmark &&
33.14 & 0.961 & - & 5.1MB & - & 12.2 min &&
- & - & - &&
29.00 & 0.938 & - \\
WaveletNeRF~\cite{Rho2022MaskedWR} && \xmark &&
31.94 & - & - & 846 KB & - & 24.0 min &&
- & - & - &&
27.77 & - & - \\
WaveletNeRF$^\ast$~\cite{Rho2022MaskedWR} && \xmark &&
25.90 & 0.891 & 0.142 & 199 KB & - & 23.6 min &&
- & - & - &&
- & - & - \\
\midrule
PlenOctree~\cite{Yu2021PlenoxelsRF} && \cmark &&
31.71 & 0.958 & 0.049 & 1976 MB & 168 & 50 hrs &&
- & - & - &&
27.99 & 0.917 & 0.131 \\
SNeRG~\cite{Hedman2021BakingNR} && \cmark &&
30.4 & 0.950 & 0.050 & 87 MB & 502 & 15 hrs &&
- & - & - &&
- & - & - \\
MobileNeRF~\cite{Chen2022MobileNeRFET} && \cmark &&
30.9 & 0.947 & 0.062 & 126 MB & 762 & 20 hrs &&
- & - & - &&
- & - & - \\
Re-Rend~\cite{Rojas2023ReReNDRR} && \cmark &&
29.0 & 0.934 & 0.080 & 199 MB & 1013 & 60 hrs &&
- & - & - &&
- & - & - \\
Gaussian Splatting$^\ast$~\cite{Kerbl20233DGS} && \cmark &&
33.80 & 0.970 & 0.030 & 67.3 MB & - & 4.9 min &&
24.95 & 0.867 & 0.109 &&
27.94 & 0.930 & 0.097 \\
PPNG-3$^\ast$ && \cmark &&
31.90 & 0.949 & 0.044 & 32.8 MB & 128 & 4.9 min &&
26.89 & 0.909 & 0.068 &&
27.83 & 0.925 & 0.112 \\
PPNG-2$^\ast$ && \cmark &&
30.99 & 0.944 & 0.053 & 2.49 MB & 127 & 9.8 min &&
26.53 & 0.894 & 0.080 &&
27.23 & 0.912 & 0.136 \\
PPNG-1$^\ast$ && \cmark &&
28.89 & 0.926 & 0.080 & 151 KB & 127 & 13.1 min &&
24.77 & 0.855 & 0.134 &&
25.68 & 0.892 & 0.178 \\
\bottomrule
\end{tabular}
}
\label{tab:quantitative}
\end{table*}





\newcommand{\zoomimage}[2]{
    \tikz{
        \node[draw opacity=1](ppng) at (0, 0) {\includegraphics[width=0.20\textwidth]{#1}};
        \node[draw=black, draw opacity=1.0, line width=.3mm, inner sep=0pt](ppng) at (-.9, -.9) {
            \includegraphics[trim=#2, clip, width=0.08\textwidth]{#1}
        };
    }
}
\newcommand{\zoomimagewithtrim}[2]{
    \tikz{
        \node[draw opacity=1](ppng) at (0, 0) {\includegraphics[trim={0cm, 0cm, 0cm, 5cm}, clip, width=0.20\textwidth]{#1}};
        \node[draw=black, draw opacity=1.0, line width=.3mm, inner sep=0pt](ppng) at (-.9, -.9) {
            \includegraphics[trim=#2, clip, width=0.08\textwidth]{#1}
        };
    }
}
\begin{figure*}[t]
    \centering
    \resizebox{0.95\textwidth}{!}{
    \begin{tabular}{c@{}c@{}c@{}c@{}c}
        \zoomimage{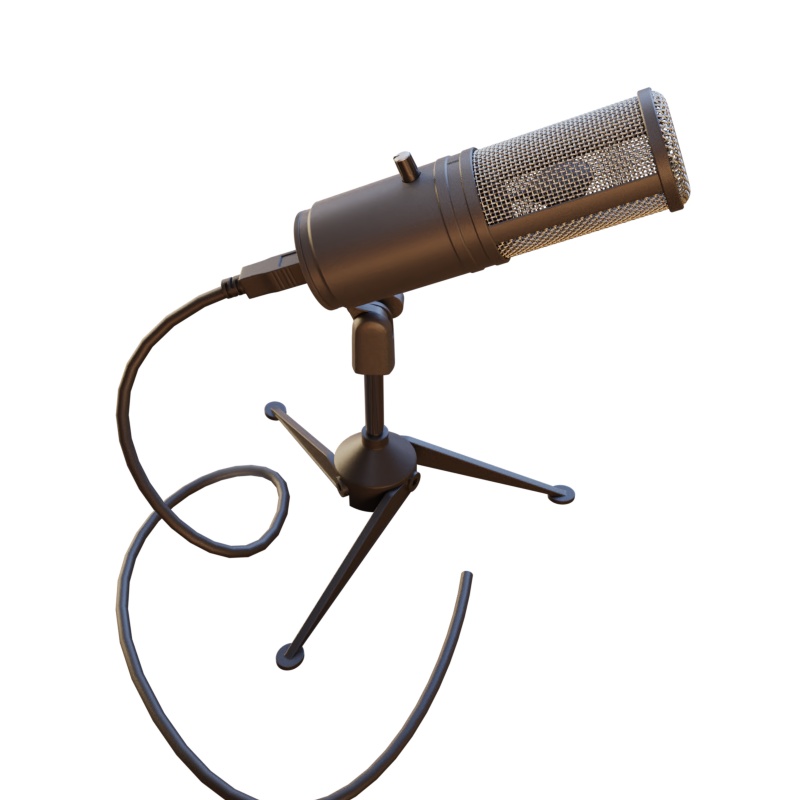}{20cm 20.5cm 4cm 3.5cm} & 
        \zoomimage{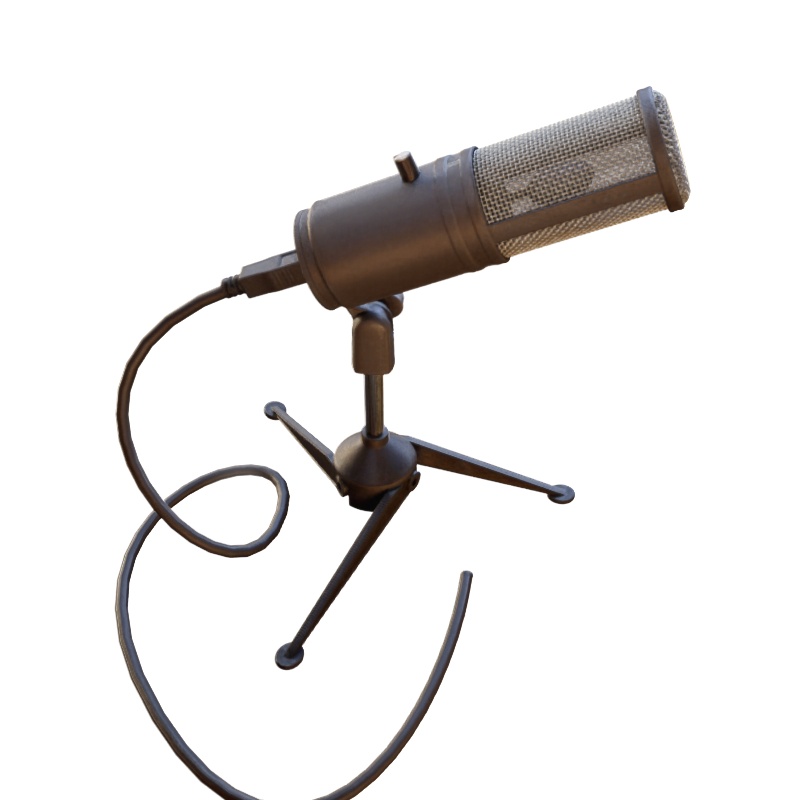}{20cm 20.5cm 4cm 3.5cm} & 
        \zoomimage{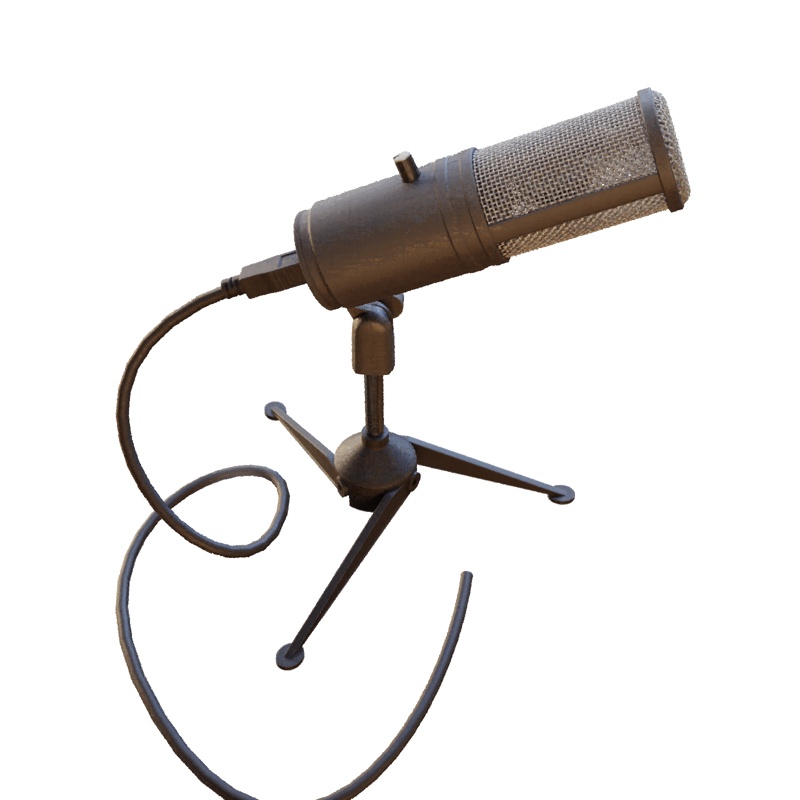}{20cm 20.5cm 4cm 3.5cm} & 
        \zoomimage{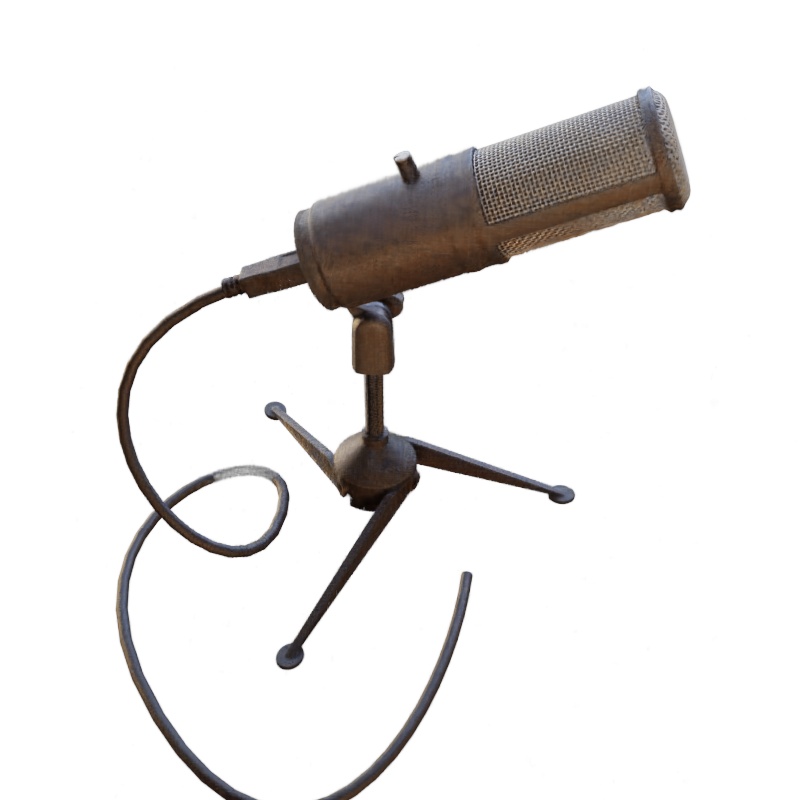}{20cm 20.5cm 4cm 3.5cm} & 
        \zoomimage{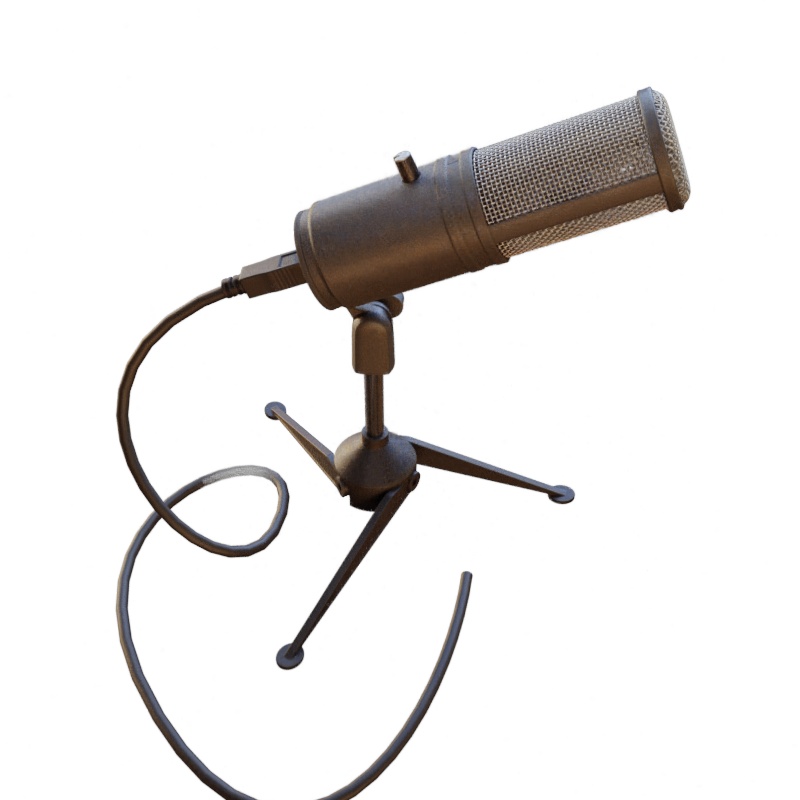}{20cm 20.5cm 4cm 3.5cm} \\
        Ground Truth & SNeRG~(15 hrs) & MobileNeRF~(21 hrs) & PPNG-1~(12.8 mins) & PPNG-2~(9.8 mins)\\
        & 18.0 MB / 32.6 dB & 52.7 MB / 32.5 dB  & 144 KB / 32.1 dB  & 2.5 MB / 34.1 dB  
        \vspace{-2mm} \\
   \zoomimagewithtrim{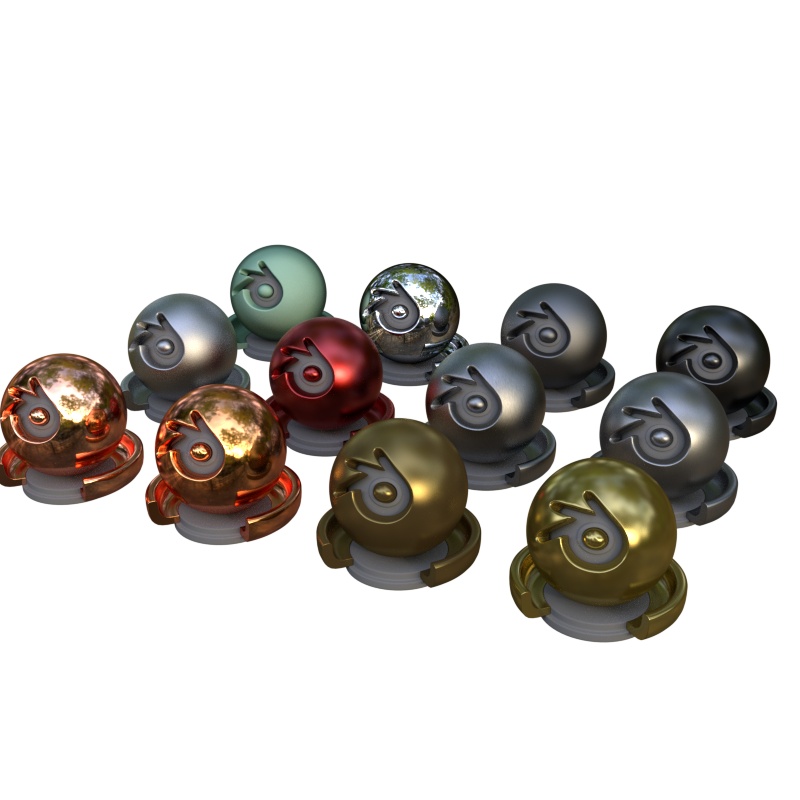}{6cm 11cm 18cm 13cm} & 
        \zoomimagewithtrim{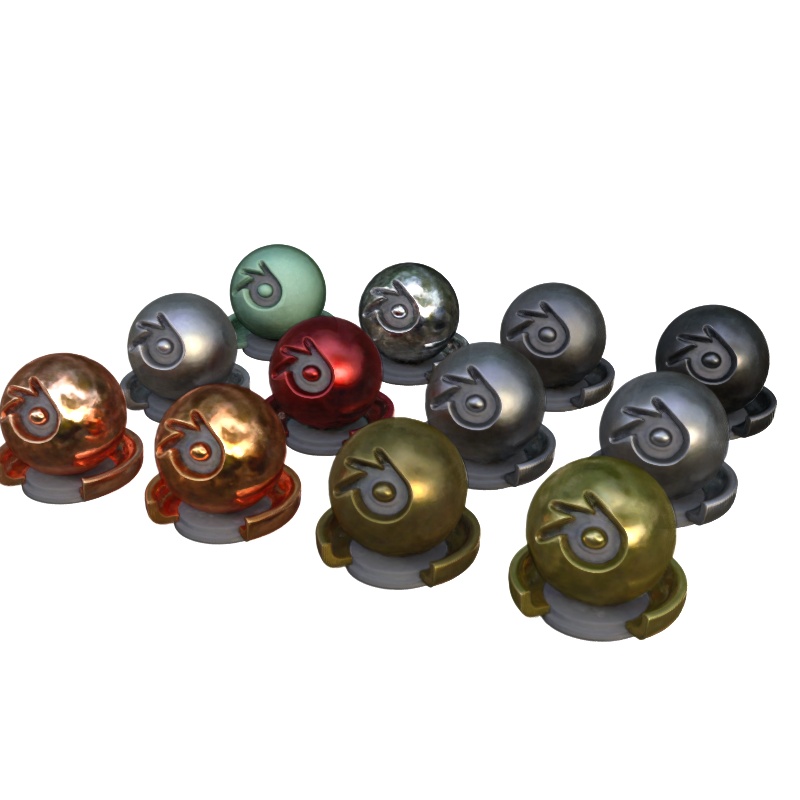}{6cm 11cm 18cm 13cm} & 
        \zoomimagewithtrim{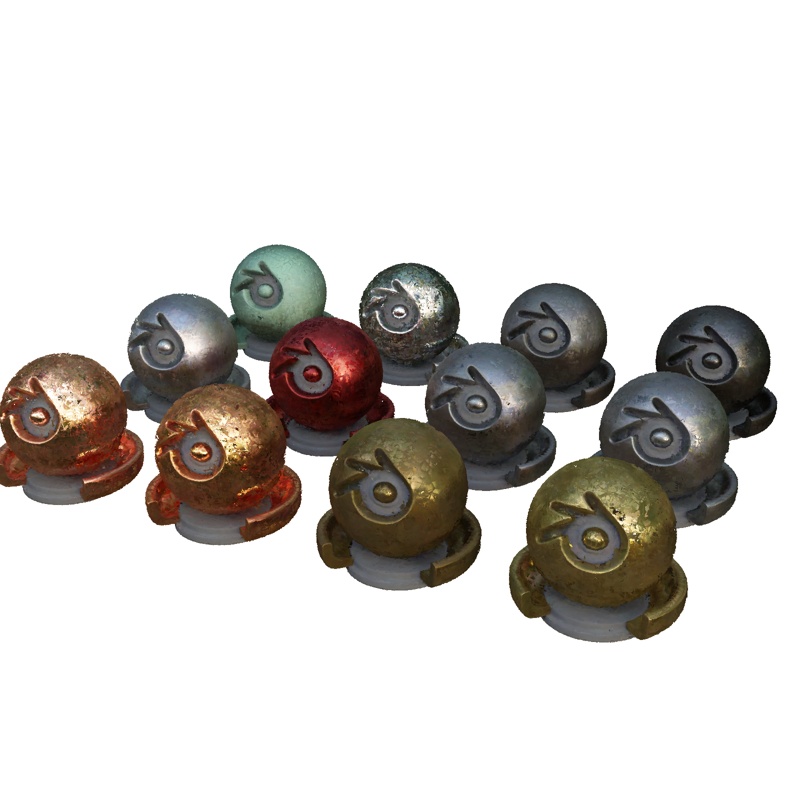}{6cm 11cm 18cm 13cm} & 
        \zoomimagewithtrim{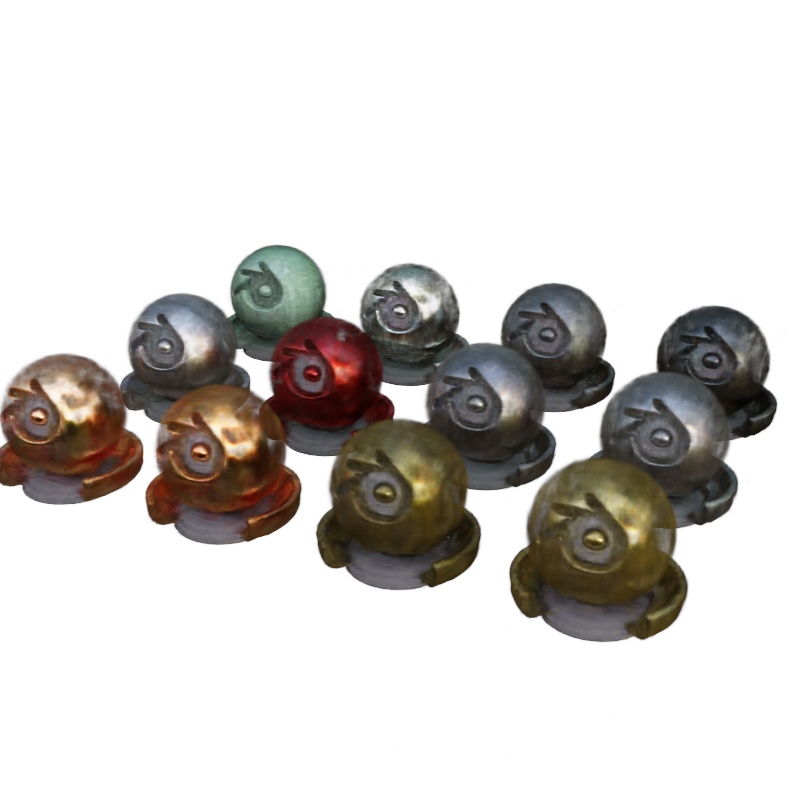}{6cm 11cm 18cm 13cm} & 
        \zoomimagewithtrim{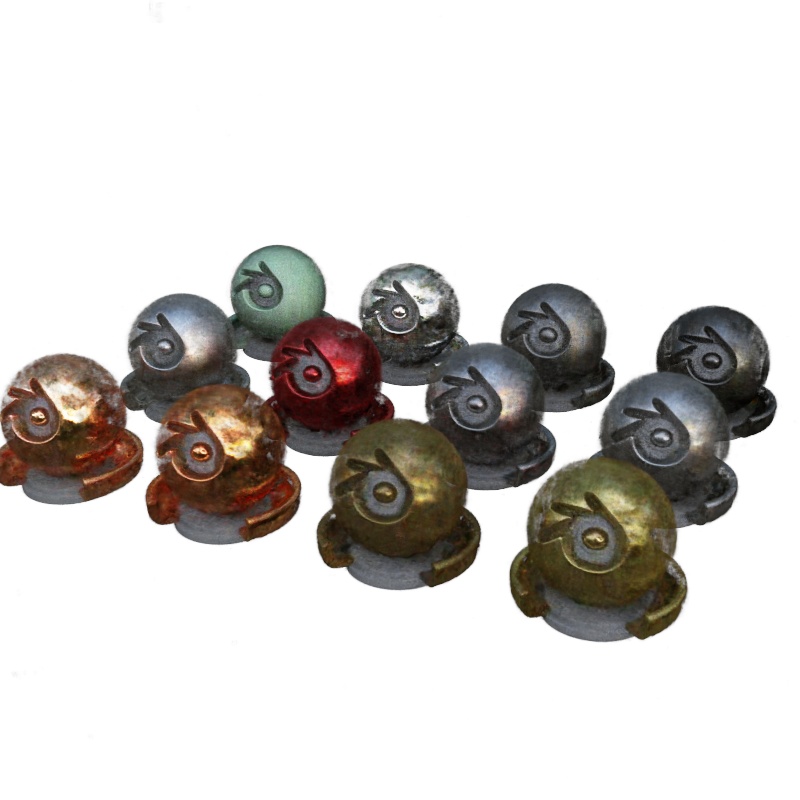}{6cm 11cm 18cm 13cm} \\
        Ground Truth & SNeRG~(15 hrs) & MobileNeRF~(21 hrs) & PPNG-1~(13.9 mins) & PPNG-2~(10.0 mins)\\
        & 62.8 MB / 27.2 dB & 191 MB / 26.7 dB & 149 KB / 26.9 dB & 2.5 MB / 27.5 dB \\
        \vspace{-3mm}
        \zoomimage{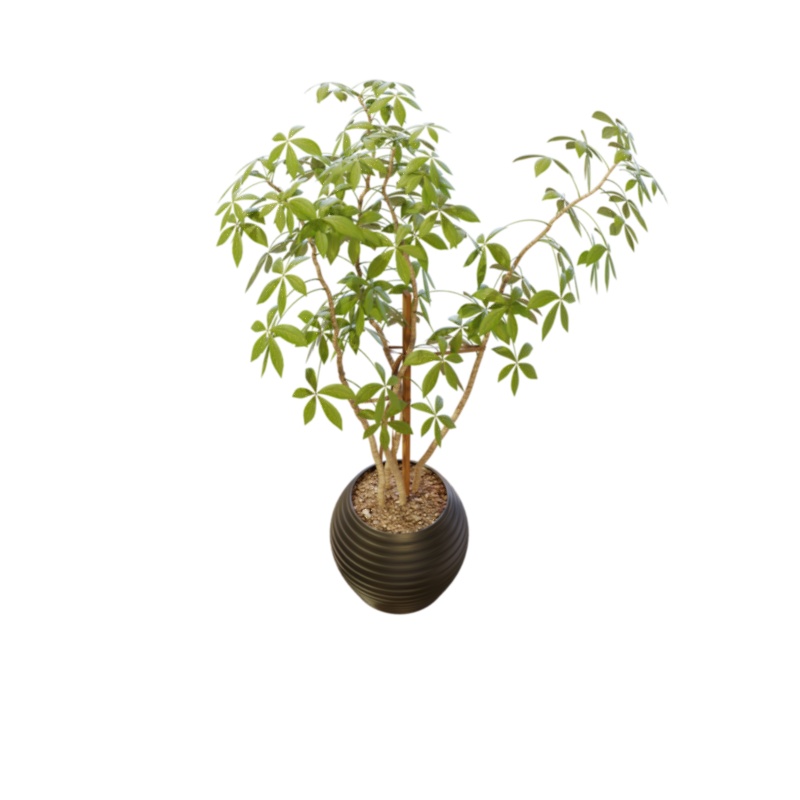}{12cm 12cm 12cm 12cm} & 
        \zoomimage{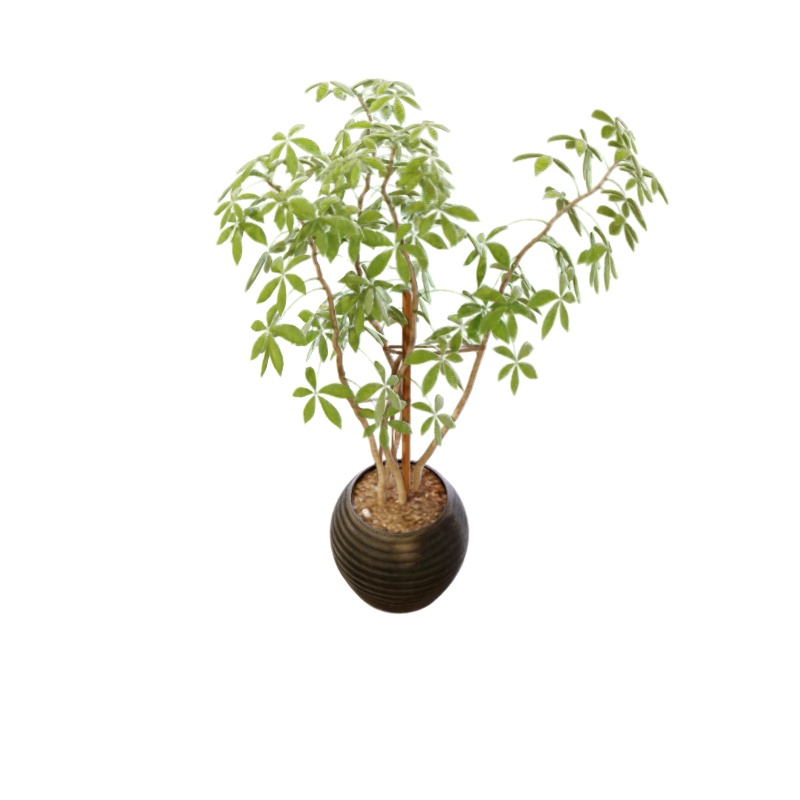}{12cm 12cm 12cm 12cm} & \zoomimage{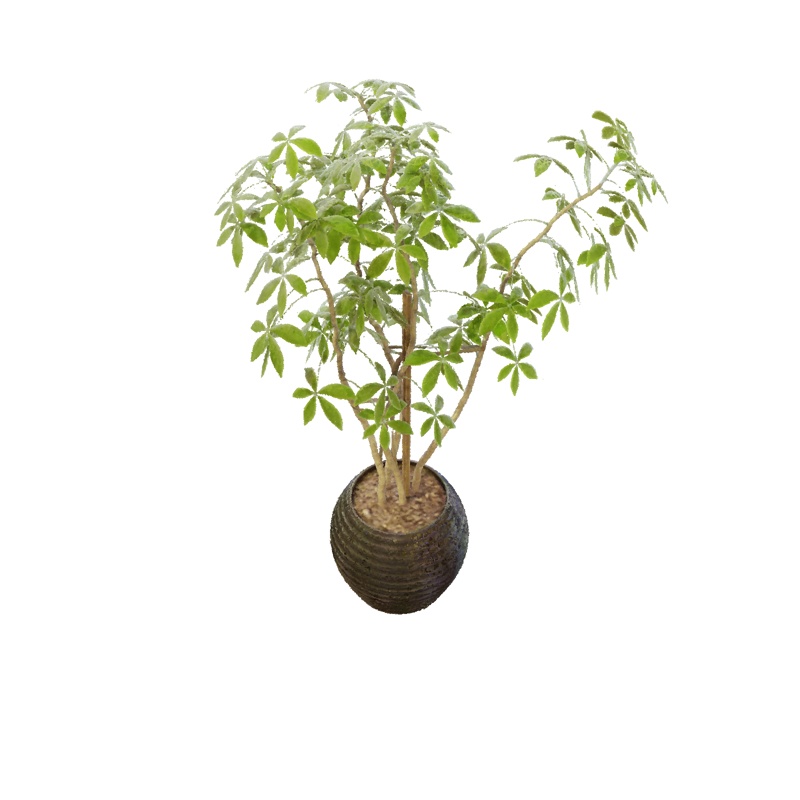}{12cm 12cm 12cm 12cm} & 
        \zoomimage{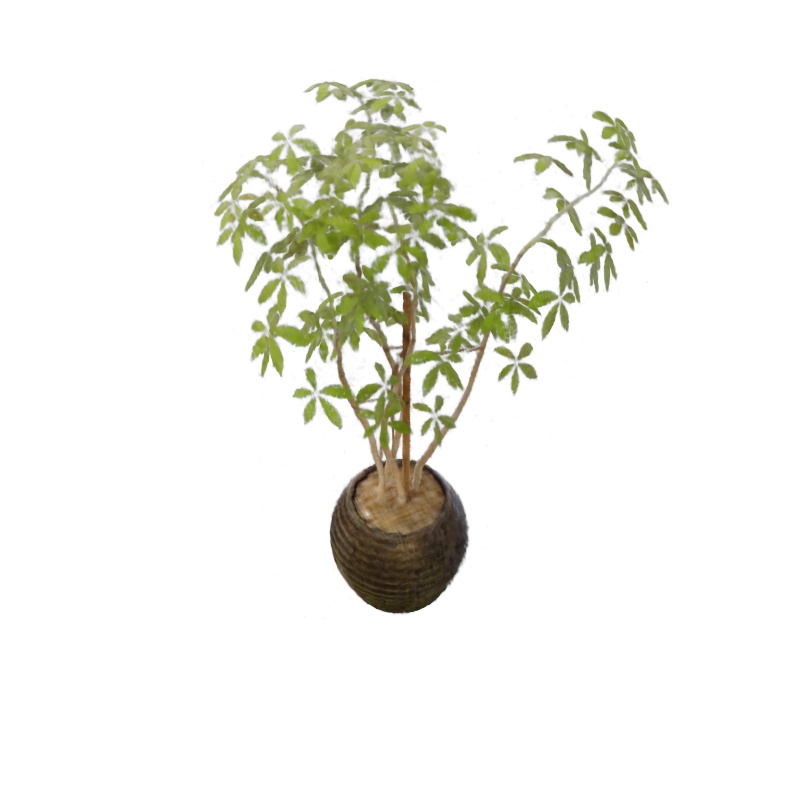}{12cm 12cm 12cm 12cm} & 
        \zoomimage{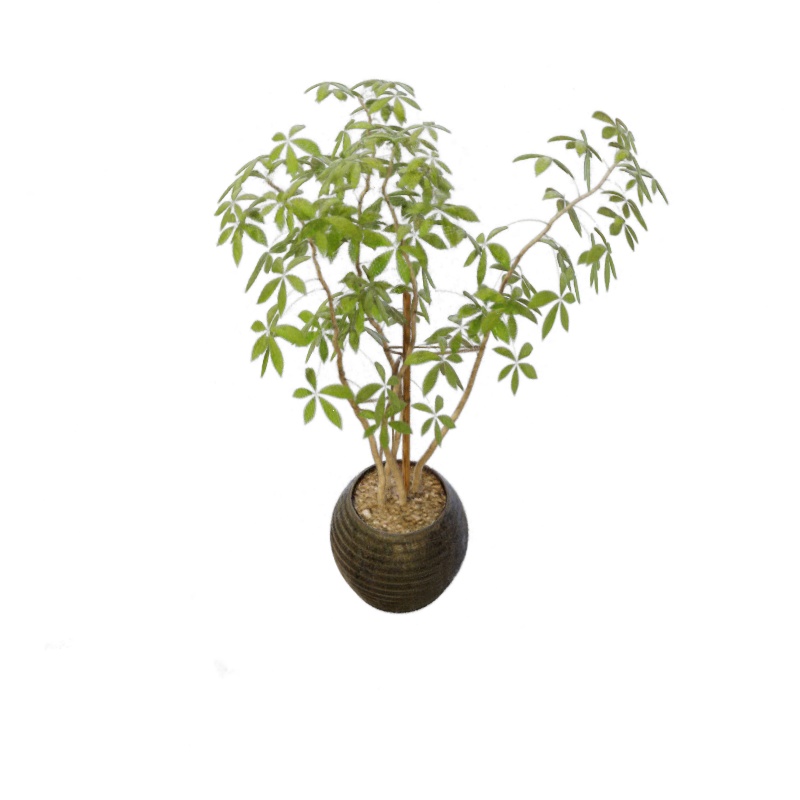}{12cm 12cm 12cm 12cm} \\
        Ground Truth & SNeRG~(15 hrs) & MobileNeRF~(21 hrs) & PPNG-1~(11.9 mins) & PPNG-2~(9.4 mins) \\
        & 30.0 MB / 29.3 dB & 81 MB / 30.2 dB & 150 KB / 28.2 dB & 2.5 MB / 31.0 dB
    \end{tabular}
    }
    \vspace{-3mm}
    \caption{{\bf Qualitative Comparison on the Synthetic NeRF Dataset:} We show qualitative results and compare real-time NeRF models (SNeRG and MobileNeRF) in terms of training time, model size, and quality. PPNG-1 delivers similar or superior visual quality compared to other web-friendly baselines while being at least {\bf 120x smaller in model size}. PPNG-2 offers enhanced quality with a model size more than 8x smaller. 
    }
    \label{fig:qualitative}
\end{figure*}

\section{Experiments}
We evaluate our model on multiple object-level datasets to validate the re-rendered models qualitatively and quantitatively. 
Then, we test our real-time renderer on various devices, including a desktop with a GPU, various laptops, and several mobile phones. 
We further conduct several analyses on how different designs contribute to the final performance and finally discuss the current limitations of our approach on unbounded scenes. 

\subsection{Experimental Details}

\begin{table*}[h]
\caption{
{\bf Ablation studies of PPNG on Synthetic NeRF dataset.} 
The reference implementation uses volume resolution $Q=80$, Max Freq = $2^3$, 1 Layer MLP, number of components ($R$) for PPNG-1 set as 8 and PPNG-2 set as 2. We mark significant improvements and losses with {\green Green} and {\red Red} respectively. 
}
\vspace{-3mm}
\resizebox{\textwidth}{!}
{
\begin{tabular}{lcccccccccccc}
\toprule
\multirow{3}{*}{Ablations} && 
\multicolumn{3}{c}{PPNG-1} &&
\multicolumn{3}{c}{PPNG-2} &&
\multicolumn{3}{c}{PPNG-3} \\
&&
PSNR & Size & Training Time && 
PSNR & Size & Training Time && 
PSNR & Size & Training Time \\
&&
({\green{+}}{\red{-}} 0.5) & ({\green{+}}{\red{-}} 20\%) & ({\green{+}}{\red{-}} 20\%) && 
({\green{+}}{\red{-}} 0.5) & ({\green{+}}{\red{-}} 20\%) & ({\green{+}}{\red{-}} 20\%) && 
({\green{+}}{\red{-}} 0.5) & ({\green{+}}{\red{-}} 20\%) & ({\green{+}}{\red{-}} 20\%) \\
\midrule
Reference in Table~\ref{tab:quantitative} &&
28.89 & 151 KB & 13.1 min && 
30.99 & 2.49 MB &  9.8 min && 
31.90 & 32.8 MB & 4.9 min \\
\midrule
Vol. Res. $Q=60$ &&
28.51 & \green 120 KB & 14.9 min && 
\red 30.63 & \green 1.42 MB & 11.0 min && 
\red 31.05 & \green 13.9 MB & 4.0 min \\
Vol. Res. $Q=100$ &&
29.09 & \red 182 KB & 13.3 min && 
31.12 & \red 3.72 MB & 9.2 min && 
32.02 & \red 64.0 MB & 5.2 min \\ \midrule
Max Freq = $2^1$ &&
28.56 & 151  KB & 15.1 min && 
\red 30.12 & 2.49 MB & 11.1 min && 
\red 30.61 & 32.8 MB & 4.1 min \\
Max Freq = $2^5$ &&
\red 28.18 & 151 KB & 12.3 min && 
\red 30.48 & 2.49 MB & 9.8 min && 
31.63 & 32.8 MB & 4.9 min \\ \midrule
2 Layer MLP &&
29.14 & 153 KB & 14.7 min && 
31.04 & 2.49 MB & 9.7 min && 
31.82 & 32.8 MB & 4.5 min \\ \midrule
\# Comp $\times$ 0.5 &&
\red 27.50 & \green 89.2 KB & \green 10.6 min && 
\red 30.09 & \green 1.26 MB & 9.31 min && 
- & - & - \\
\# Comp $\times$ 2 &&
\green 29.87 & \red 274 KB & \red 24.2 min && 
\green 31.50 & \red 4.94 MB & \red 17.0 min && 
- & - & - \\
\bottomrule
\end{tabular}
}
\label{tab:ablation}
\vspace{-3mm}
\end{table*}

\noindent\textbf{Datasets:} 
We evaluate on Synthetic NeRF~\cite{Mildenhall2020NeRFRS}, Blended MVS~\cite{Yao2019BlendedMVSAL} and Tanks and Temples~\cite{Knapitsch2017TanksAT} datasets with a resolution of $800\times800$, $768\times576$, $1920\times1080$ respectively. 
We use the author-provided training/testing splits for Synthetic NeRF, and use processed scenes and train/test splits provided by NSVF~\cite{Liu2020NeuralSV} for Blended MVS and Tanks and Temples. 
We measure PSNR, LPIPS, and SSIM for visual quality, and report rendering speed in FPS, model size in MB, and optimization time for each scene.

\noindent\textbf{Baselines:} 
Our goal is to achieve compact, real-time, and web-compatible neural rendering. To the best of our knowledge, there is no preceding work that achieves our proposed level of compression (KB-level). To ensure a fair comparison, we primarily evaluate and compare our approach against the current best real-time, web-ready NeRF approaches, including SNeRG~\cite{Hedman2021BakingNR}, MobileNeRF~\cite{Chen2022MobileNeRFET} and Re-render~\cite{Rojas2023ReReNDRR}. 
Additionally, we also reference state-of-the-art and classic NeRF approaches~\cite{Mildenhall2020NeRFRS,Liu2020NeuralSV,Yu2021PlenOctreesFR,Wu2021DIVeRRA,Chen2022TensoRFTR} for context, despite them not being in the same categories as our proposed approach.

\begin{table}[htbp!]
\vspace{-3mm}
\centering
\caption{{\bf Rendering Speed on Different Devices.} We report the rendering FPS for PPNG-2 on various devices using the Lego scene~\cite{Mildenhall2020NeRFRS} at an $800\times800$ resolution on a web browser. For mobile devices, all measurements were conducted in battery mode, without external power connected.
}
\vspace{-3mm}
\resizebox{0.6\linewidth}{!}{
\begin{tabular}{l|c}
\toprule
Device & FPS \\
\midrule
iPhone 10 Pro Max& 20 \\
iPhone 14 & 30 \\
iPhone 15 Pro & 50 \\
M1 iPad& 40 \\
M1 Macbook Pro& 45 \\
M3 Macbook Air& 50 \\
M3 Max Macbook Pro& 100 \\
Desktop with Nvidia 3090 GPU & 127 \\
\bottomrule
\end{tabular}
\label{tab:fps}
}
\end{table}
\subsection{Experimental Results}

\noindent\textbf{Qualitative results:}
Figure~\ref{fig:qualitative} presents a comprehensive comparison of qualitative results. We note that the quality is reasonable for all competing algorithms, but our proposed approach achieves a significant reduction in model size (over 40x to 1500x) and over 50x reduction in training time. We would like to particularly highlight that our approach, despite its extreme compactness, effectively captures thin structures (such as the stems of plants), reflective materials (e.g., metal balls), as well as repetitive high-frequency patterns.

\noindent\textbf{Quantitative results:} 
We conduct two major quantitative evaluations. 
The first focuses on real-time, web-compatible approaches using the Synthetic NeRF dataset. Figure~\ref{fig:quantitative} displays the results. This study shows that our smallest model, PPNG1, significantly outperforms all other methods in terms of model size ({\bf being 2-3 orders of magnitude smaller}) and training speed (20 times faster), while maintaining competitive rendering quality (PSNR = 28.5 dB). Our most robust model, PPNG3, has a model size over 5 times smaller and a training speed 40 times faster, and it achieves the best rendering quality among these real-time NeRF methods (PSNR = 31.9 db). Importantly, our GPU memory during rendering is only {\bf 47 MB}, making it particularly suitable for low-cost mobile devices ({ $>$10 times smaller than other competing methods}). More details on vram consumption can be found in appendix.  

\newcommand{\zoomimageunbounded}[3]{
    \tikz{
        \node[draw opacity=1](ppng) at (0, 0) {\includegraphics[width=0.25\textwidth]{#1}};
        \node[draw=black, draw opacity=1.0, line width=.3mm, inner sep=0pt](ppng) at (-1.2, -.7) {
            \includegraphics[trim=#2, clip, width=0.07\textwidth]{#1}
        };
        \node[draw=black, draw opacity=1.0, line width=.3mm, inner sep=0pt](ppng) at (1.2, .7) {
            \includegraphics[trim=#3, clip, width=0.07\textwidth]{#1}
        };
    }
}
\begin{figure*}[t]
    \centering
    \resizebox{0.93\textwidth}{!}{
    \begin{tabular}{c@{}c@{}c@{}c}
        \zoomimageunbounded{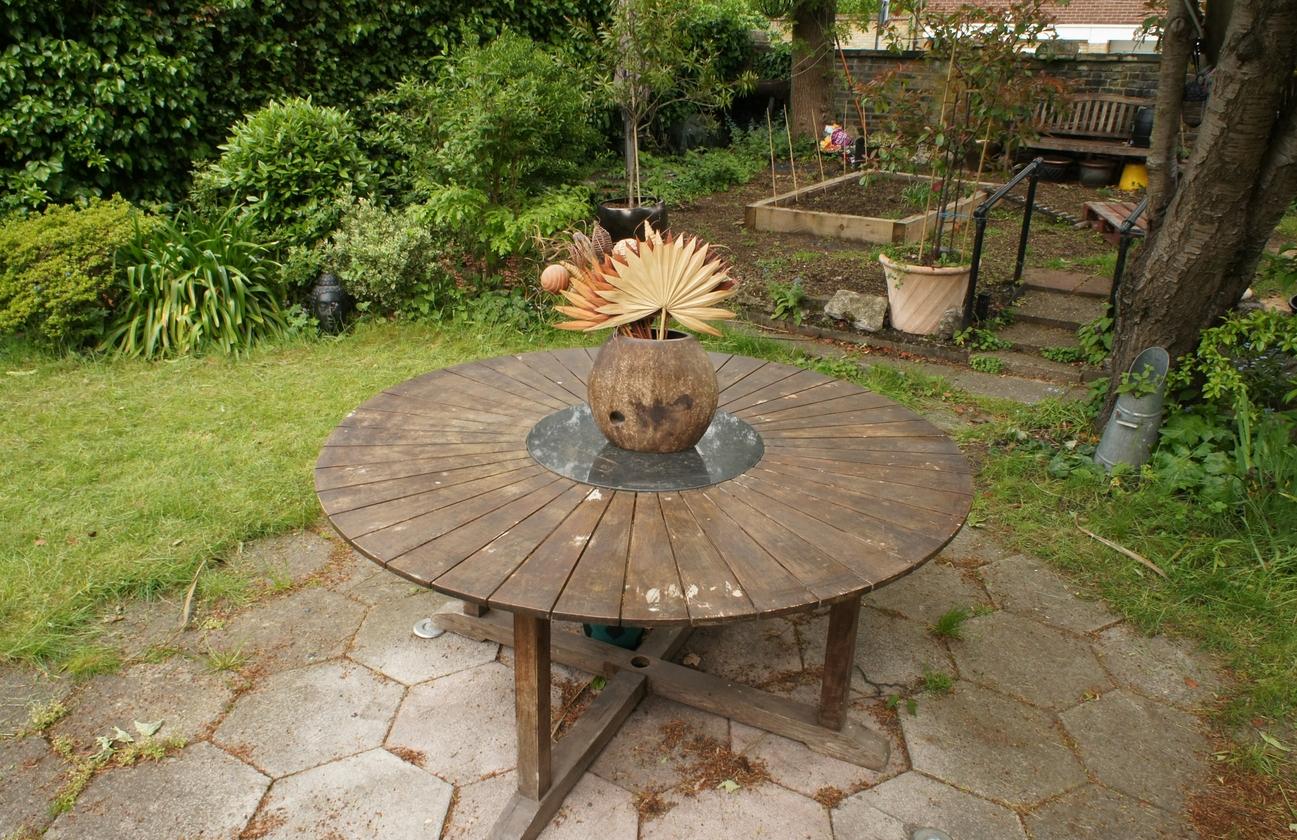}{15cm 8cm 25cm 16cm}{15cm 20cm 25cm 4cm} &
        \zoomimageunbounded{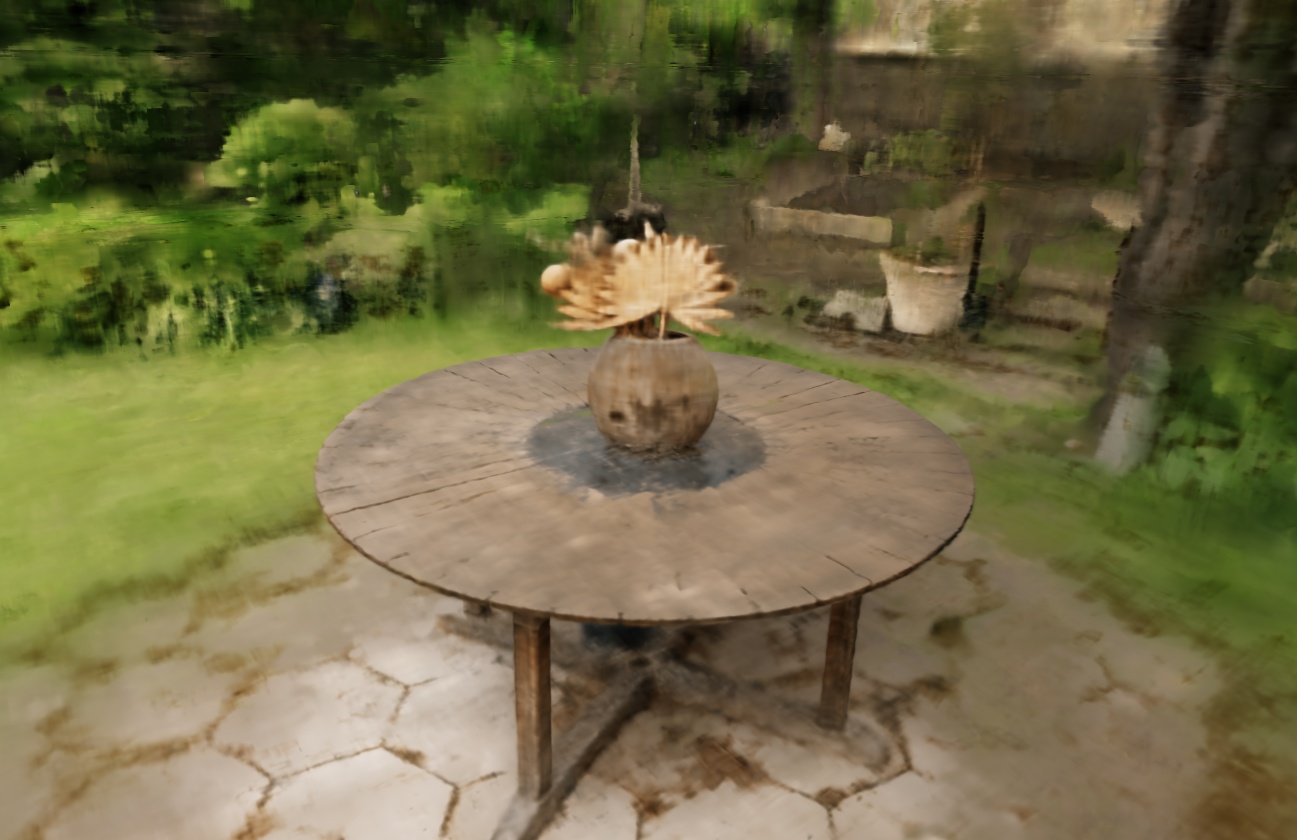}{15cm 8cm 25cm 16cm}{15cm 20cm 25cm 4cm} &
        \zoomimageunbounded{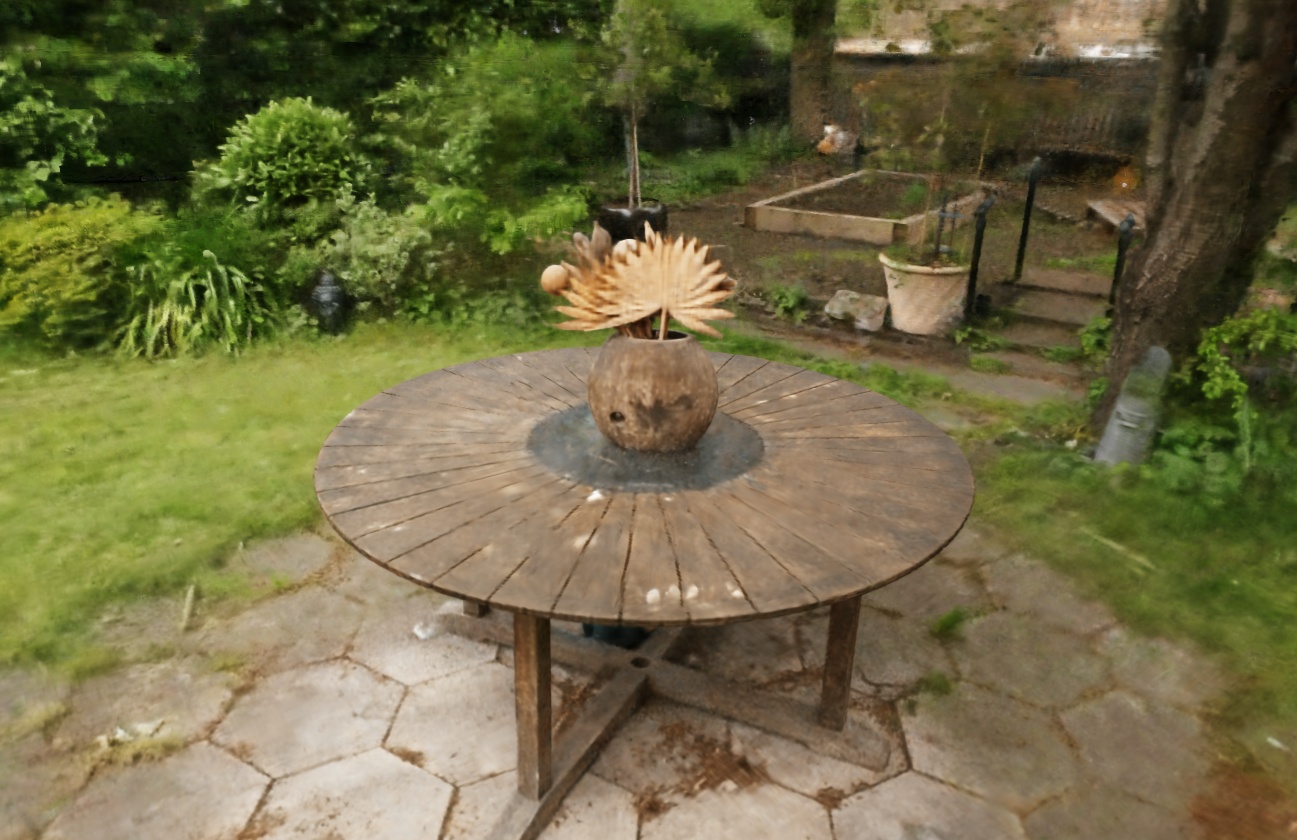}{15cm 8cm 25cm 16cm}{15cm 20cm 25cm 4cm} &
        \zoomimageunbounded{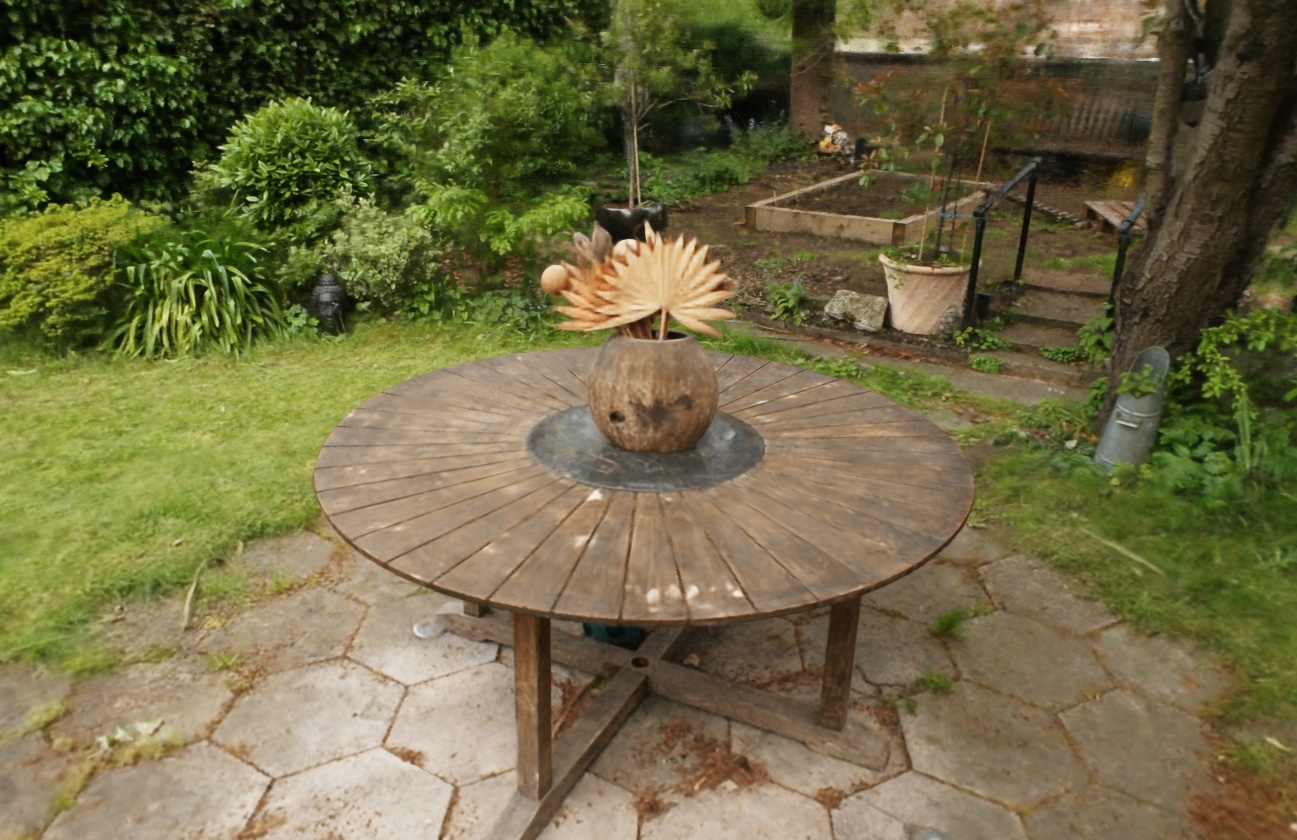}{15cm 8cm 25cm 16cm}{15cm 20cm 25cm 4cm} \\
        \zoomimageunbounded{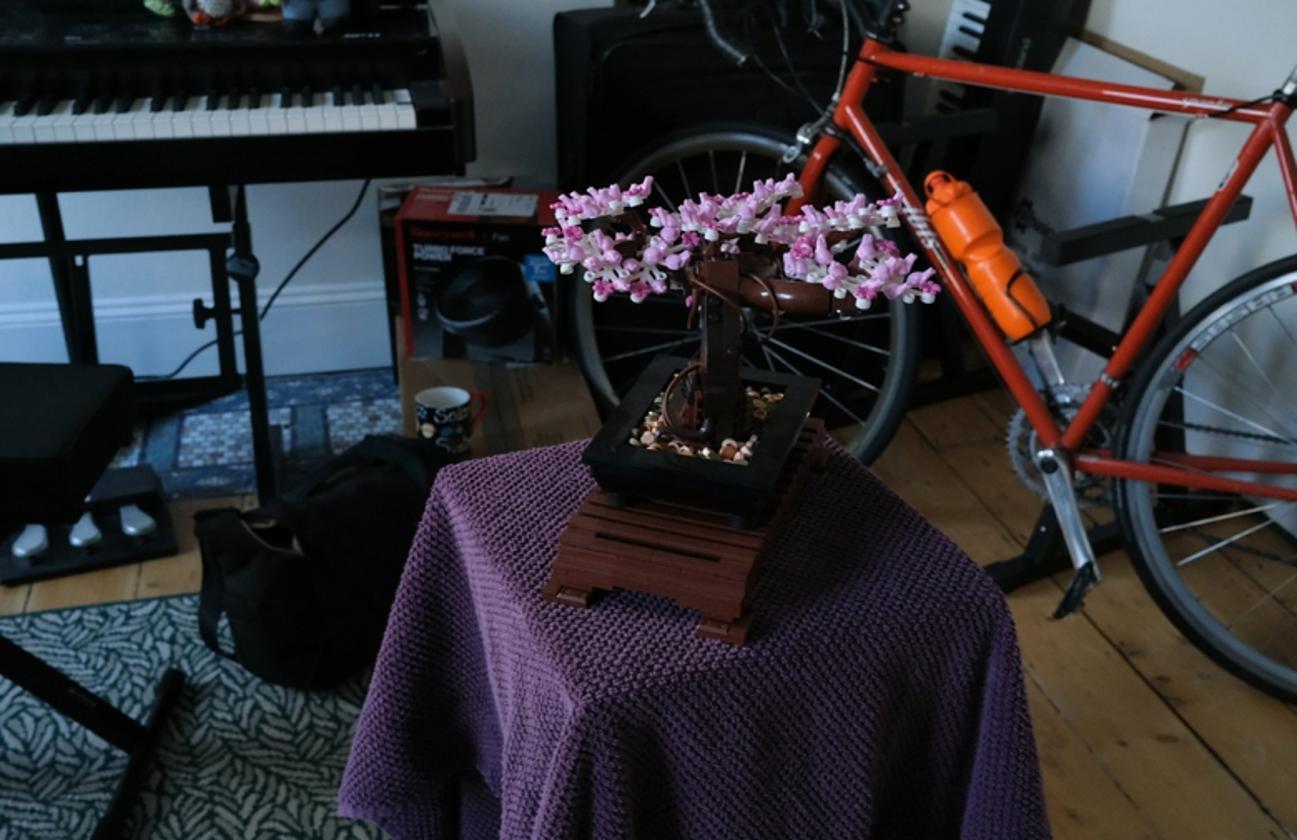}{15cm 8cm 25cm 16cm}{10cm 22cm 30cm 2cm} &
        \zoomimageunbounded{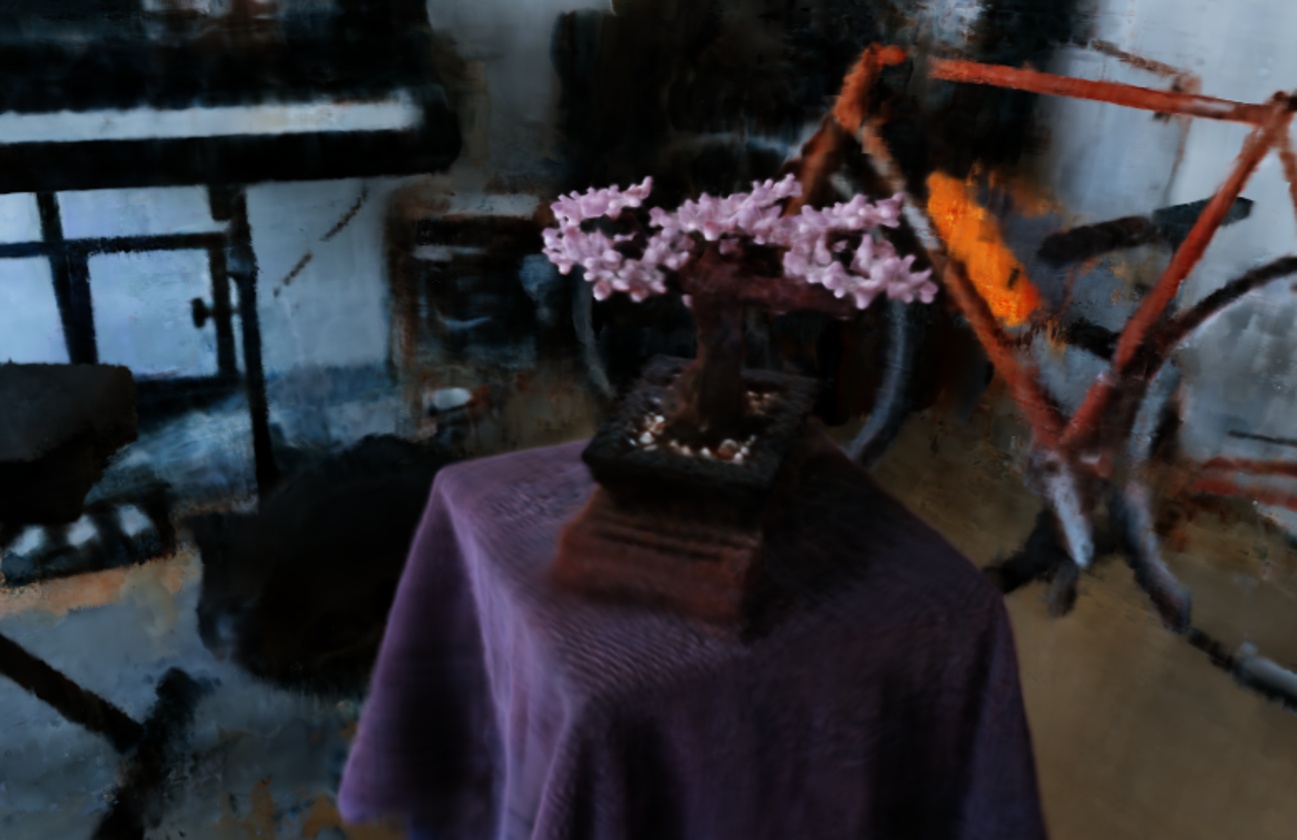}{15cm 8cm 25cm 16cm}{10cm 22cm 30cm 2cm} &
        \zoomimageunbounded{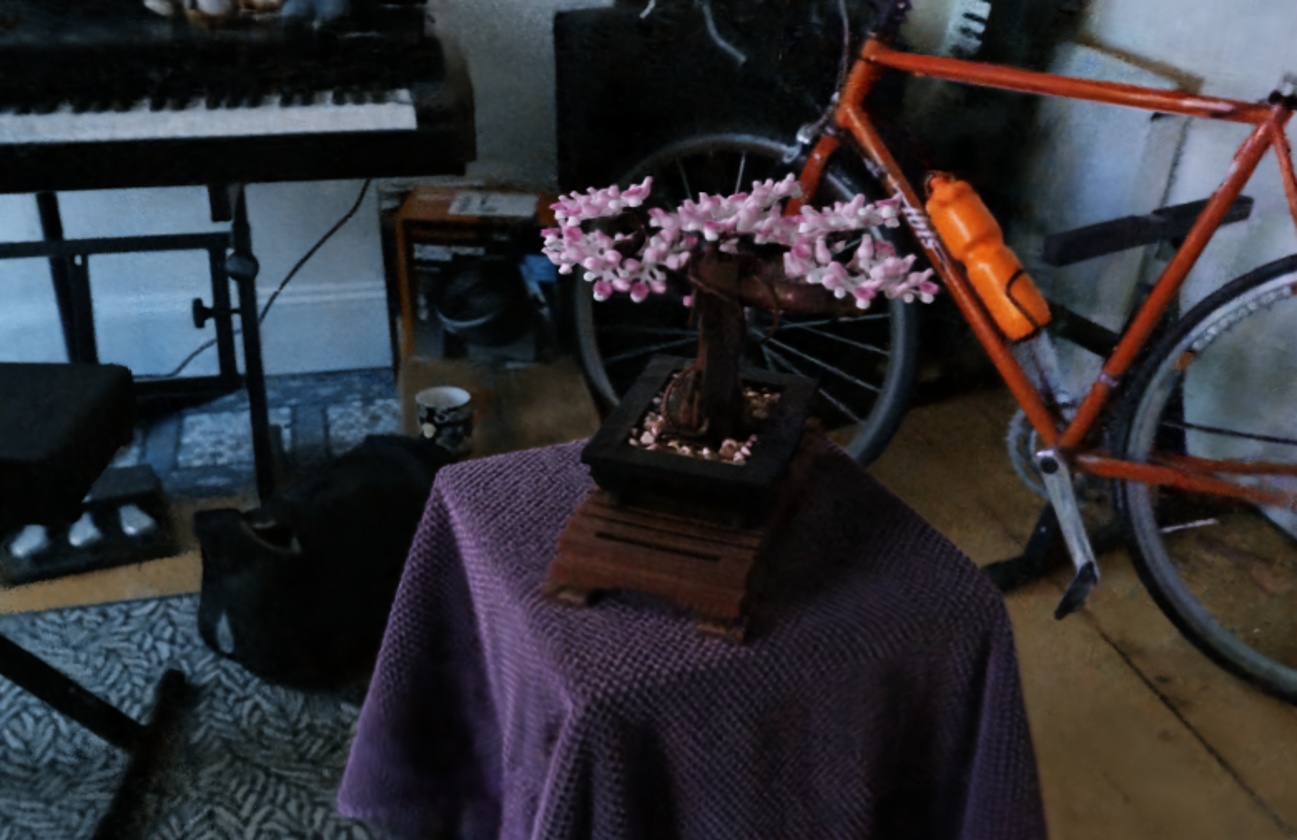}{15cm 8cm 25cm 16cm}{10cm 22cm 30cm 2cm} &
        \zoomimageunbounded{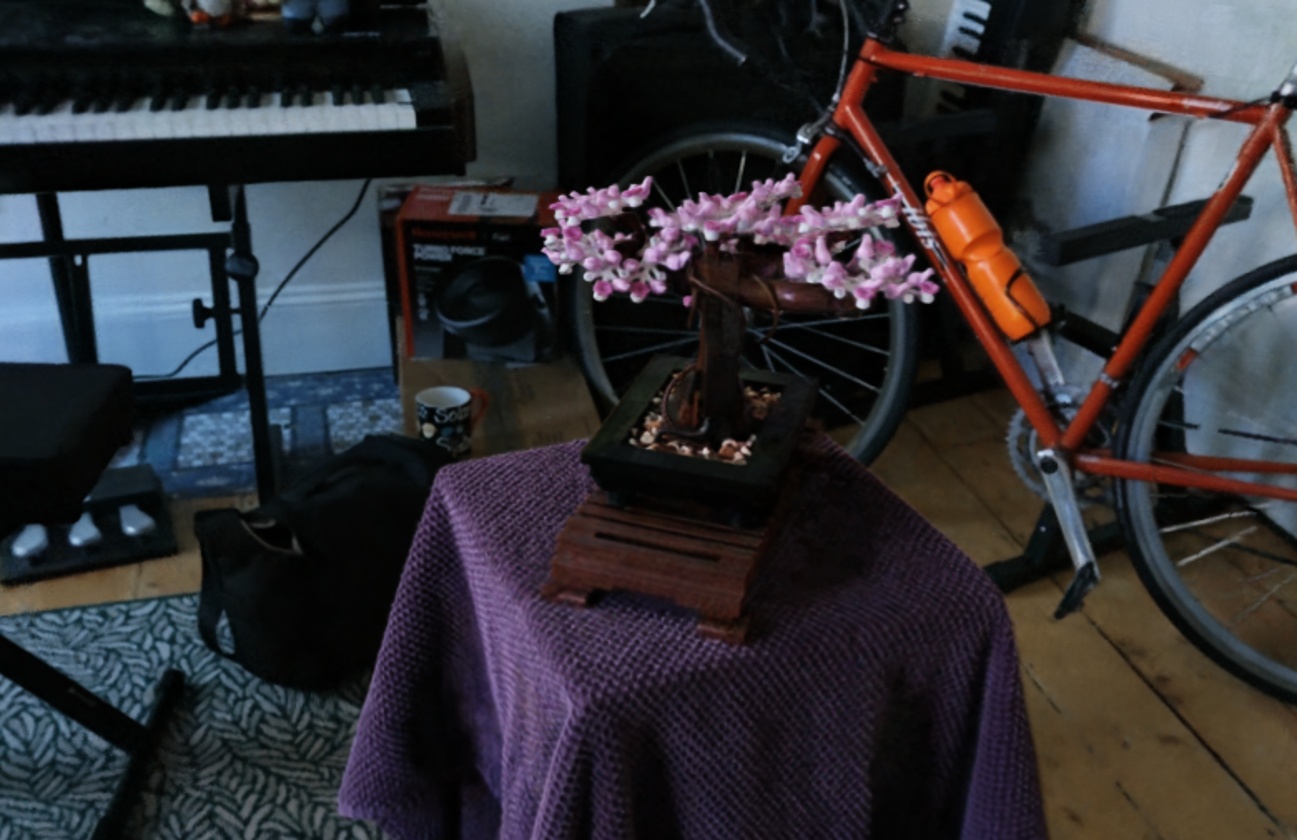}{15cm 8cm 25cm 16cm}{10cm 22cm 30cm 2cm} \\
         Ground Truth & PPNG-1 (512 KB) & PPNG-2 (2.9 MB) & PPNG-3 (33.3 MB)
    \end{tabular}
    }
    \vspace{-3mm}
    \caption{{\bf Qualitative Results on Unbounded 360$^\circ$ Scenes:} We highlight the background region in the top right corner and the central region in the bottom left corner. PPNG-3 provides compelling results with detailed textures in both cases. Factorized representations reach their capacity limits in such scenes. PPNG-1, with only 128 KB parameters, fails to recreate fine details in both the central and background regions, and PPNG-2 also cannot recreate details in the background regions due to capacity with limited volume size.}
    \label{fig:360}
\end{figure*}
Our second quantitative evaluation comprehensively compares the most representative and state-of-the-art novel view synthesis models across various datasets. 
We evaluate the rendering quality, speed, training time, and model size. 
We broadly categorize these models into two groups: mobile-friendly and non-friendly approaches, differentiated based on their GL compatibility.
Table~\ref{tab:quantitative} presents the results. 
Our general observations are: 1) {\it Implicit approaches} offer the best quality and relatively compact size, but they are not mobile compatible, and their rendering speed tends to be slow; 2) {\it Real-time, mobile-compatible methods} tend to sacrifice some rendering quality for speed and typically have a larger model size; 3) {\it Our approaches} achieve the best trade-off in terms of model size, rendering quality, and training time. 
In particular, our model size significantly outperforms all competing baselines in model size and has one of the fastest training speeds. 

To further provide a comparison for low-bit model sizes, we optimized the open source SotA low-bit model, WaveletNeRF~\cite{Rho2022MaskedWR} to a 200KB size to our best effort with an experimental combination (i.e., $4\times$ smaller feature length, $2\times$ smaller feature grid, and a larger ($1e-8$) mask loss).
PPNG outperforms optimized WaveletNeRF by a large margin at the Internet-friendly size, demonstrating the advanced compression capacity of periodic encoding over sparsity encoding.
The rendering quality of PPNG-1 and PPNG-2 is comparable to other real-time methods, while the quality of PPNG-3 is comparable to implicit approaches.
\begin{table}[htbp!]
\vspace{-3mm}
\caption{{\bf Quantitative evaluation on unbounded 360$^\circ$ scenes.} We let $^\ast$ to denote author measured time; $^\dagger$ to denote reported time on paper.
}
\vspace{-3mm}
\resizebox{\linewidth}{!}{
\begin{tabular}{lcccccc}
\toprule
&& PSNR & SSIM & LPIPS & Size & Training Time\\
\midrule
MobileNeRF~\cite{Chen2022MobileNeRFET}  && 22.0 & 0.470 & 0.470 & 347 MB & 21+ hours$^\dagger$ \\
BakedSDF~\cite{Yariv2023BakedSDFMN}     && 24.5 & 0.697 & 0.309 & 457 MB & 7+ hours$^\ast$ \\
MERF~\cite{Reiser2023MERFMR}            && 25.2 & 0.722 & 0.311 & 162 MB & 2 hours (8 GPUs)$^\ast$ \\
SMERF~\cite{Duckworth2023SMERFSM}       && 28.0 & 0.728 & 0.212 & 139 MB & 17+hours$^\dagger$ \\
\midrule
PPNG-1                                  && 20.2 & 0.476 & 0.658 & 512 KB & 20.7 min\\
PPNG-2                                  && 21.9 & 0.543 & 0.499 & 2.93 MB & 15.0 min \\
PPNG-3                                  && 23.7 & 0.618 & 0.392 & 33.3 MB & 7.8 min \\
\bottomrule
\end{tabular}
}
\vspace{-6mm}
\label{tab:360}
\end{table}

\noindent\textbf{Multi-platform analysis:} 
Table~\ref{tab:fps} shows that PPNG can be efficiently rendered with various devices including mobile devices. On supplementary material, we demonstrate that mobile phones can load multiple scenes at once. 

\subsection{Ablation studies}
We evaluate how each component of our model impacts the performance and report the results in Table~\ref{tab:ablation}.  

\noindent\textbf{Number of MLP layers:}
We demonstrate that adding one additional layer to the shallow MLP improves the PSNR by 0.25 dB in PPNG-1, yet does not alter the quality much for PPNG-2 and PPNG-3. 
While a deeper MLP is known to offer a stronger capacity for modeling complex appearances~\cite{Mller2022InstantNG}, they also increase the computation required per sample. Therefore, we use a one-layer MLP to ensure speed and compatibility in our final model. 

\noindent\textbf{Levels of quantization:}
A finer feature grid improves performance (with $Q=100$) but increases file and memory size, which grows rapidly at a rate of $O(Q^3)$ for PPNG-3. Since PPNG-1 and PPNG-2 are converted into PPNG-3 at rendering time, we consider $Q=80$ to be an appropriate level, effectively balancing quality and memory size.

\noindent\textbf{Frequency Ranges:}
Choosing the frequencies for sinusoidal positional encoding is crucial. We show that using too low or high frequency can degrade performance. Additionally, optimal frequency may vary based on levels of quantization and scene scale. 

\noindent\textbf{Number of Components:}
The number of factorized components is crucial for balancing model size and performance in PPNG-1 and PPNG-2. Fewer components reduce the model size to under 100KB, while more components improve rendering quality.

\subsection{Limitations}
Plenoptic PNG is designed for scenes with a limited range. Although it can model unbounded scenes with reasonable quality at a tiny size (as shown in Figure~\ref{fig:360}), it may not perform as effectively as large real-time models (see Table~\ref{tab:360}). In the future, we plan to extend PPNG to include contracted space modeling and blocks to address this limitation.  


\section{Conclusion}
We present Plenoptic PNG (PPNG), a highly compact representation for real-time, web-compatible free-viewpoint rendering. 
PPNG leverages Fourier feature modeling and volume factorization to achieve a small model size and fast training time. 
Compared to other real-time NeRF models, PPNG offers the smallest size and quickest training with minimal quality loss. 
Additionally, PPNG can be efficiently loaded on lightweight devices like mobile phones using GL-texturing. We believe PPNG will enable new applications requiring easy sharing of 3D immersive visual content.

\clearpage
\noindent\textbf{Acknowledgements}
We thank David Forsyth and Thomas Müller for insightful feedback for the paper.
This work is supported by NSF grants 2331878, 2340254, 2312102 and 2020227.

{
    \small
    \bibliographystyle{splncs04}
    \bibliography{references}
}

\newpage
\appendix
\section{Training Details}

We follow default parameters of Instant-NGP~\cite{Mller2022InstantNG} for training. Specifically we minimize the Huber Loss with 50,000 steps using Adam~\cite{Kingma2014AdamAM} optimizer. 
During training our voxel-based density grid cache has resolution of 128. For object level scenes, we use single scale voxel grid. 
For unbounded scenes, we use voxel grids with five scales, where higher scale covers half of each dimension (i.e, 0.5 width, 0.5 height and 0.5 depth) with the same resolution centered at (0.5, 0.5, 0.5).

\section{Additional Qualitative Results}
We provide additional qualitative results for all the datasets.

\section{Additional Quantiative Results}
\noindent\textbf{Comparisons with real-time methods}.~We provide a quantitative comparison with real-time Web-compatible methods in Table~\ref{tab:comparison}. This table shows that our approach benefits from a small model size, fast training speed, minimal VRAM requirements, and real-time rendering capabilities, while maintaining comparable rendering quality.

\noindent\textbf{Detailed quantitative performance per scene}.~We show detailed quantitative results of our PPNG model across the NeRF Synthetics, NSVF Synthetic, BlendedMVS, and Tanks and Temples datasets from Table 7. to Table 18.

\begin{table}[h]
\caption{Comparison with the real-time WebGL rendered models. We use values provided from~\cite{Rojas2023ReReNDRR} for SNeRG~\cite{Hedman2021BakingNR}, MobileNeRF~\cite{Chen2022MobileNeRFET}, Re-Rend~\cite{Rojas2023ReReNDRR}, and author provided values for PlenOctree~\cite{Yu2021PlenOctreesFR} }
\resizebox{\linewidth}{!}
{
\begin{tabular}{lccccc}
\toprule
Model Name &&  PSNR & Size & GPU Usage & Training Time \\
\midrule
SNeRG~\cite{Hedman2021BakingNR}         && 30.4 & 87 MB & 3627 MB & 15 hrs \\
MobileNeRF~\cite{Chen2022MobileNeRFET}  && 30.9 & 126 MB & 570 MB & 20 hrs \\
PlenOctree~\cite{Yu2021PlenOctreesFR}   && 31.7 & 1976 MB & 1690 MB      & 50 hrs \\
Re-Rend~\cite{Rojas2023ReReNDRR}        && 29.0 & 199 MB & 532 MB & 60 hrs \\
\midrule
PPNG-1                                  && 28.8 & 151 KB & 47 MB & 13.1 min \\
PPNG-2                                  && 31.0 & 2.49 MB & 47 MB & 9.8 min \\
PPNG-3                                  && 31.5 & 32.8 MB & 47 MB & 4.9 min \\
\bottomrule
\end{tabular}
}
\label{tab:comparison}
\end{table}

\newpage
\begin{table}[h]
\caption{PSNR evaluation for for Tanks and Temples dataset.  }
\centering
\resizebox{\linewidth}{!}
{
\begin{tabular}{lccccc}
\toprule
&  Barn & Caterpillar & Family & Ignatius & Truck\\
\midrule
PPNG-1 & 24.52 & 22.78 & 29.97 & 26.71 & 24.42\\
PPNG-2 & 25.71 & 24.32 & 32.50 & 27.16 & 26.46\\
PPNG-3 & 26.42 & 24.88 & 33.28 & 27.51 & 27.05\\
\bottomrule
\end{tabular}
}
\label{tab:appendix_tnt_psnr}
\end{table}

\begin{table}[h]
\caption{SSIM evaluation for for Tanks and Temples dataset.  }
\centering
\resizebox{\linewidth}{!}
{
\begin{tabular}{lccccc}
\toprule
&  Barn & Caterpillar & Family & Ignatius & Truck\\
\midrule
PPNG-1 & 0.827 & 0.882 & 0.937 & 0.937 & 0.881\\
PPNG-2 & 0.848 & 0.900 & 0.958 & 0.943 & 0.910\\
PPNG-3 & 0.869 & 0.915 & 0.968 & 0.951 & 0.923\\
\bottomrule
\end{tabular}
}
\label{tab:appendix_tnt_ssim}
\end{table}

\begin{table}[h]
\caption{LPIPS~(AlexNet) evaluation for for Tanks and Temples dataset.  }
\centering
\resizebox{\linewidth}{!}
{
\begin{tabular}{lccccc}
\toprule
&  Barn & Caterpillar & Family & Ignatius & Truck\\
\midrule
PPNG-1 & 0.327 & 0.200 & 0.085 & 0.086 & 0.197\\
PPNG-2 & 0.274 & 0.152 & 0.046 & 0.077 & 0.132\\
PPNG-3 & 0.208 & 0.131 & 0.036 & 0.072 & 0.112\\
\bottomrule
\end{tabular}
}
\label{tab:appendix_tnt_lpips}
\end{table}

\begin{table}[h]
\caption{PSNR evaluation for for BlendedMVS dataset.  }
\centering
{
\begin{tabular}{lcccc}
\toprule
& Character & Fountain & Jade & Statues\\
\midrule
PPNG-1 & 25.4 & 24.14 & 24.87 & 24.67 \\
PPNG-2 & 28.88 & 26.07 & 25.14 & 26.04 \\
PPNG-3 & 29.65 & 26.52 & 24.91 & 26.48 \\
\bottomrule
\end{tabular}
}
\label{tab:appendix_blendedmvs_psnr}
\end{table}

\begin{table}[h]
\caption{SSIM evaluation for for BlendedMVS dataset.  }
\centering
{
\begin{tabular}{lcccc}
\toprule
& Character & Fountain & Jade & Statues\\
\midrule
PPNG-1 & 0.911 & 0.823 & 0.849 & 0.838 \\
PPNG-2 & 0.956 & 0.881 & 0.861 & 0.879 \\
PPNG-3 & 0.965 & 0.901 & 0.870 & 0.900 \\
\bottomrule
\end{tabular}
}
\label{tab:appendix_blendedmvs_ssim}
\end{table}

\begin{table}[h]
\caption{LPIPS~(AlexNet) evaluation for for BlendedMVS dataset.  }
\centering
{
\begin{tabular}{lcccc}
\toprule
& Character & Fountain & Jade & Statues\\
\midrule
PPNG-1 & 0.076 & 0.18 & 0.128 & 0.152 \\
PPNG-2 & 0.029 & 0.095 & 0.102 & 0.092 \\
PPNG-3 & 0.023 & 0.078 & 0.096 & 0.073 \\
\bottomrule
\end{tabular}
}
\label{tab:appendix_blendedmvs_lpips}
\end{table}

\newpage

\begin{table*}[h]
\caption{PSNR evaluation for NeRF synthetic dataset.  }
\centering
{
\begin{tabular}{lcccccccc}
\toprule
& chair & drums & ficus & hotdog & lego & materials & mic & ship \\
\midrule
PPNG-1 & 29.69 & 23.14 & 28.2 & 33.82 & 30.39 & 26.89 & 32.08 & 26.9\\
PPNG-2 & 32.5 & 25.07 & 31.04 & 35.22 & 33.37 & 27.5 & 34.11 & 29.13 \\
PPNG-3 & 33.54 & 25.41 & 31.6 & 36.17 & 34.54 & 28.53 & 35.18 & 30.27\\
\bottomrule
\end{tabular}
}
\label{tab:appendix_nerf_psnr}
\end{table*}

\begin{table*}[h]
\caption{SSIM evaluation for for NeRF synthetic dataset.  }
\centering
{
\begin{tabular}{lcccccccc}
\toprule
& chair & drums & ficus & hotdog & lego & materials & mic & ship \\
\midrule
PPNG-1 & 0.94 & 0.904 & 0.937 & 0.966 & 0.937 & 0.918 & 0.969 & 0.84\\
PPNG-2 & 0.968 & 0.919 & 0.96 & 0.975 & 0.967 & 0.915 & 0.978 & 0.869\\
PPNG-3 & 0.973 & 0.915 & 0.964 & 0.978 & 0.974 & 0.926 & 0.979 & 0.884\\
\bottomrule
\end{tabular}
}
\label{tab:appendix_nerf_ssim}
\end{table*}

\begin{table*}[h]
\caption{LPIPS~(AlexNet) evaluation for for NeRF synthetic dataset.  }
\centering
{
\begin{tabular}{lcccccccc}
\toprule
& chair & drums & ficus & hotdog & lego & materials & mic & ship \\
\midrule
PPNG-1  & 0.056 & 0.116 & 0.046 & 0.041 & 0.04 & 0.09 & 0.044 & 0.207 \\
PPNG-2  & 0.022 & 0.077 & 0.034 & 0.028 & 0.018 & 0.094 & 0.027 & 0.135 \\
PPNG-3 & 0.015 & 0.075 & 0.031 & 0.022 & 0.014 & 0.072 & 0.021 & 0.1\\
\bottomrule
\end{tabular}
}
\label{tab:appendix_nerf_lpips}
\end{table*}

\begin{table*}[h]
\caption{PSNR evaluation for Synthetic NSVF dataset.  }
\centering
{
\begin{tabular}{lcccccccc}
\toprule
& Bike & Lifestyle & Palace & Robot & Spaceship & Steamtrain & Toad & Wineholder \\
\midrule
PPNG-1 & 26.42 & 28.07 & 31.85 & 30.87 & 29.89 & 30.89 & 25.99 & 26.42 \\
PPNG-2 & 34.71 & 30.74 & 34.42 & 34.61 & 30.54 & 33.21 & 32.65 & 28.83 \\
PPNG-3 & 35.64 & 31.43 & 36.27 & 35.5 & 31.0 & 33.92 & 34.15 & 29.7 \\
\bottomrule
\end{tabular}
}
\label{tab:appendix_nsvf_psnr}
\end{table*}

\begin{table*}[h]
\caption{SSIM evaluation for for Synthetic NSVF dataset.  }
\centering
{
\begin{tabular}{lcccccccc}
\toprule
& Bike & Lifestyle & Palace & Robot & Spaceship & Steamtrain & Toad & Wineholder \\
\midrule
PPNG-1 & 0.955 & 0.927 & 0.934 & 0.973 & 0.966 & 0.971 & 0.89 & 0.927 \\
PPNG-2 & 0.984 & 0.946 & 0.962 & 0.987 & 0.966 & 0.98 & 0.966 & 0.951 \\
PPNG-3 & 0.987 & 0.954 & 0.975 & 0.989 & 0.967 & 0.984 & 0.977 & 0.96 \\
\bottomrule
\end{tabular}
}
\label{tab:appendix_nsvf_ssim}
\end{table*}

\begin{table*}[h]
\caption{LPIPS~(AlexNet) evaluation for for Synthetic NSVF dataset.  }
\centering
{
\begin{tabular}{lcccccccc}
\toprule
& Bike & Lifestyle & Palace & Robot & Spaceship & Steamtrain & Toad & Wineholder \\
\midrule
PPNG-1  & 0.028 & 0.072 & 0.044 & 0.023 & 0.037 & 0.027 & 0.09 & 0.063 \\
PPNG-2  & 0.008 & 0.046 & 0.02 & 0.009 & 0.038 & 0.017 & 0.025 & 0.035 \\
PPNG-3 & 0.006 & 0.038 & 0.012 & 0.008 & 0.037 & 0.015 & 0.015 & 0.027 \\
\bottomrule
\end{tabular}
}
\label{tab:appendix_nsvf_lpips}
\end{table*}

\clearpage 

\begin{figure*}[t]
    \centering
    \begin{tabular}{c@{}c@{}c}
        \includegraphics[angle=270,origin=c,width=0.23\textwidth]{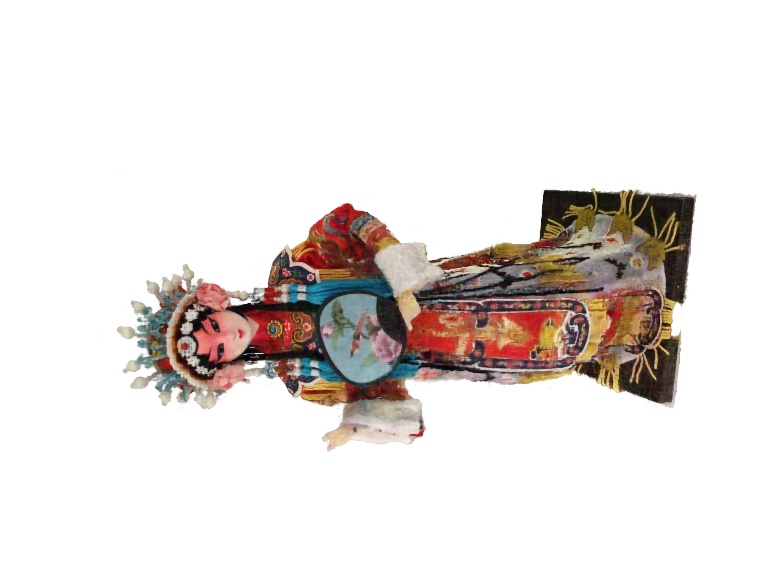} &
        \includegraphics[angle=270,origin=c,width=0.23\textwidth]{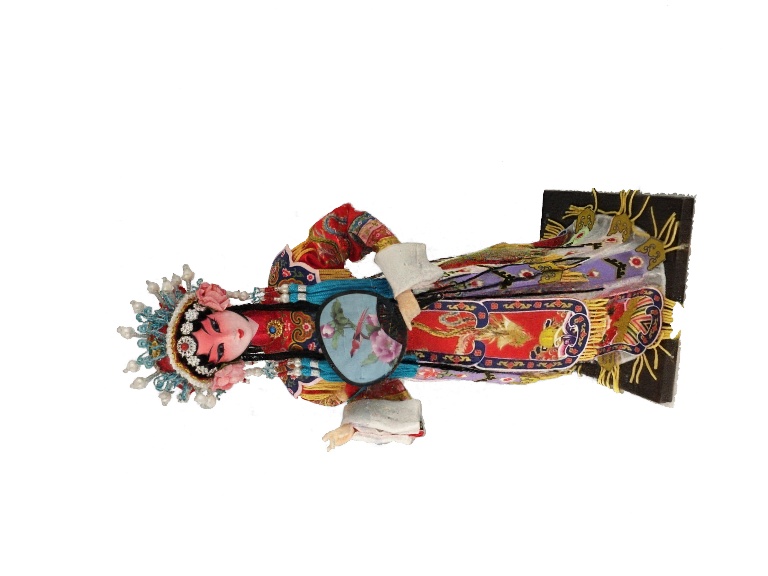} &
        \includegraphics[angle=270,origin=c,width=0.23\textwidth]{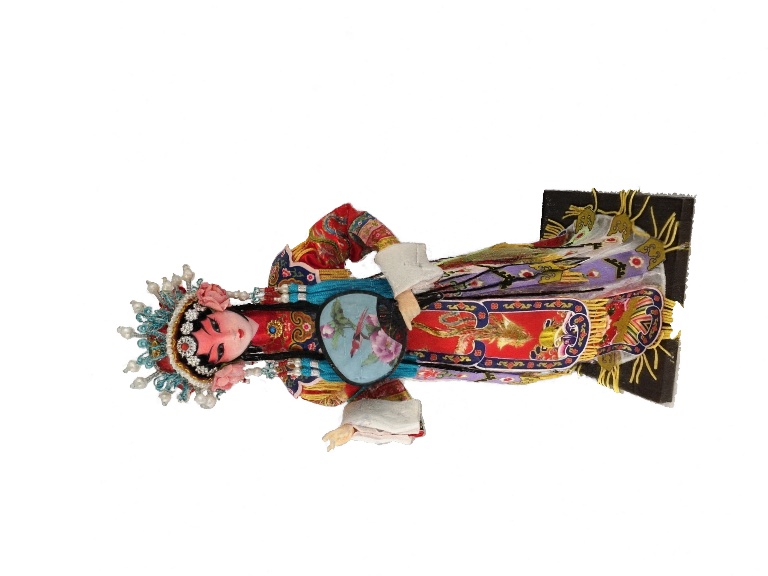}\\
        \includegraphics[width=0.23\textwidth]{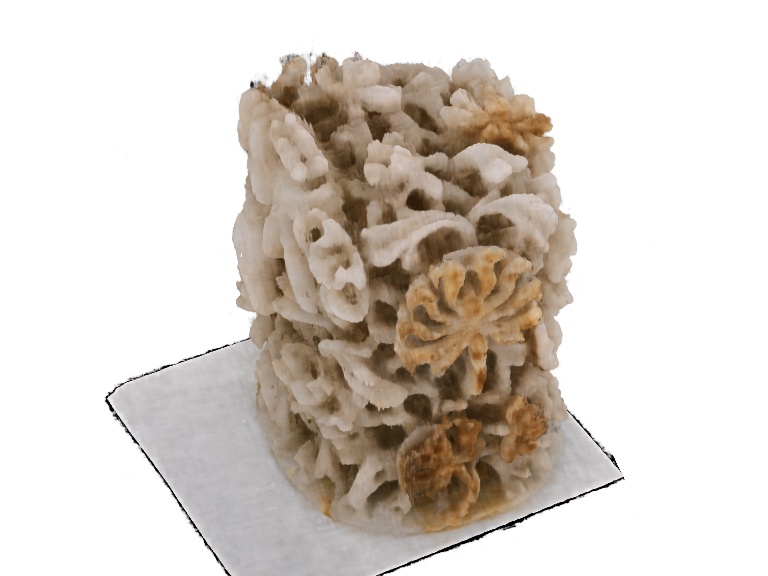} &
        \includegraphics[width=0.23\textwidth]{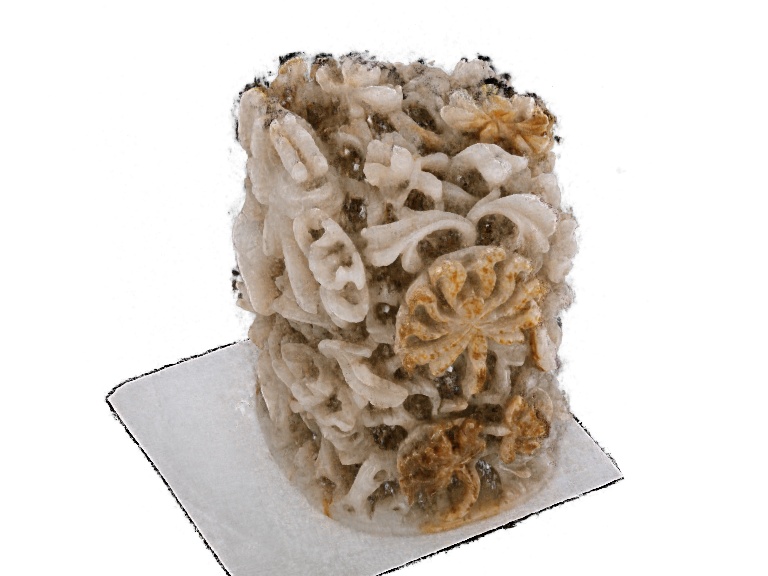} &
        \includegraphics[width=0.23\textwidth]{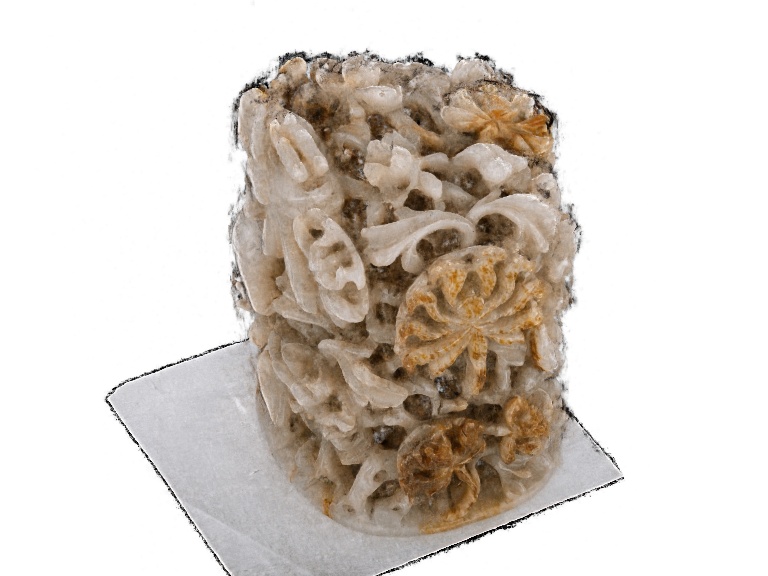}\\
        \includegraphics[angle=90,origin=c,width=0.23\textwidth]{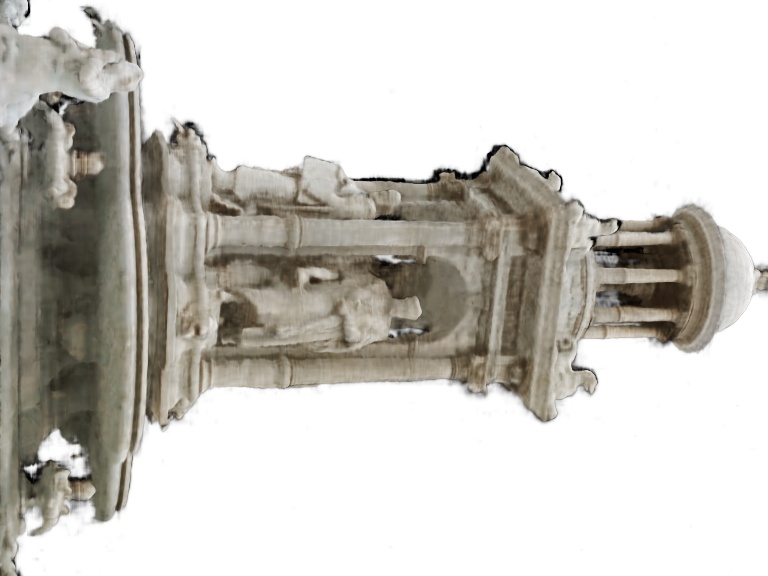} &
        \includegraphics[angle=90,origin=c,width=0.23\textwidth]{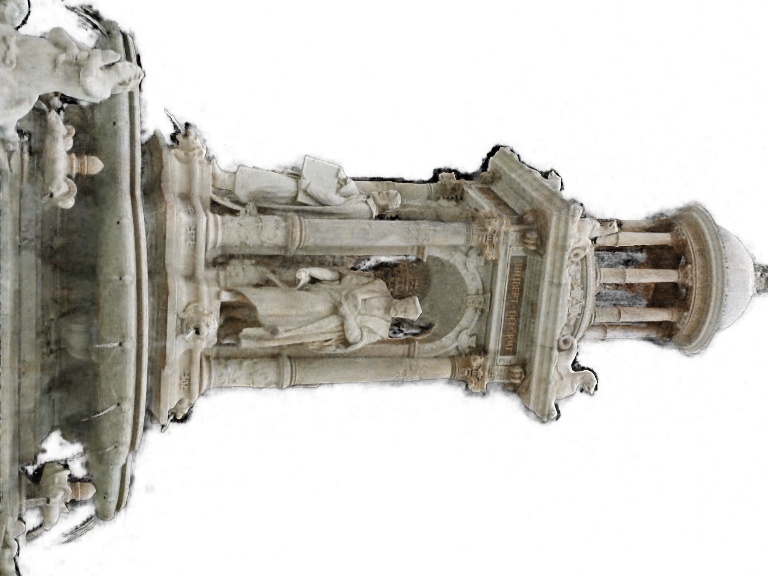} &
        \includegraphics[angle=90,origin=c,width=0.23\textwidth]{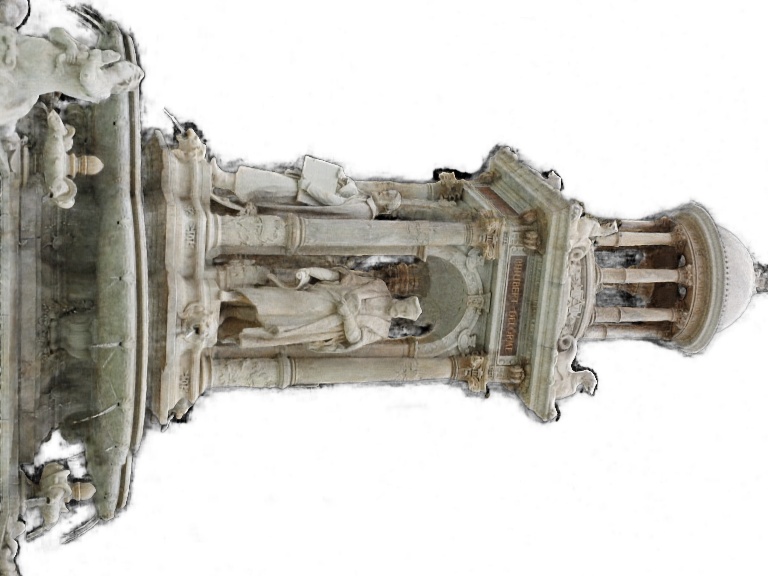}\\
        \includegraphics[angle=270,origin=c,width=0.23\textwidth]{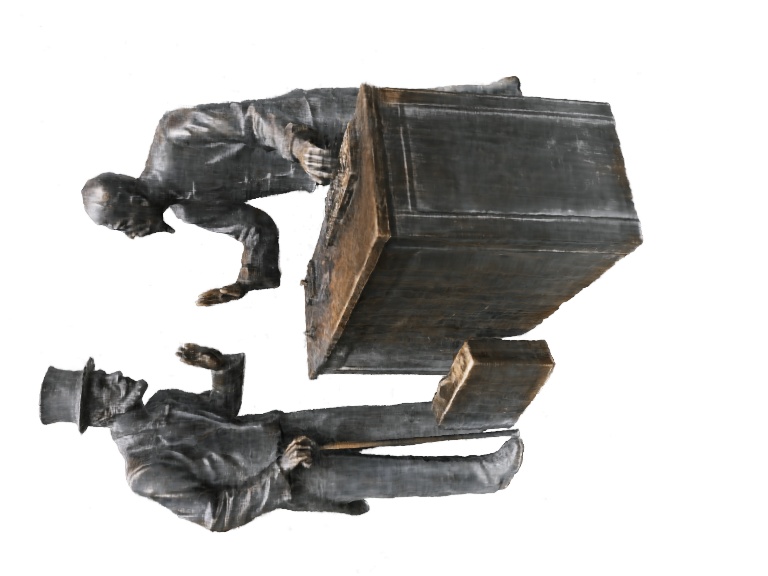} &
        \includegraphics[angle=270,origin=c,width=0.23\textwidth]{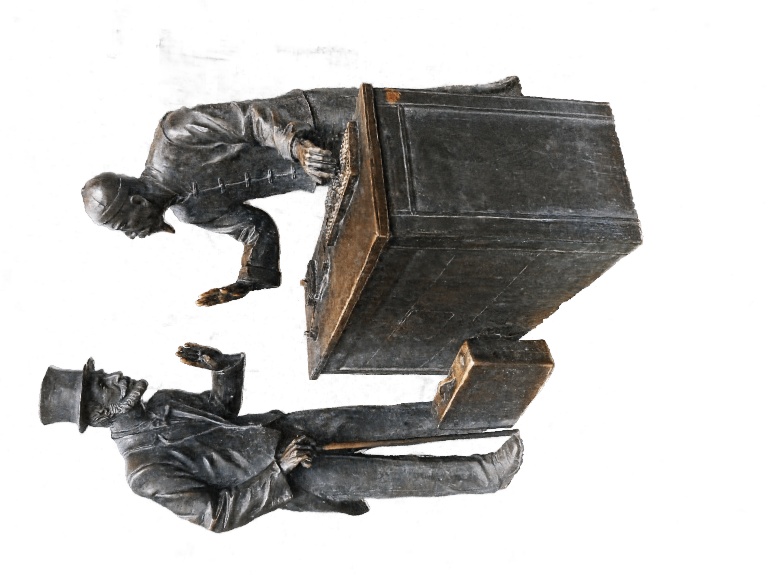} &
        \includegraphics[angle=270,origin=c,width=0.23\textwidth]{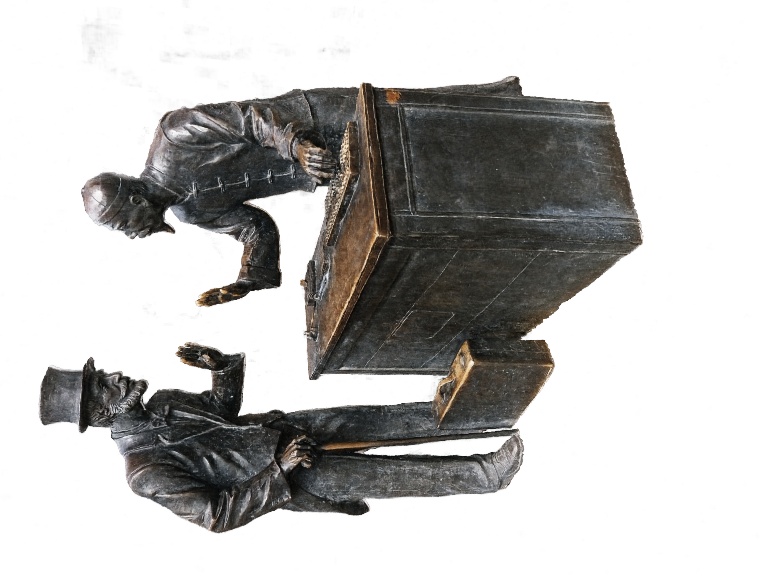}\\
         PPNG-1 & PPNG-2 & PPNG-3
    \end{tabular}
    \caption{Qualitative results for Blended MVS dataset.}
    \label{fig:appendix_blended}
\end{figure*}
\begin{figure*}[t]
    \centering
    \begin{tabular}{c@{}c@{}c@{}c@{}c@{}c}
        \includegraphics[width=0.15\textwidth]{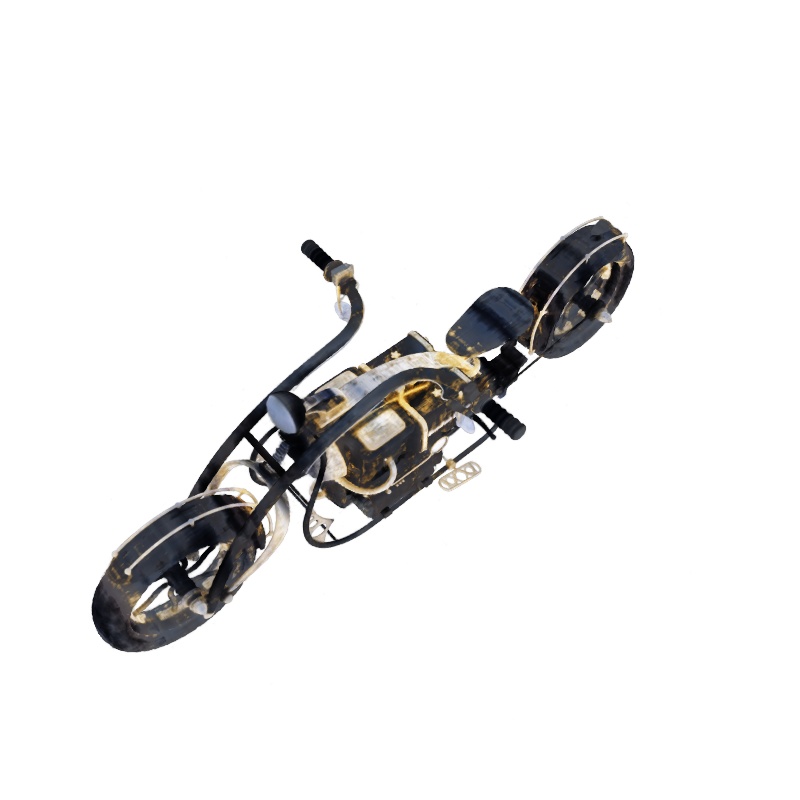} &
        \includegraphics[width=0.15\textwidth]{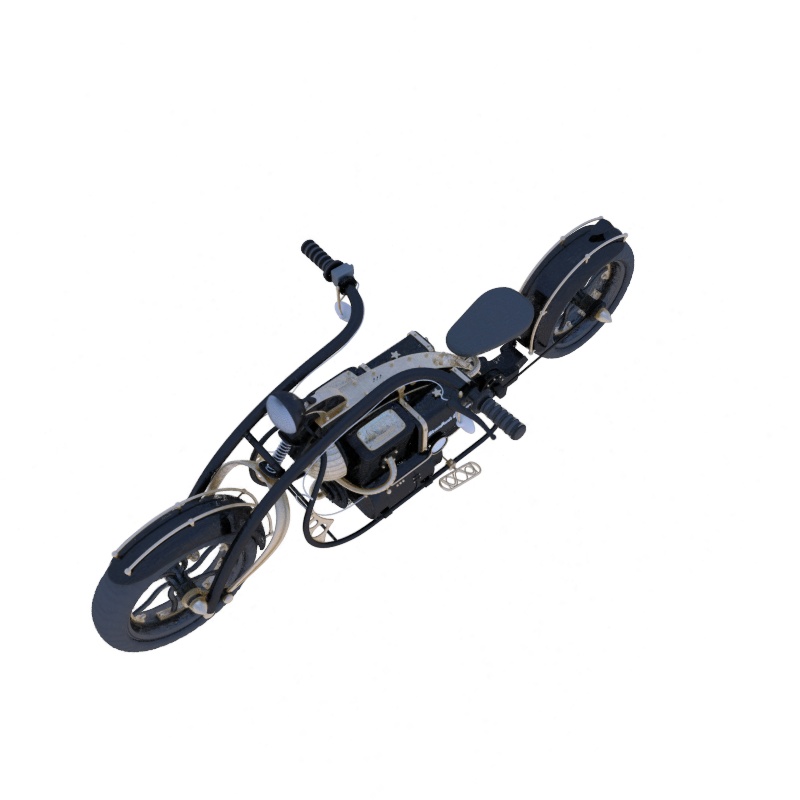} &
        \includegraphics[width=0.15\textwidth]{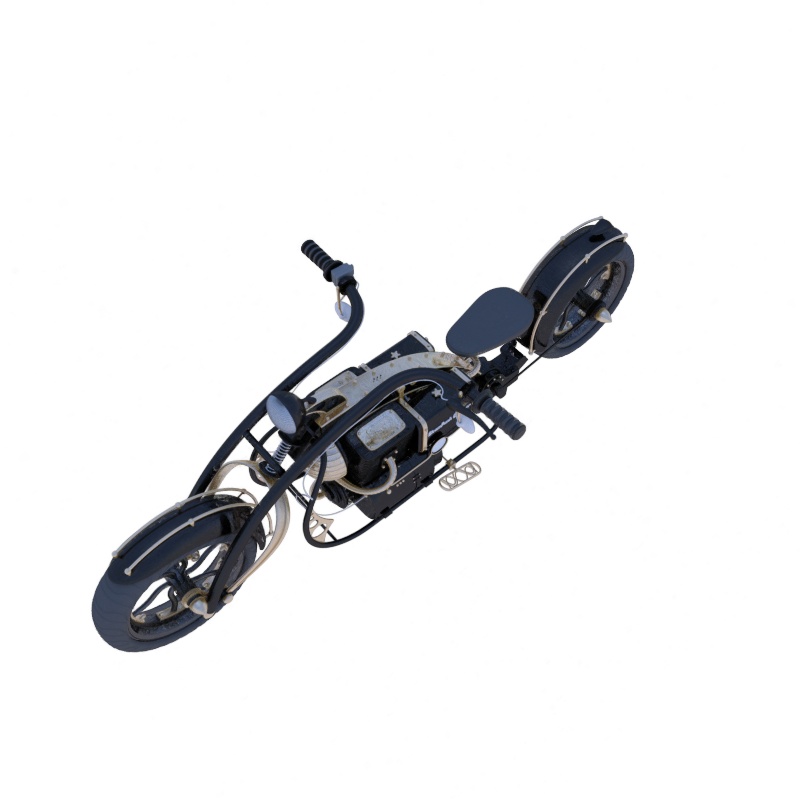} &
        \includegraphics[width=0.15\textwidth]{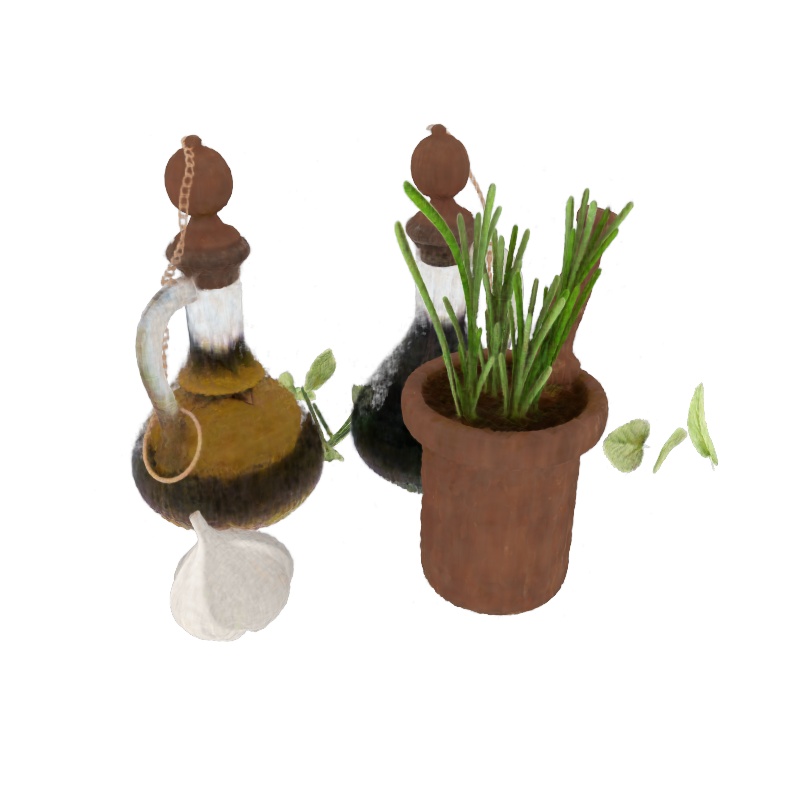} &
        \includegraphics[width=0.15\textwidth]{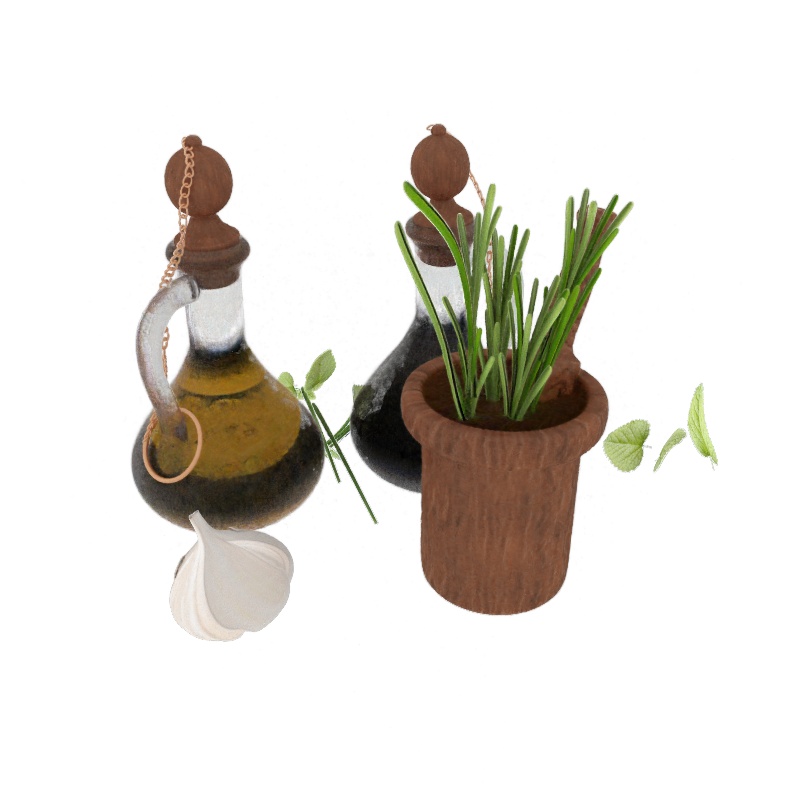} &
        \includegraphics[width=0.15\textwidth]{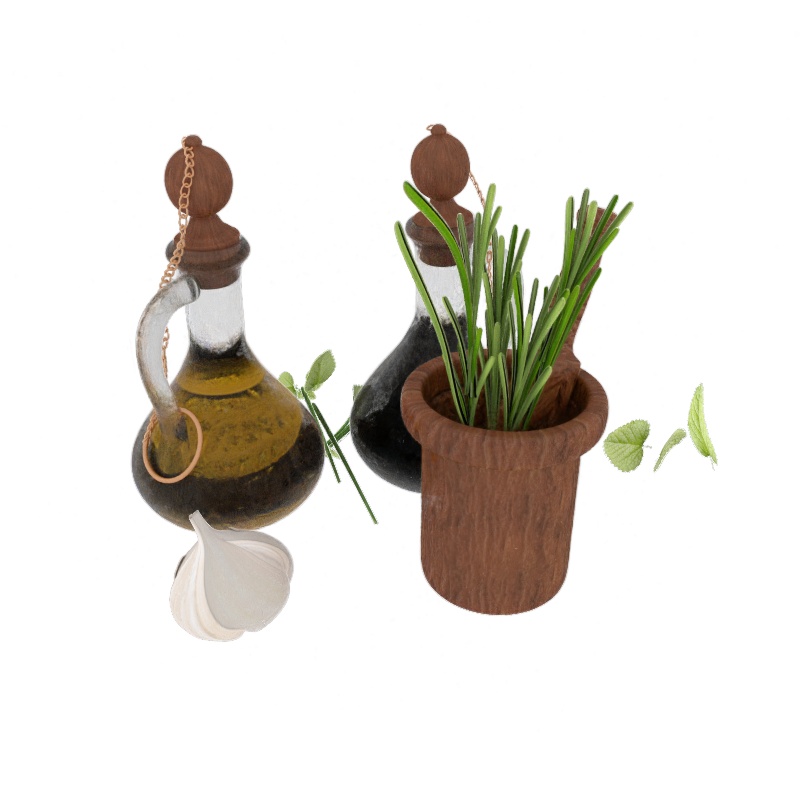}\\
        \includegraphics[width=0.15\textwidth]{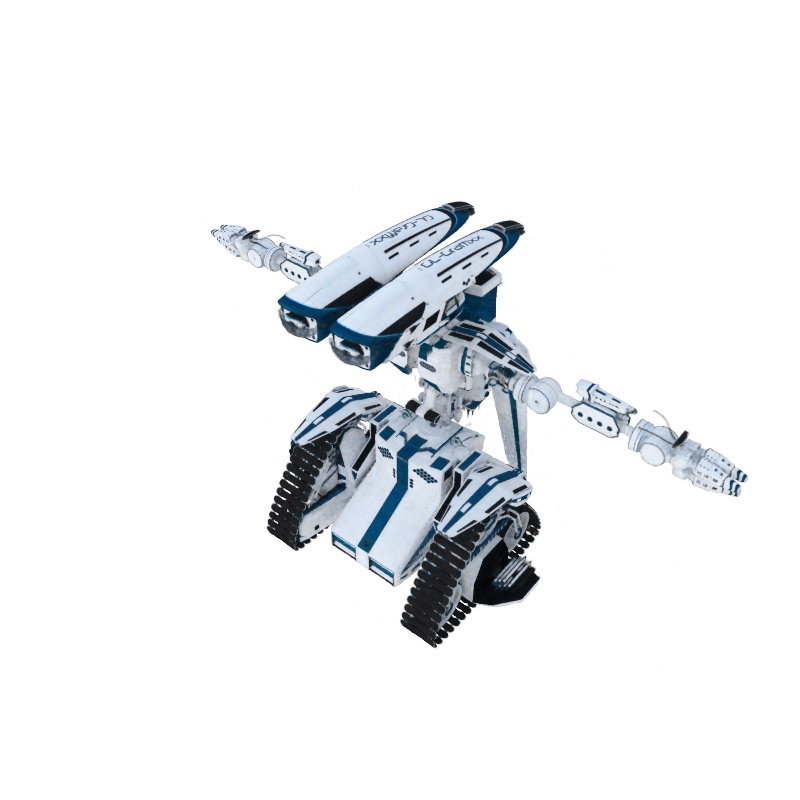} &
        \includegraphics[width=0.15\textwidth]{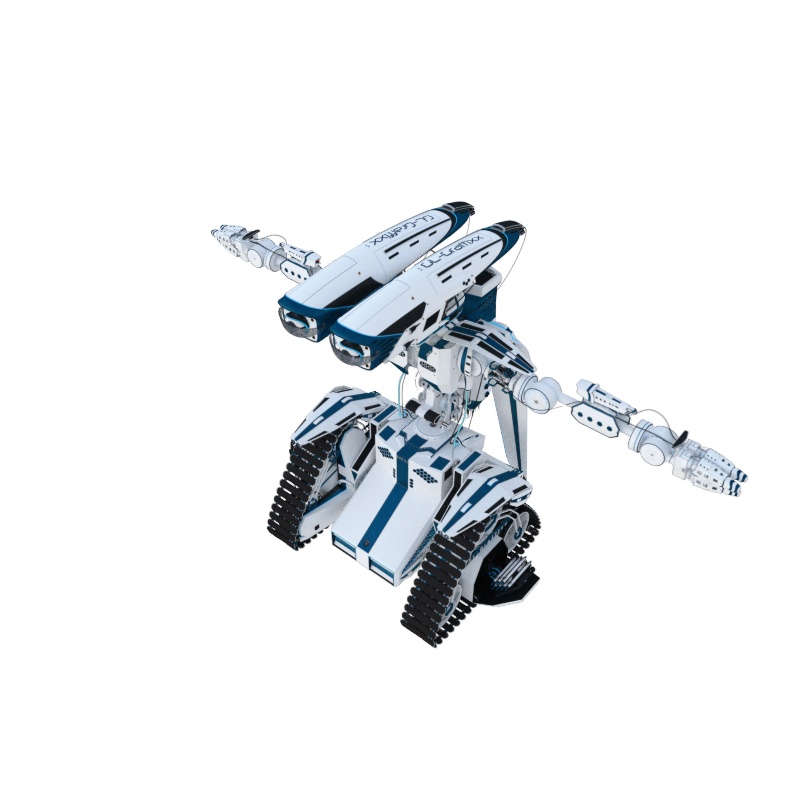} &
        \includegraphics[width=0.15\textwidth]{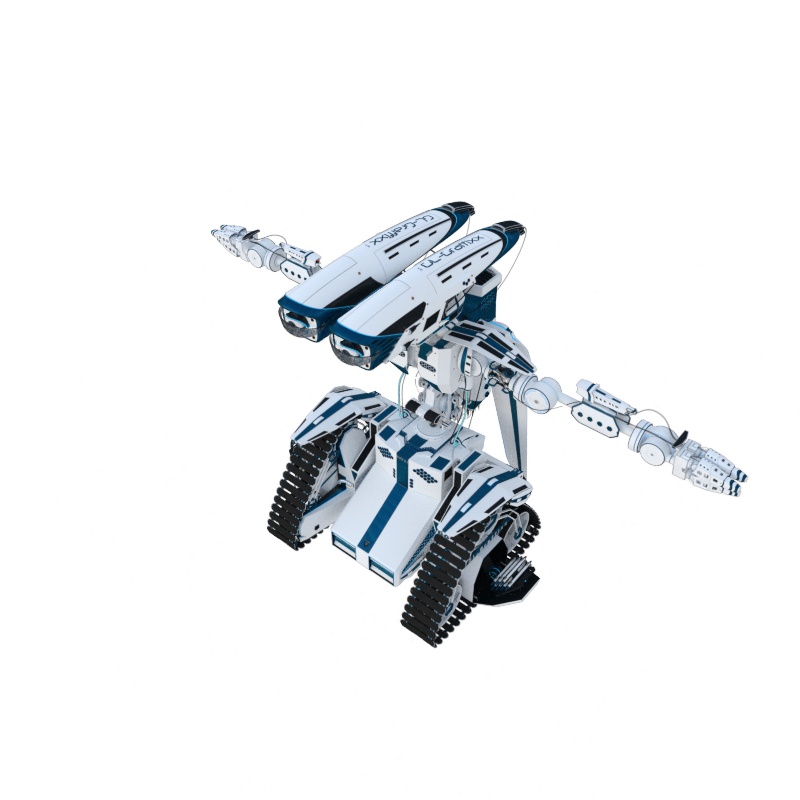} &
        \includegraphics[width=0.15\textwidth]{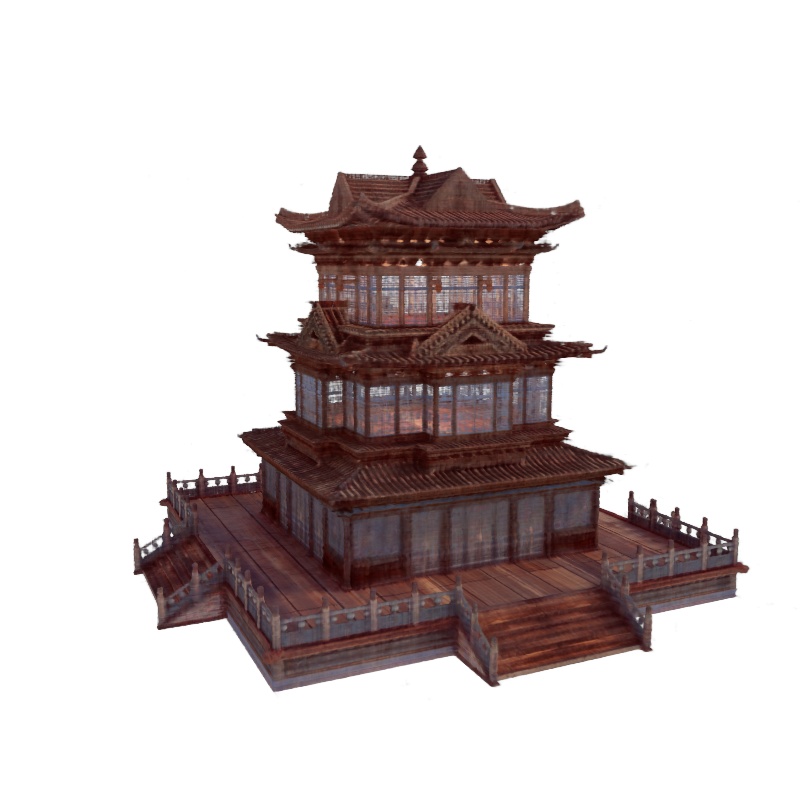} &
        \includegraphics[width=0.15\textwidth]{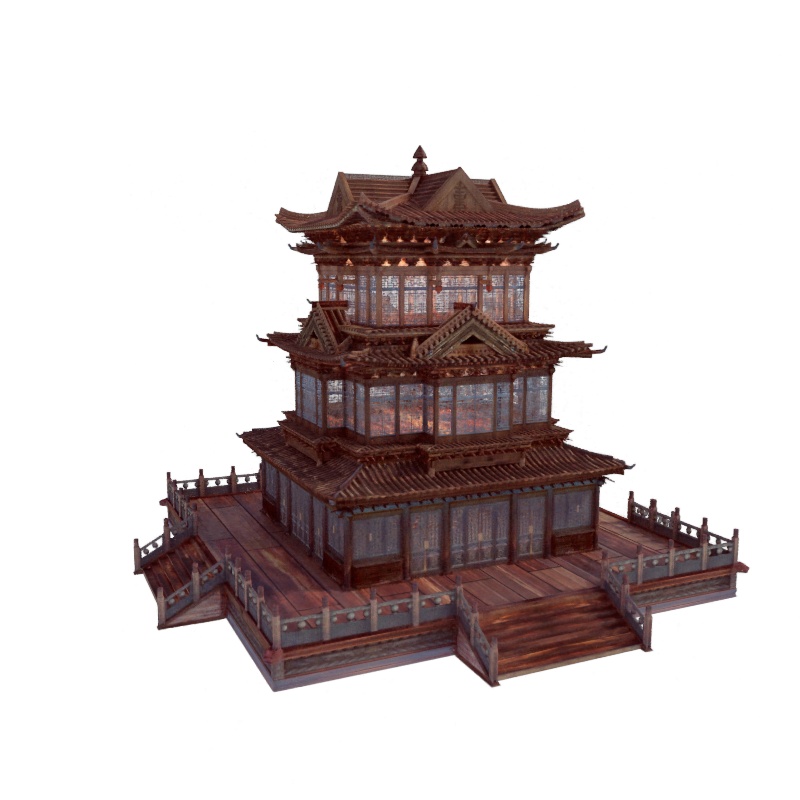} &
        \includegraphics[width=0.15\textwidth]{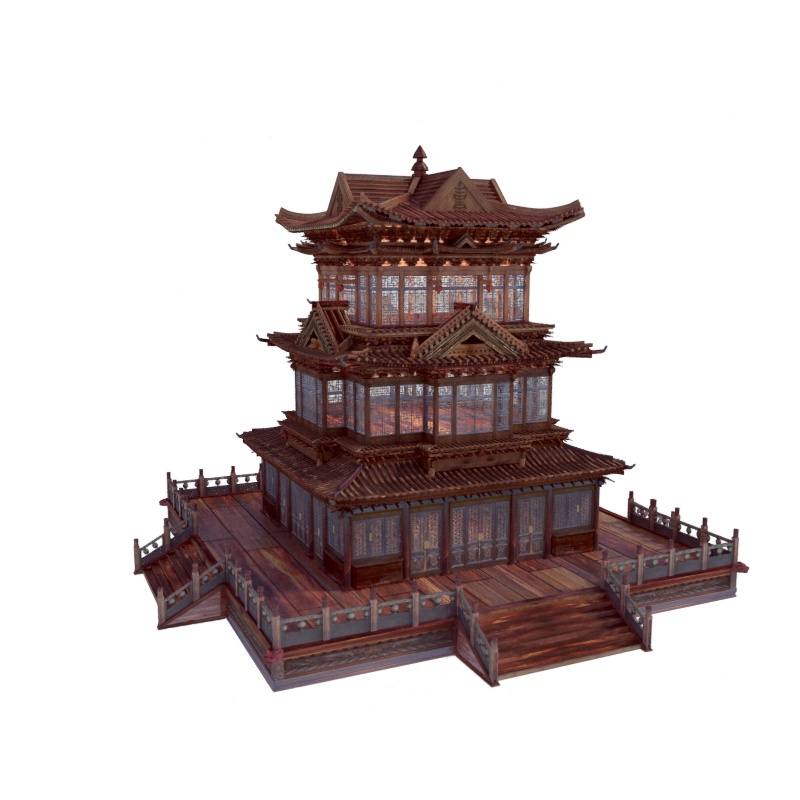}\\
        \includegraphics[width=0.15\textwidth]{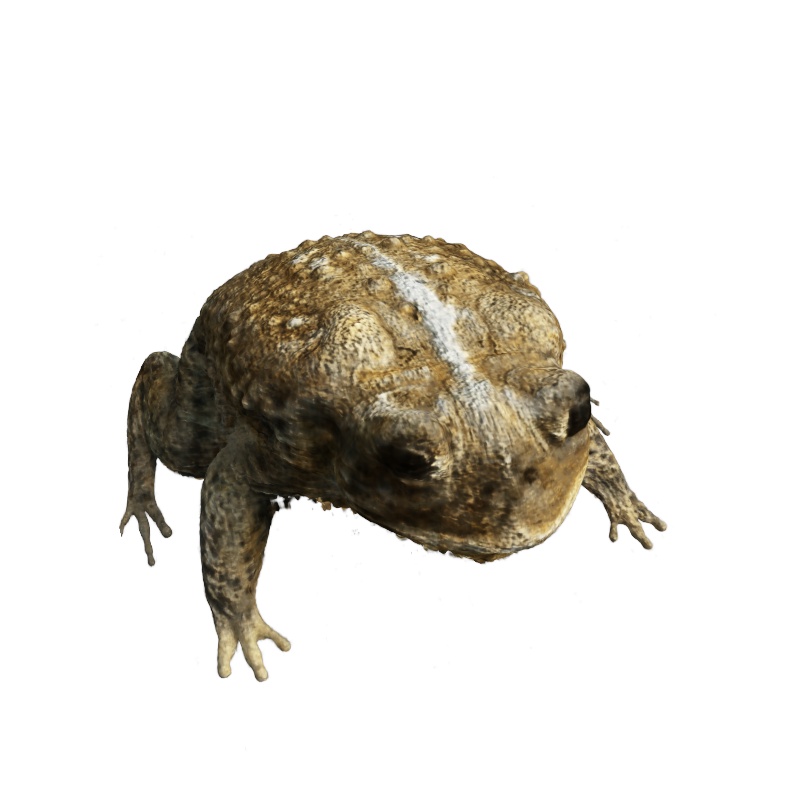} &
        \includegraphics[width=0.15\textwidth]{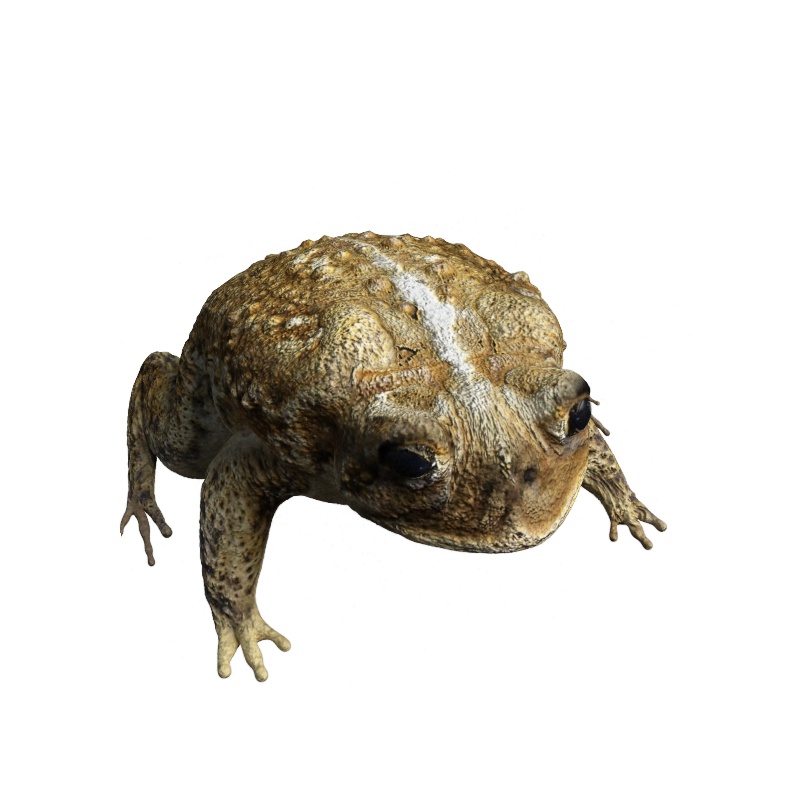} &
        \includegraphics[width=0.15\textwidth]{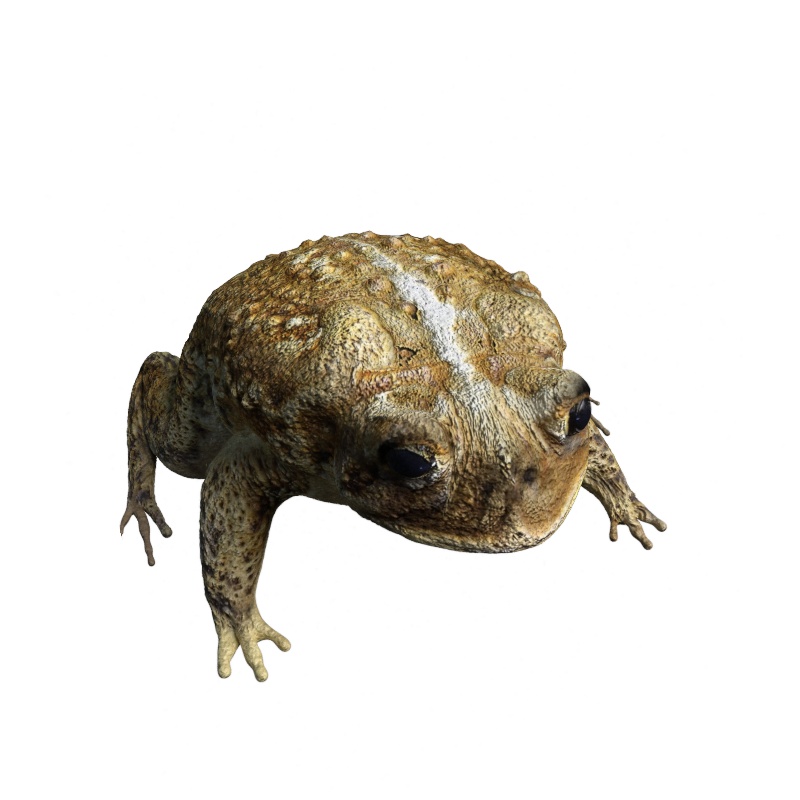} &
        \includegraphics[width=0.15\textwidth]{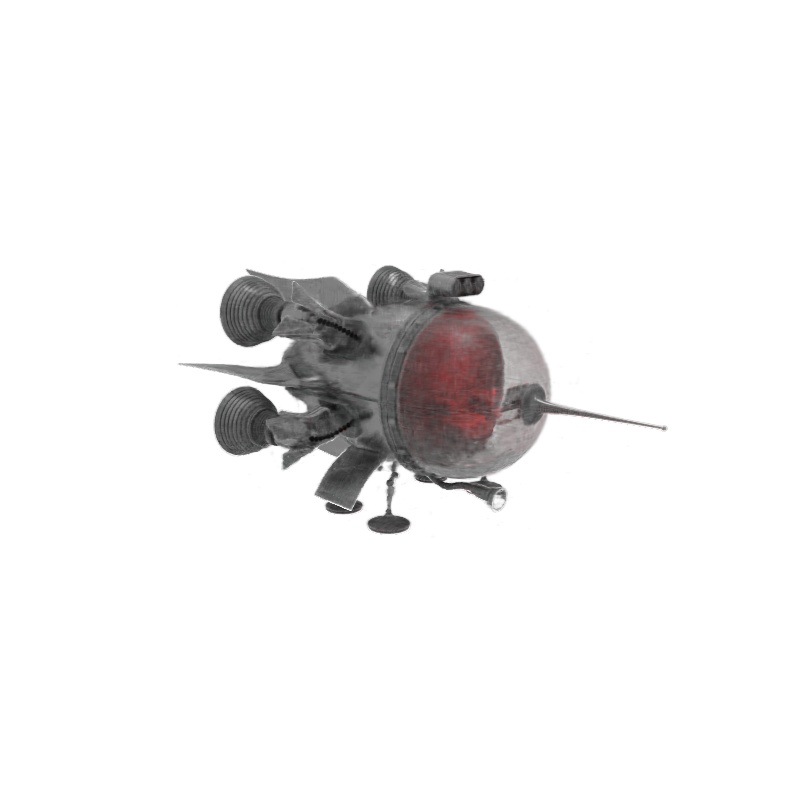} &
        \includegraphics[width=0.15\textwidth]{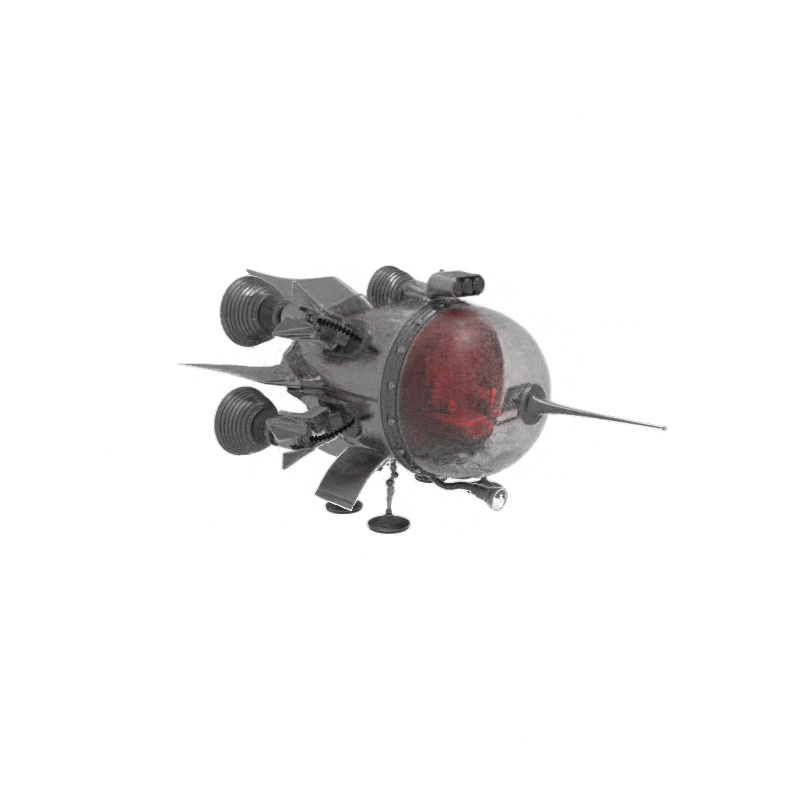} &
        \includegraphics[width=0.15\textwidth]{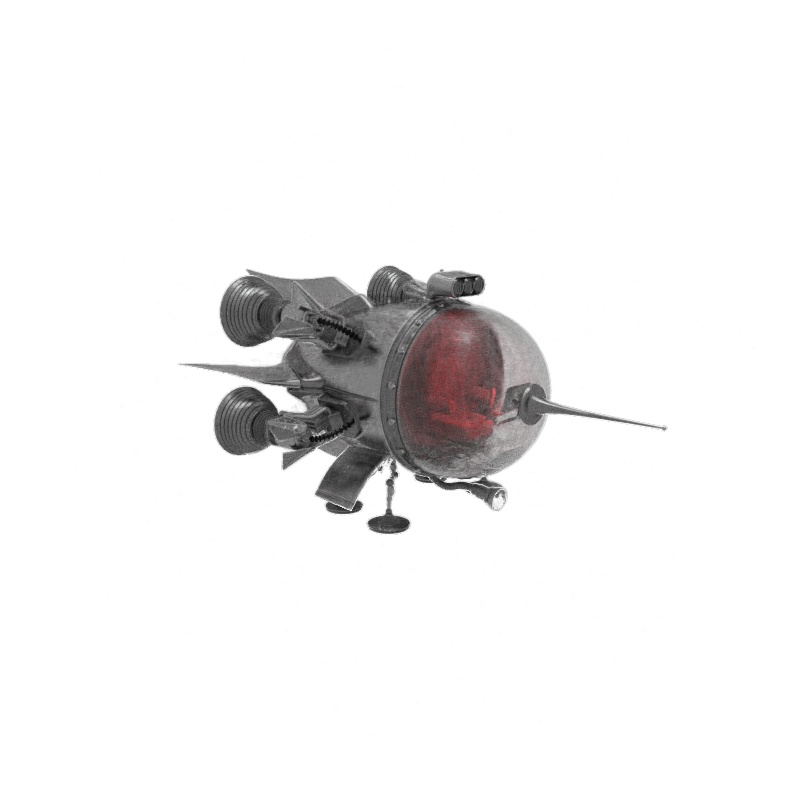}\\
        \includegraphics[width=0.15\textwidth]{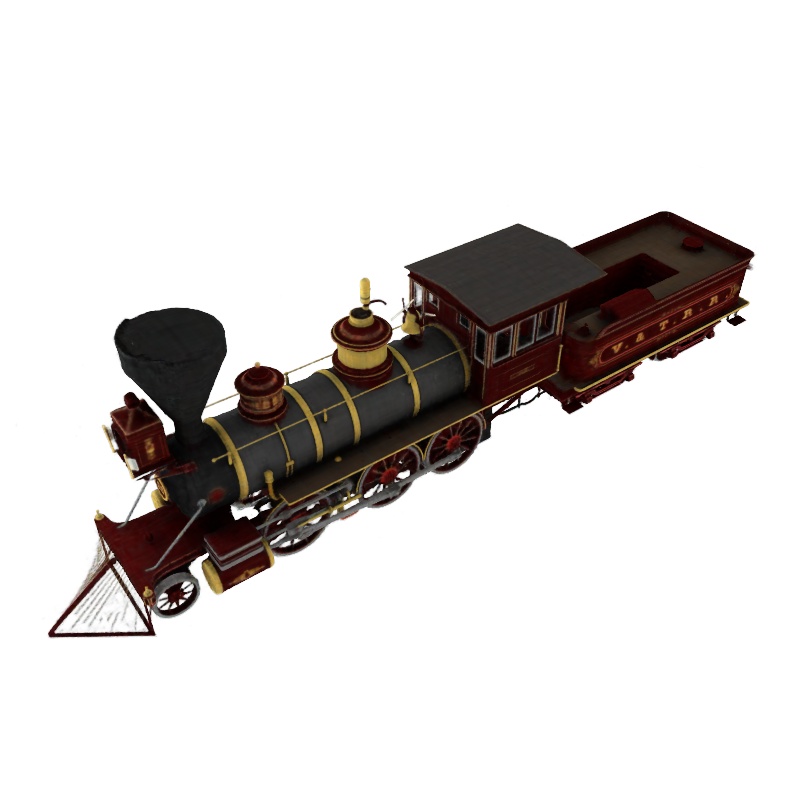} &
        \includegraphics[width=0.15\textwidth]{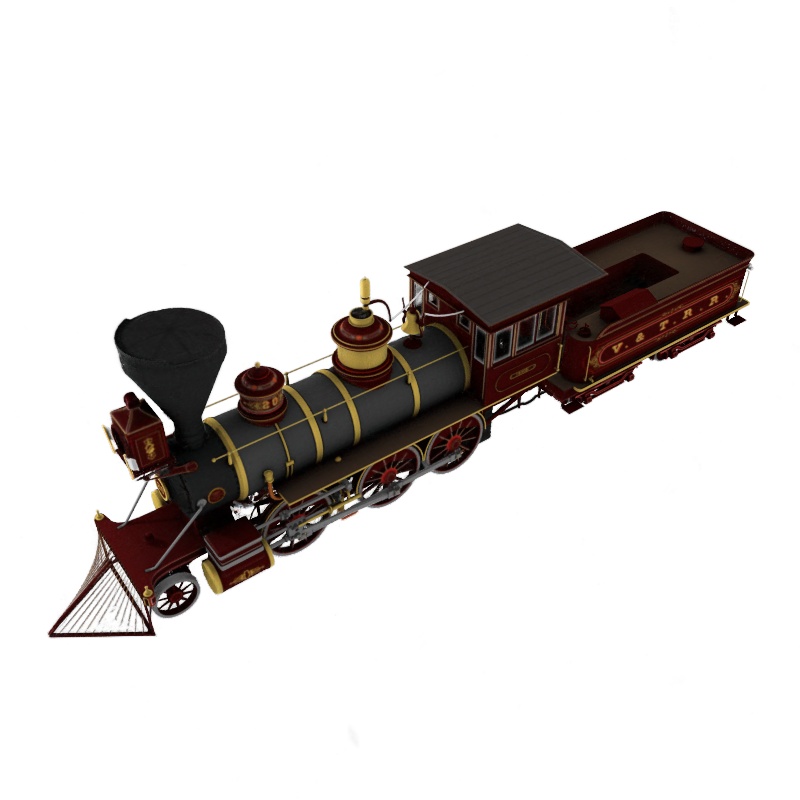} &
        \includegraphics[width=0.15\textwidth]{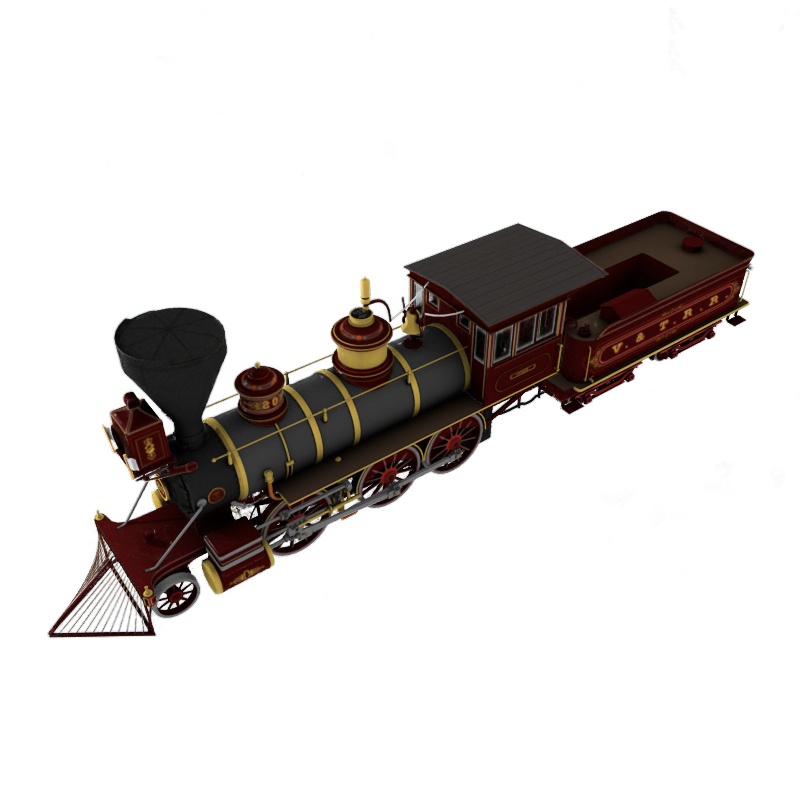} &
        \includegraphics[width=0.15\textwidth]{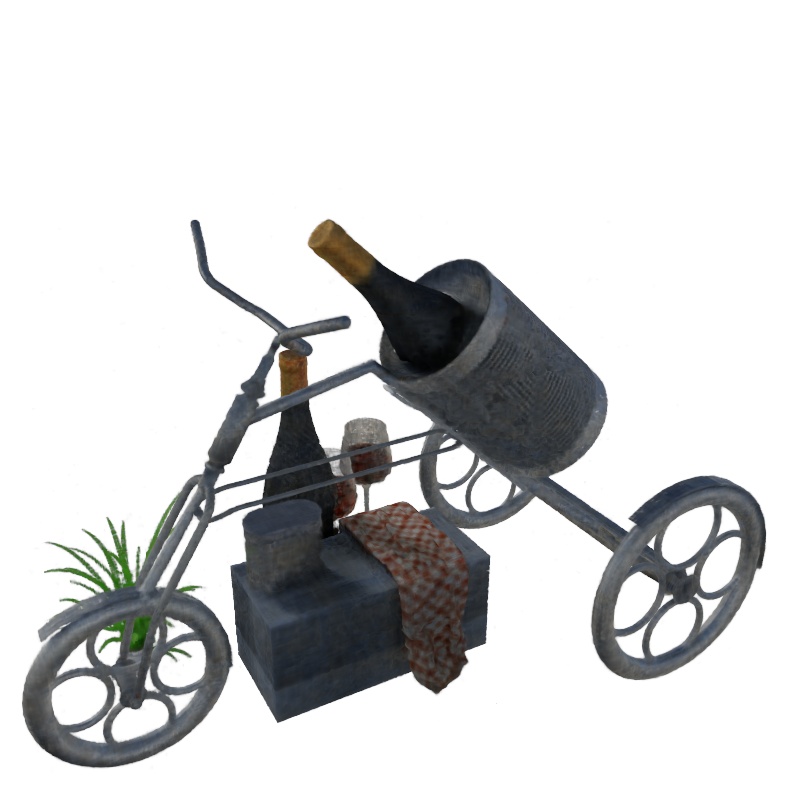} &
        \includegraphics[width=0.15\textwidth]{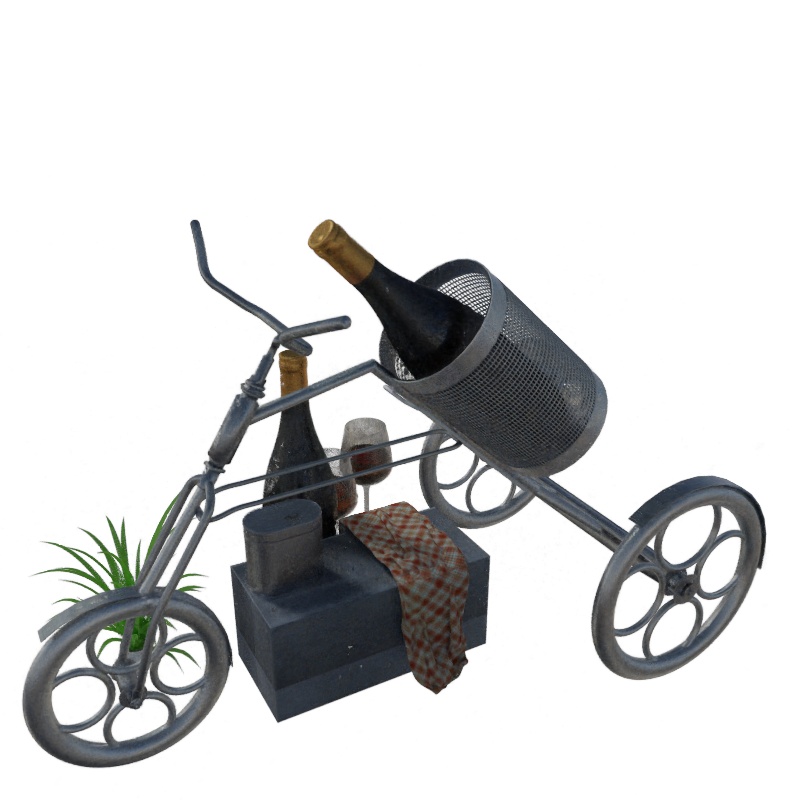} &
        \includegraphics[width=0.15\textwidth]{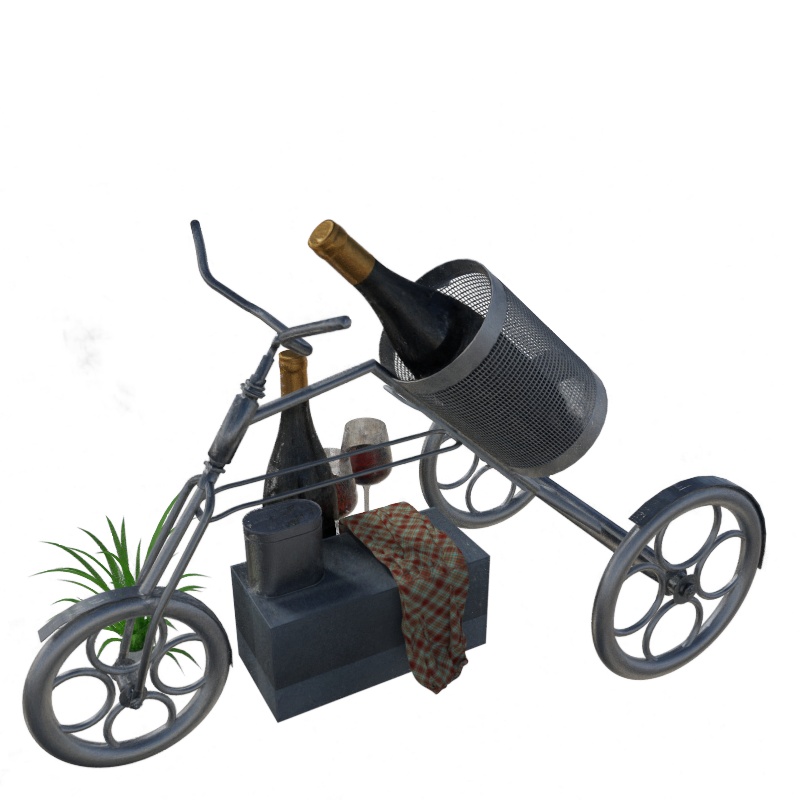}\\
         PPNG-1 & PPNG-2 & PPNG-3 & PPNG-1 & PPNG-2 & PPNG-3
    \end{tabular}
    \caption{Qualitative results for Synthetic NSVF dataset}
    \label{fig:appendix_nsvf}
\end{figure*}
\begin{figure*}[t]
    \centering
    \begin{tabular}{c@{}c@{}c}
        \includegraphics[width=0.3\textwidth]{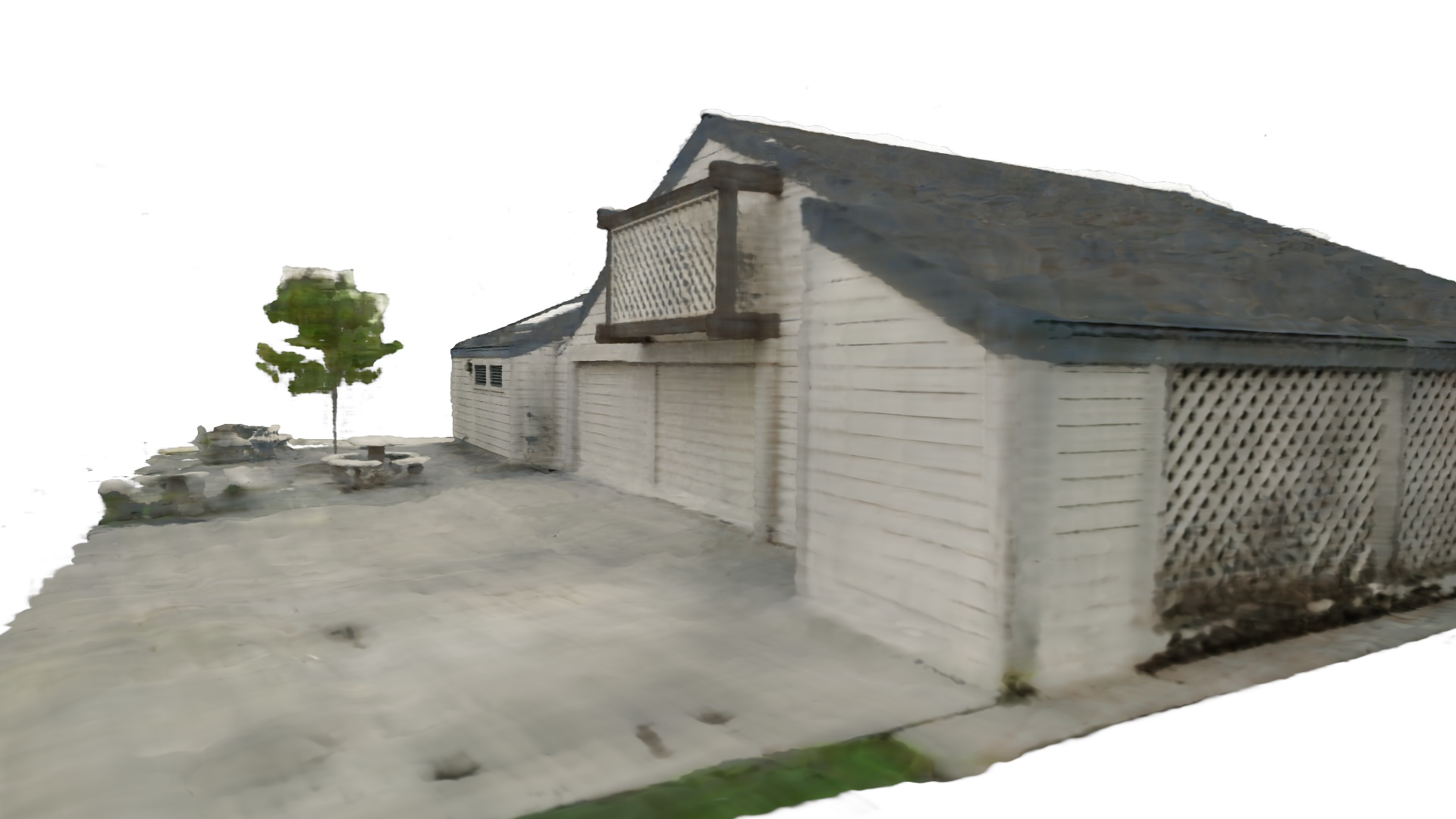} &
        \includegraphics[width=0.3\textwidth]{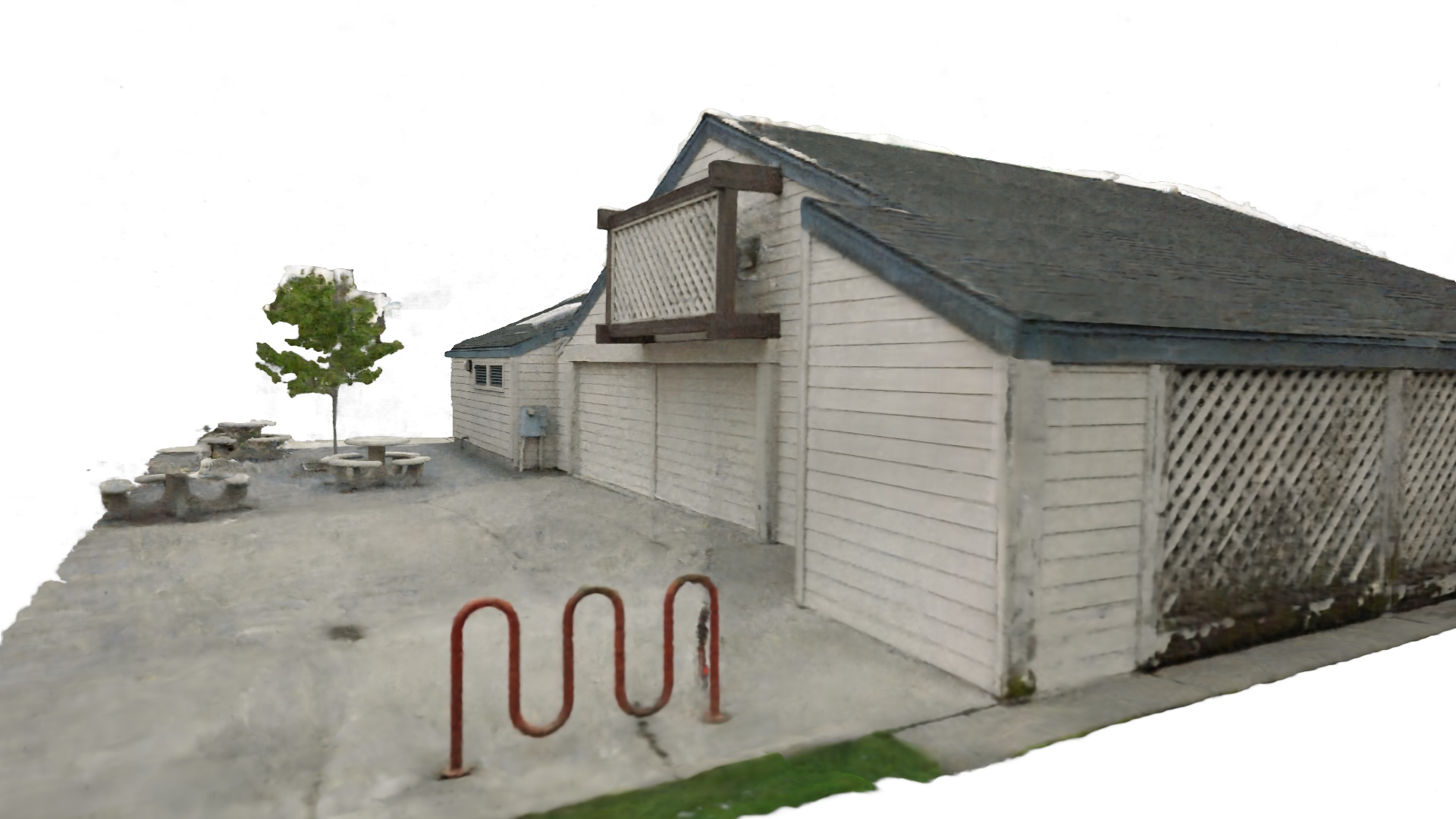} &
        \includegraphics[width=0.3\textwidth]{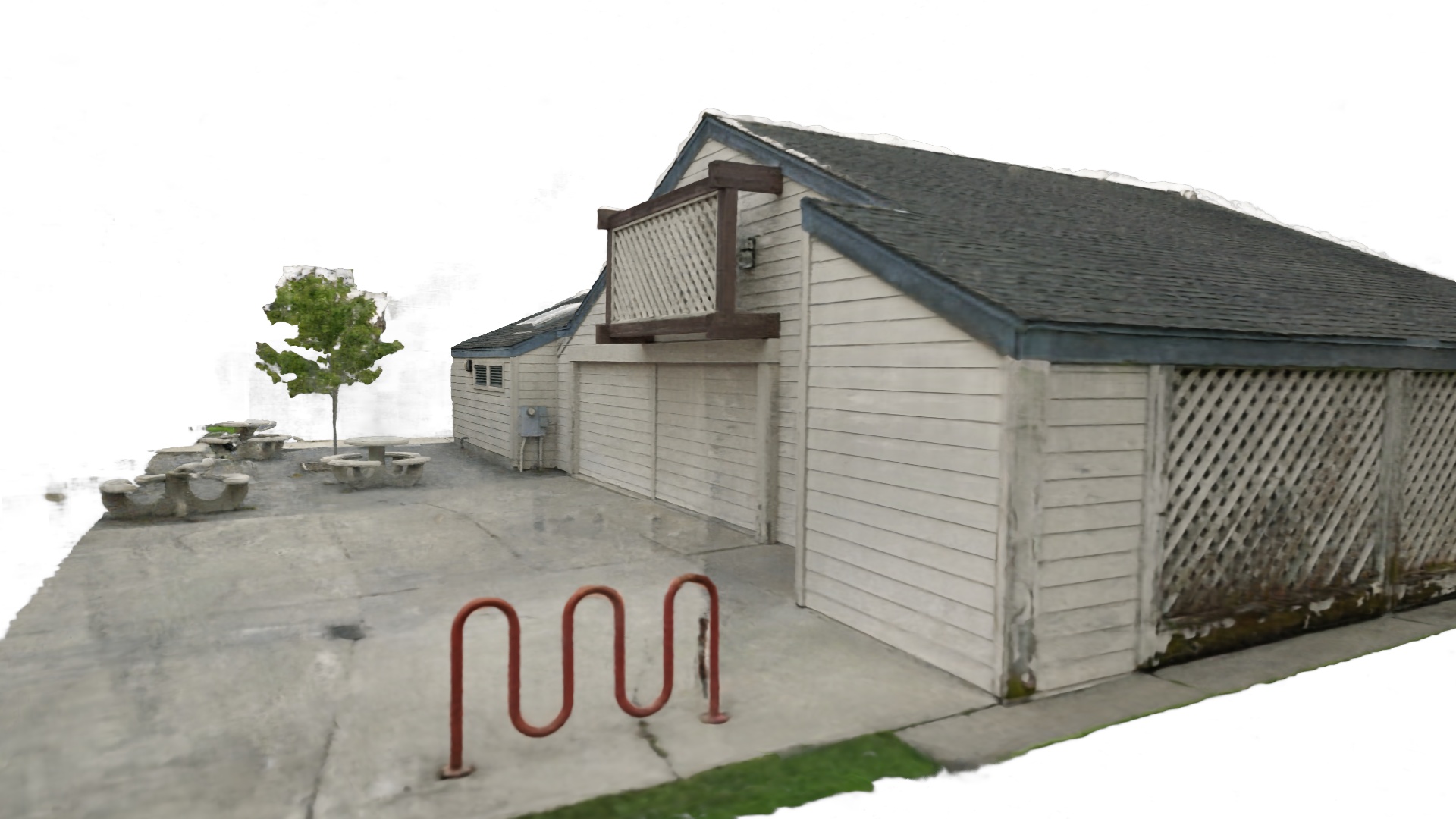}\\
        \includegraphics[width=0.3\textwidth]{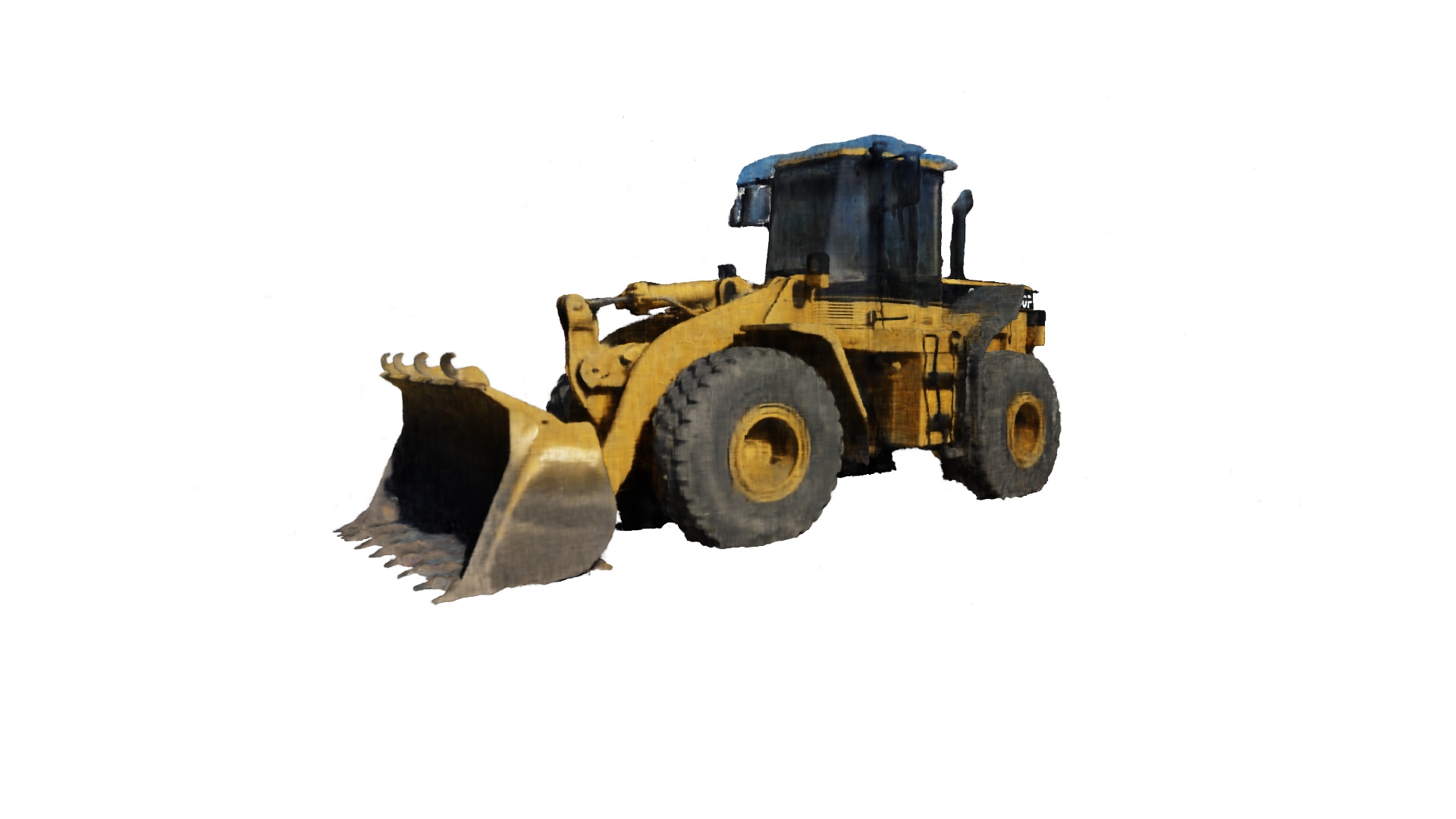} &
        \includegraphics[width=0.3\textwidth]{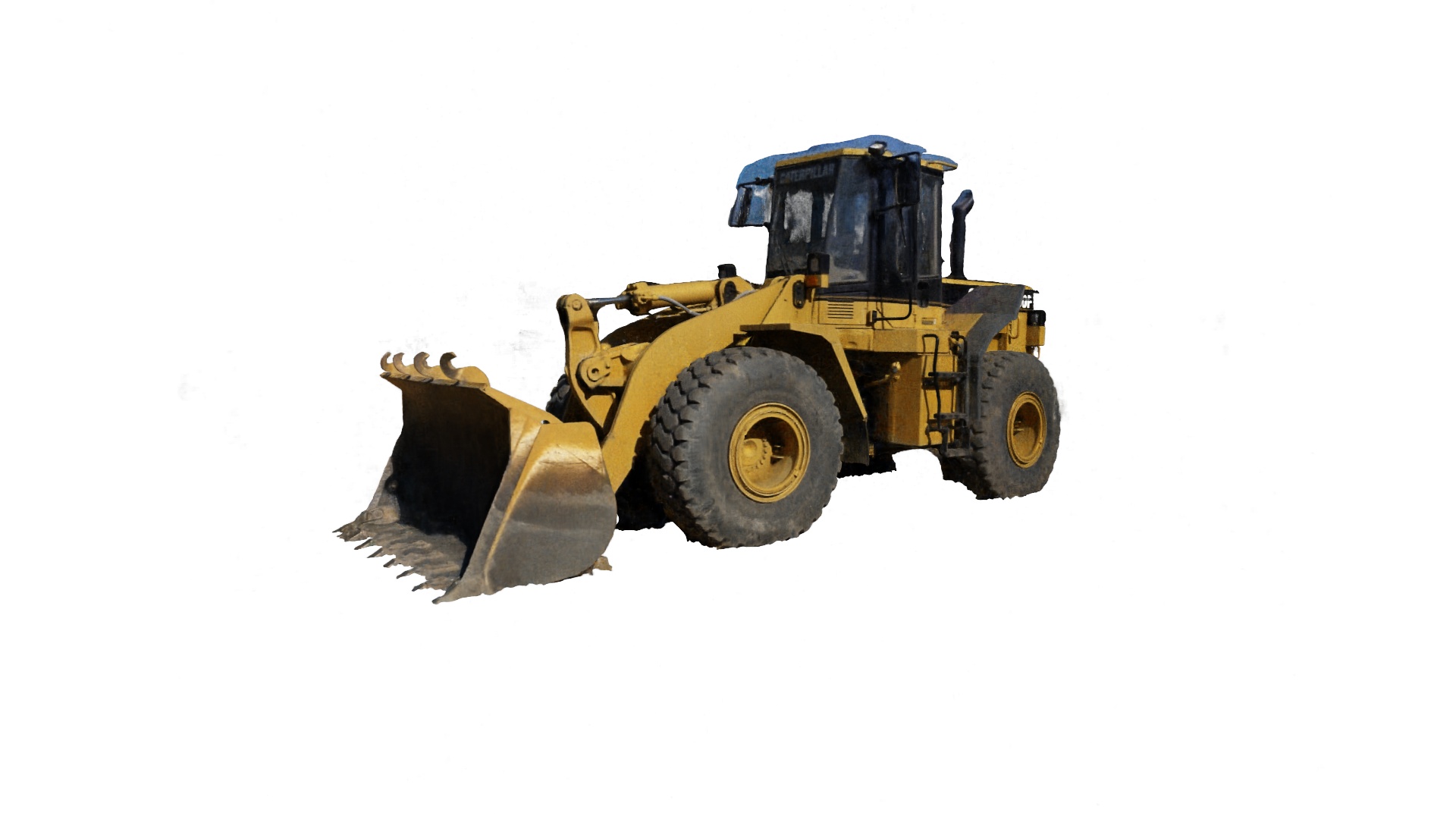} &
        \includegraphics[width=0.3\textwidth]{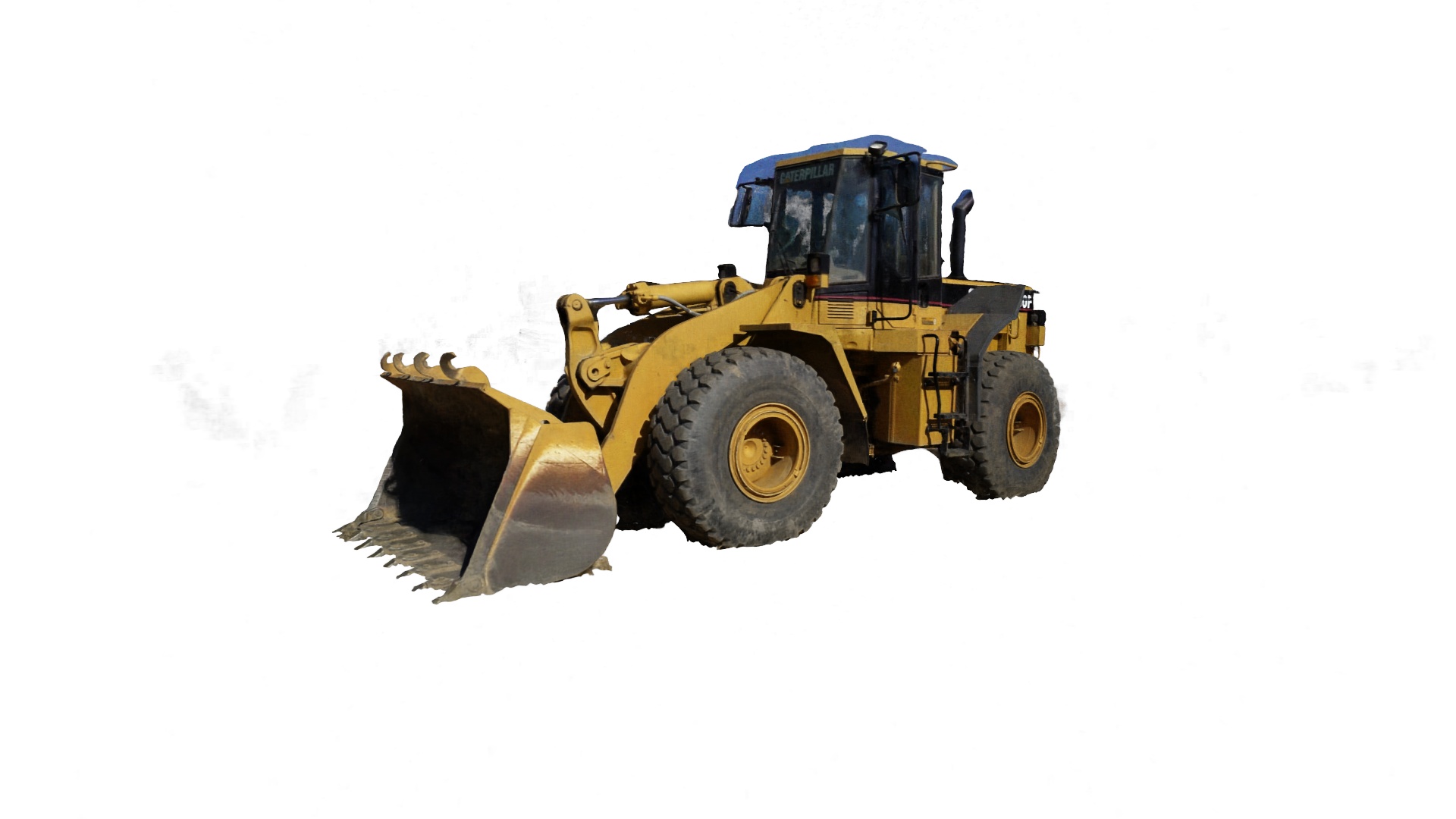}\\
        \includegraphics[width=0.3\textwidth]{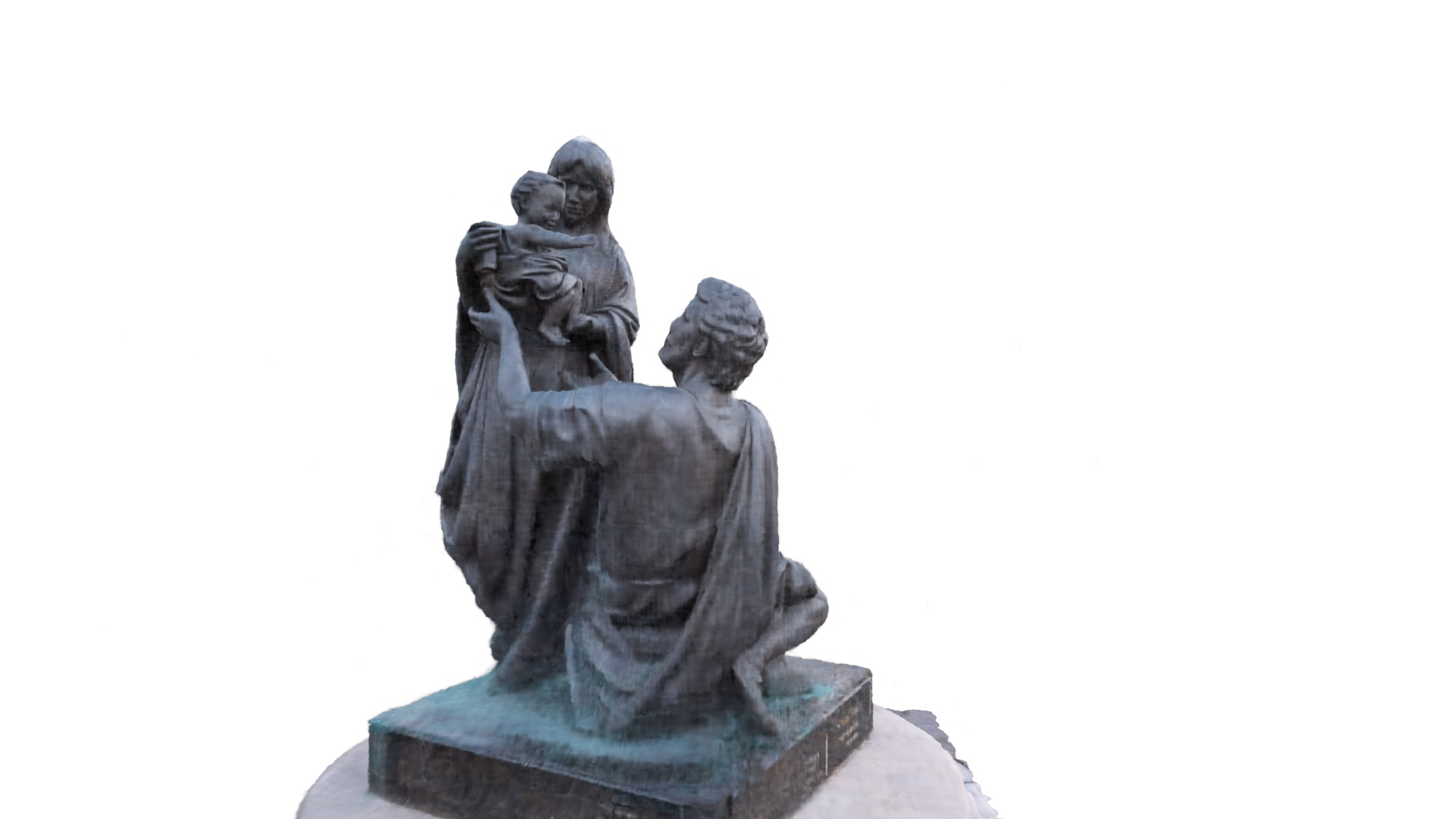} &
        \includegraphics[width=0.3\textwidth]{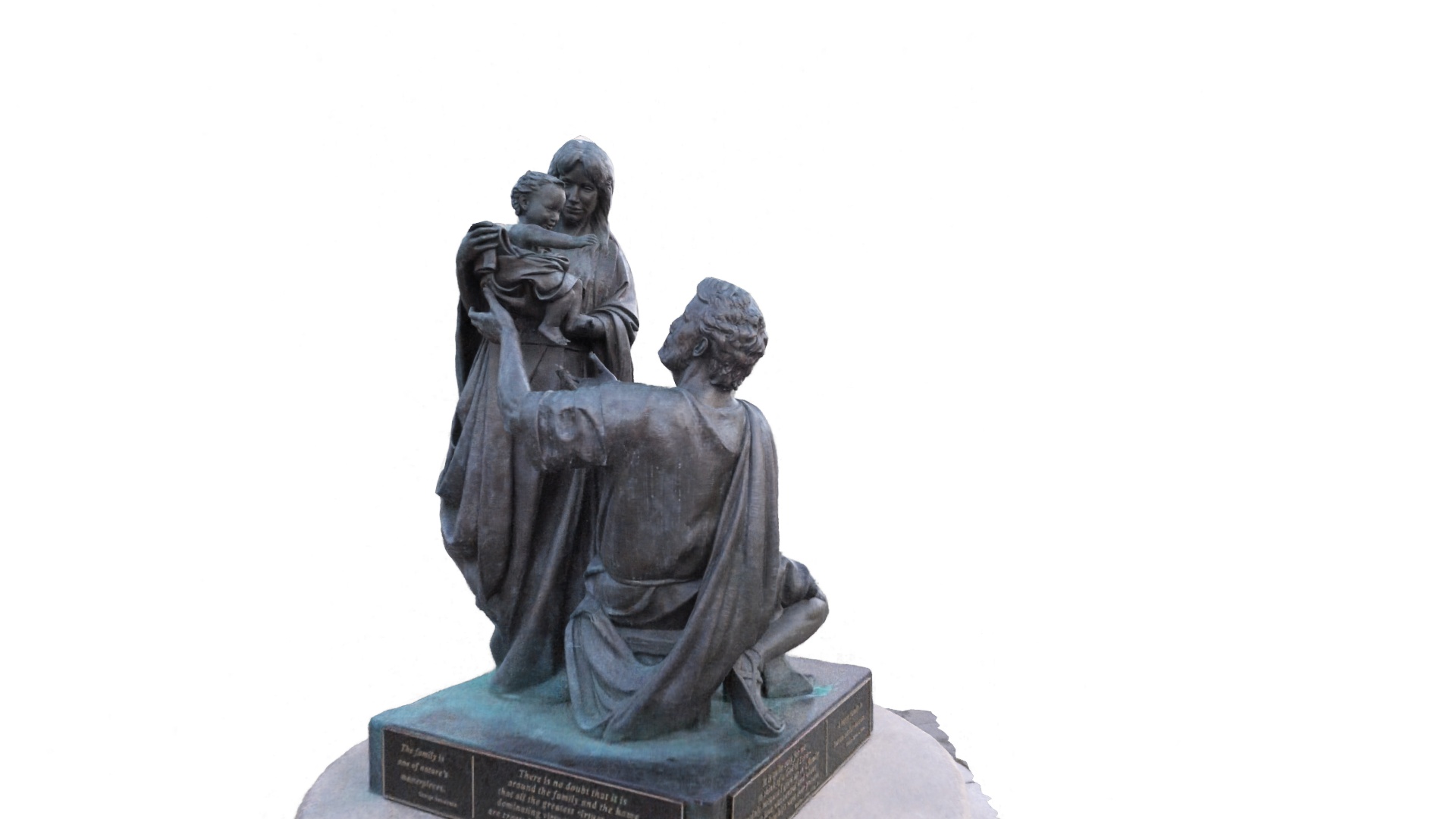} &
        \includegraphics[width=0.3\textwidth]{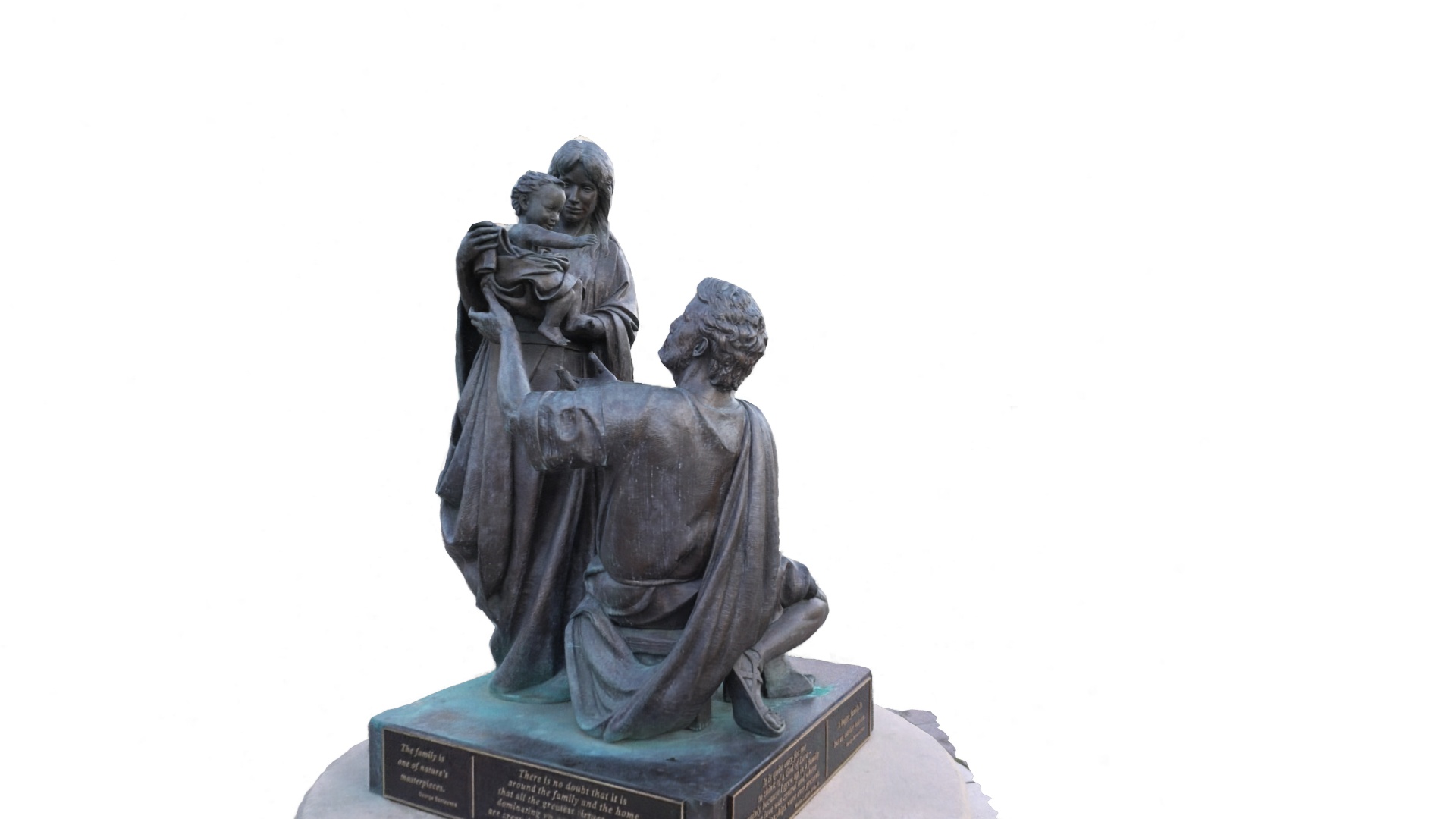}\\
        \includegraphics[width=0.3\textwidth]{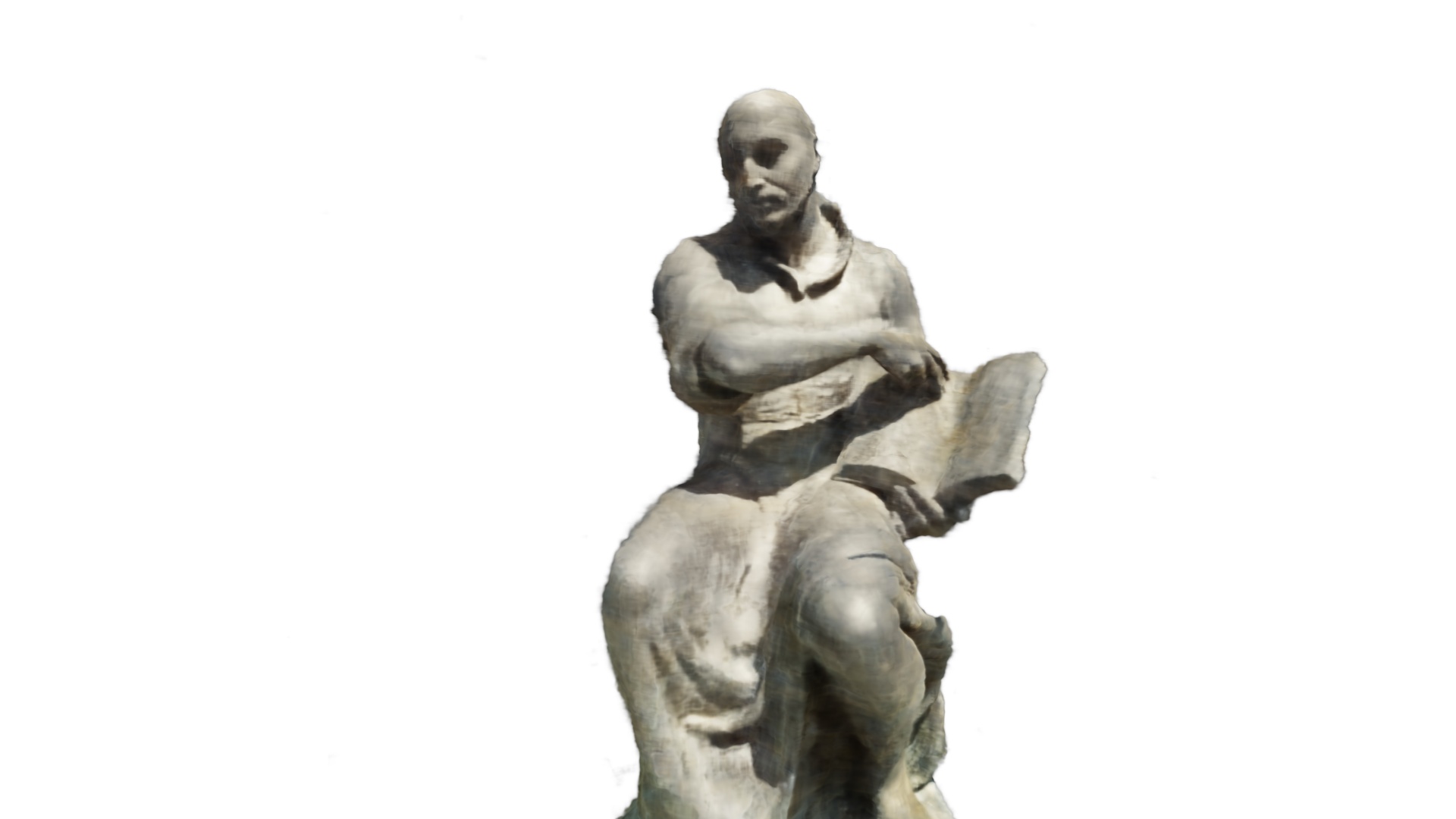} &
        \includegraphics[width=0.3\textwidth]{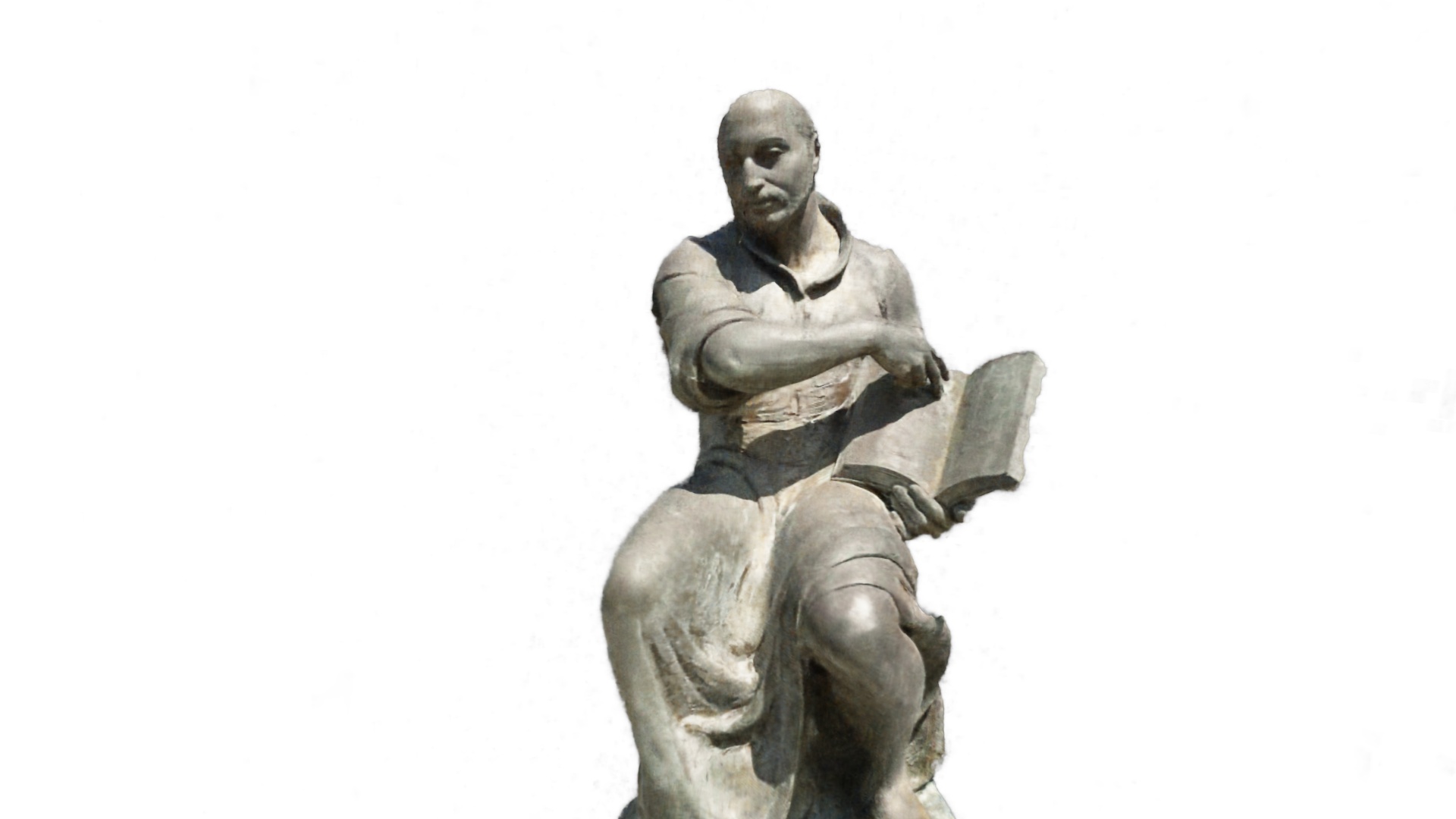} &
        \includegraphics[width=0.3\textwidth]{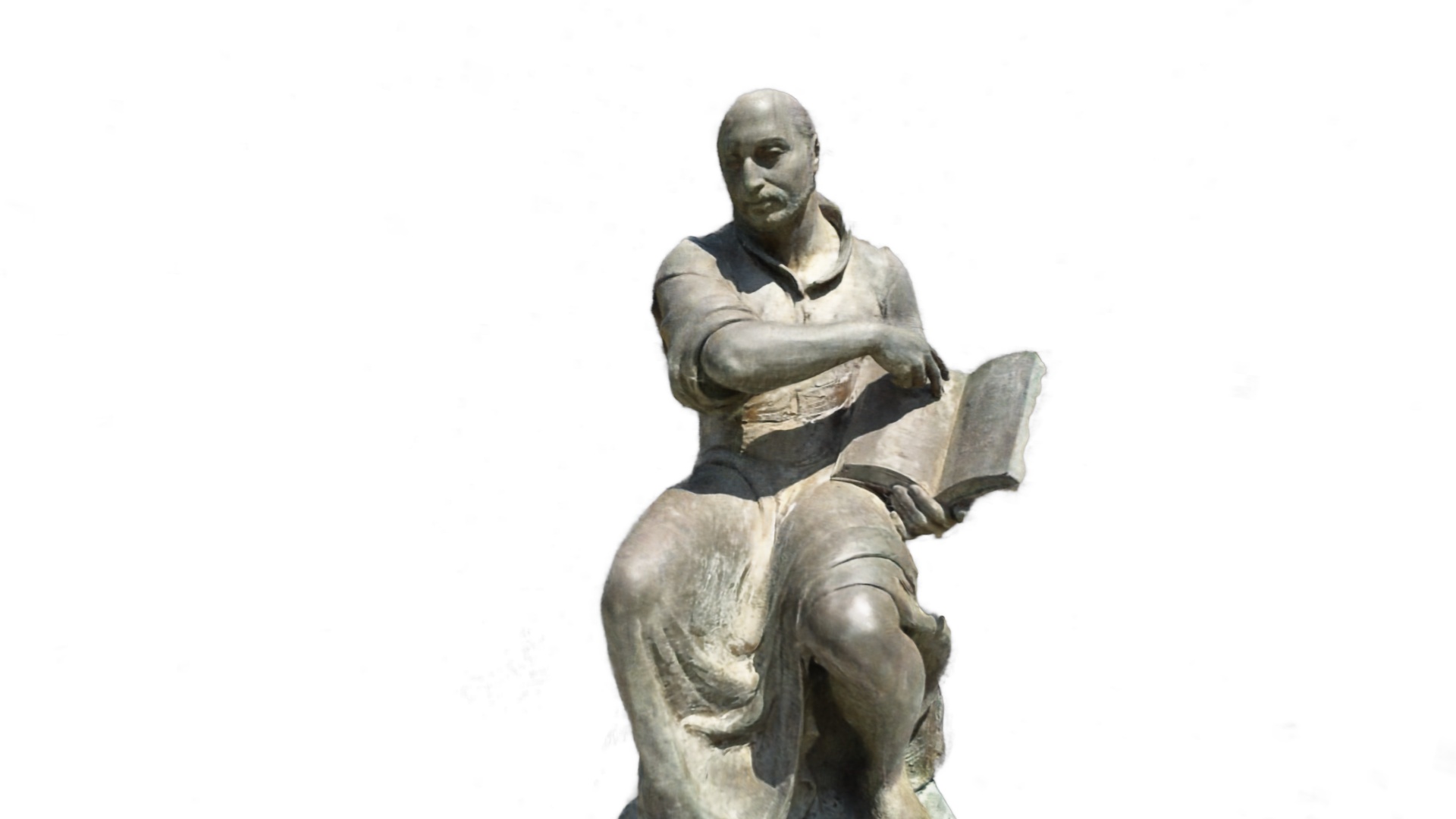}\\
        \includegraphics[width=0.3\textwidth]{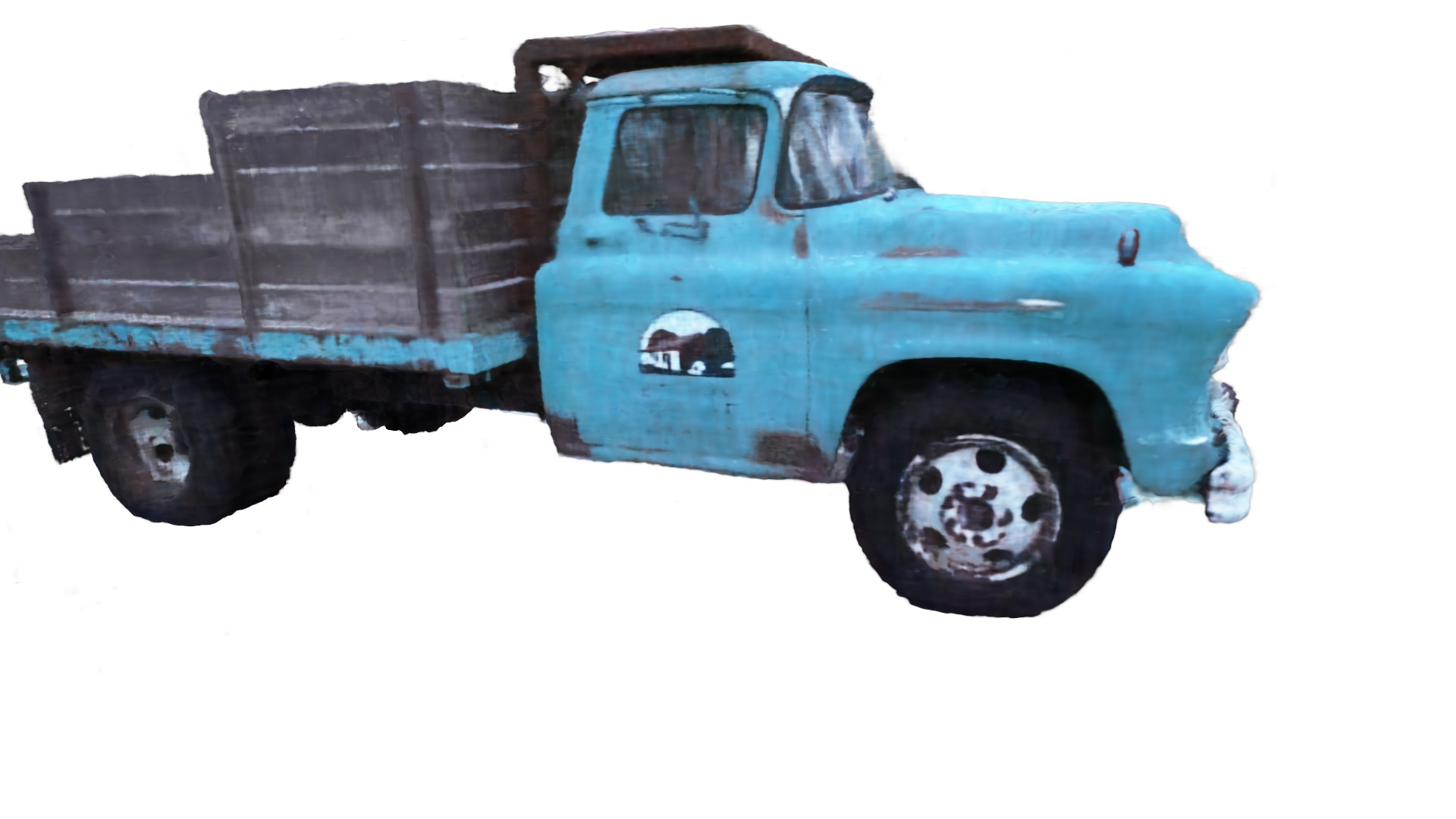} &
        \includegraphics[width=0.3\textwidth]{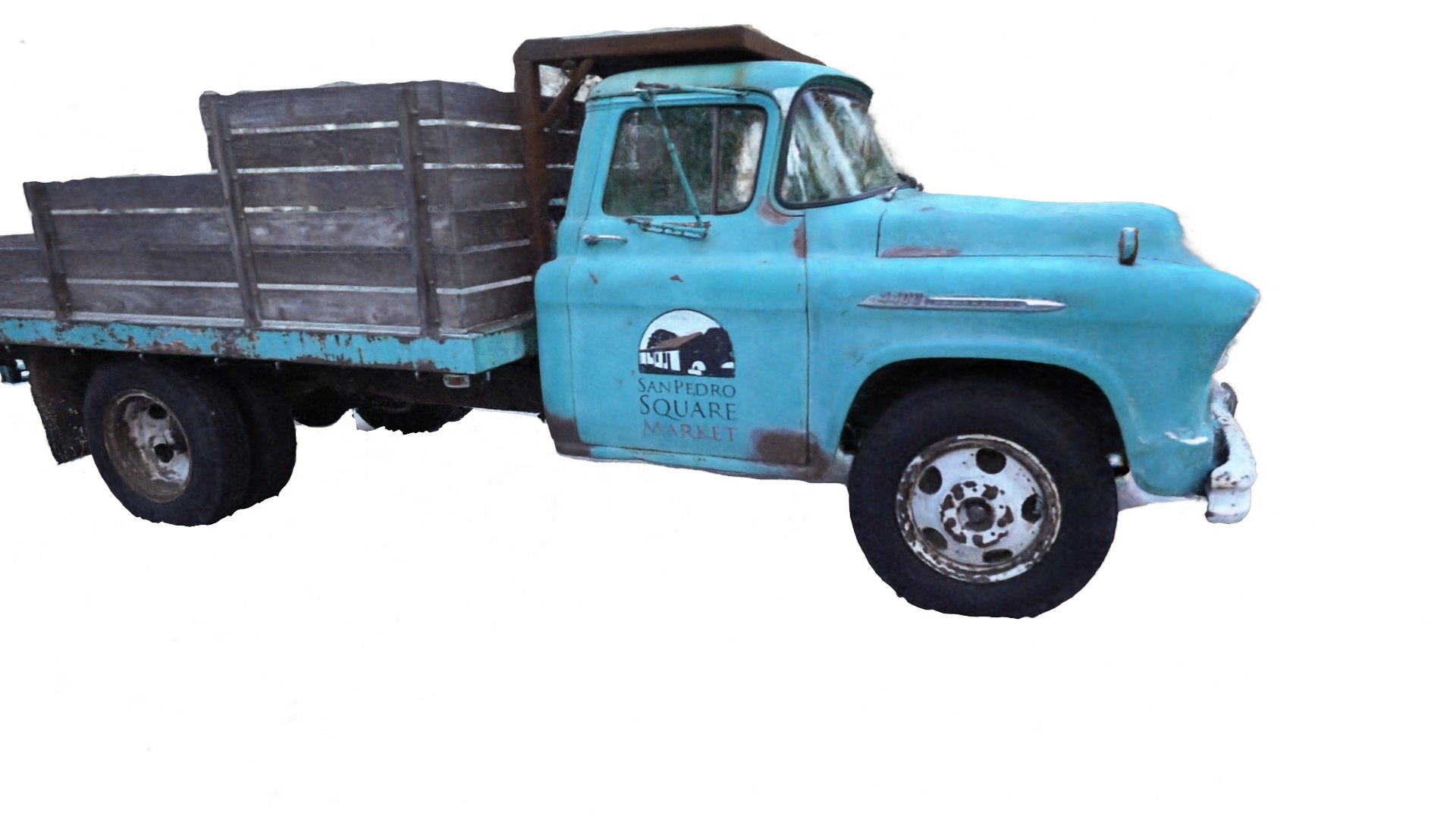} &
        \includegraphics[width=0.3\textwidth]{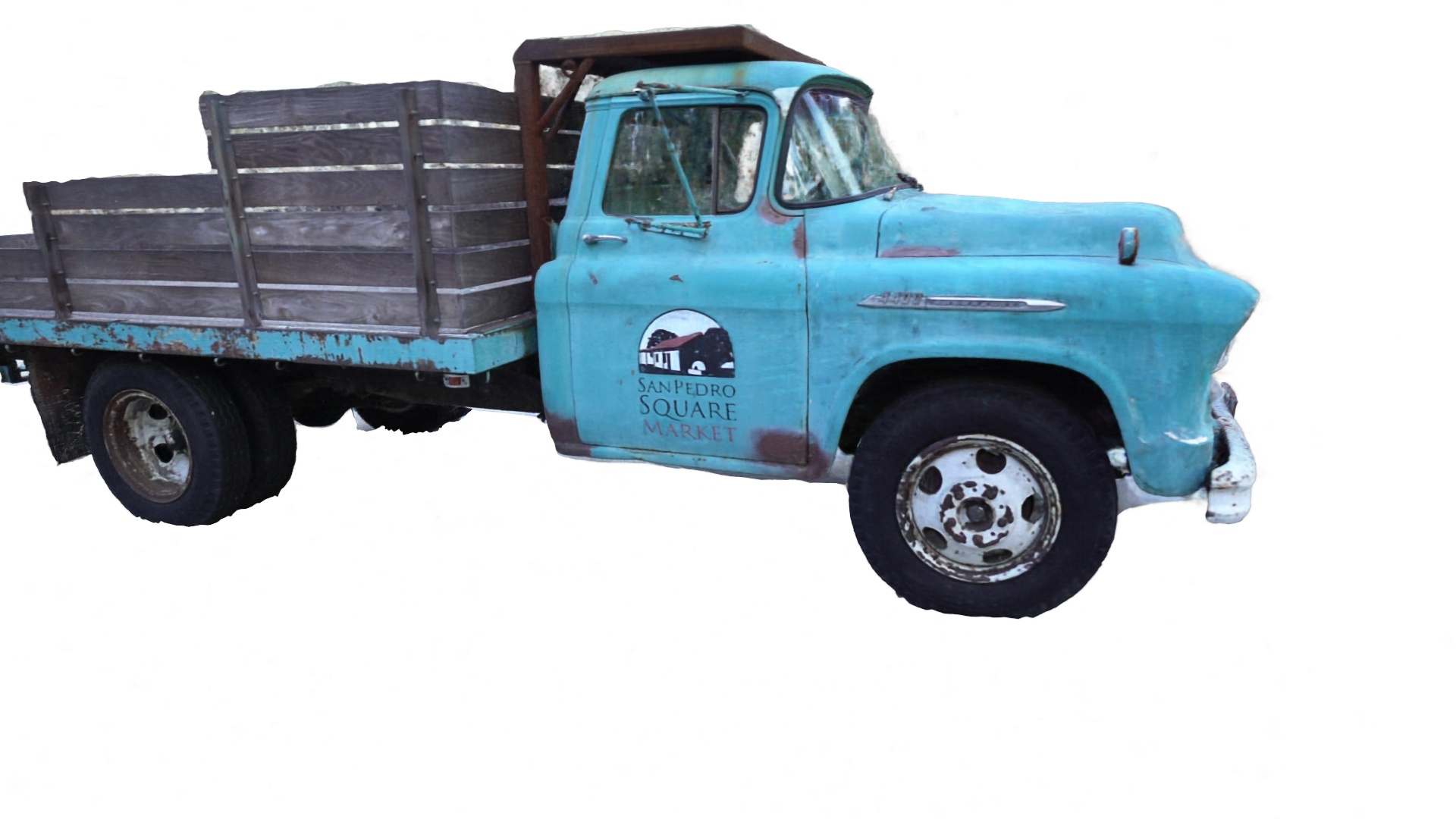}\\
         PPNG-1 & PPNG-2 & PPNG-3
    \end{tabular}
    \caption{Qualitative results for Tanks and Temples dataset}
    \label{fig:appendix_tnt}
\end{figure*}
\begin{figure*}[t]
    \centering
    \resizebox{\textwidth}{!}{
    \begin{tabular}{c@{}c@{}c}
        \includegraphics[width=0.3\textwidth]{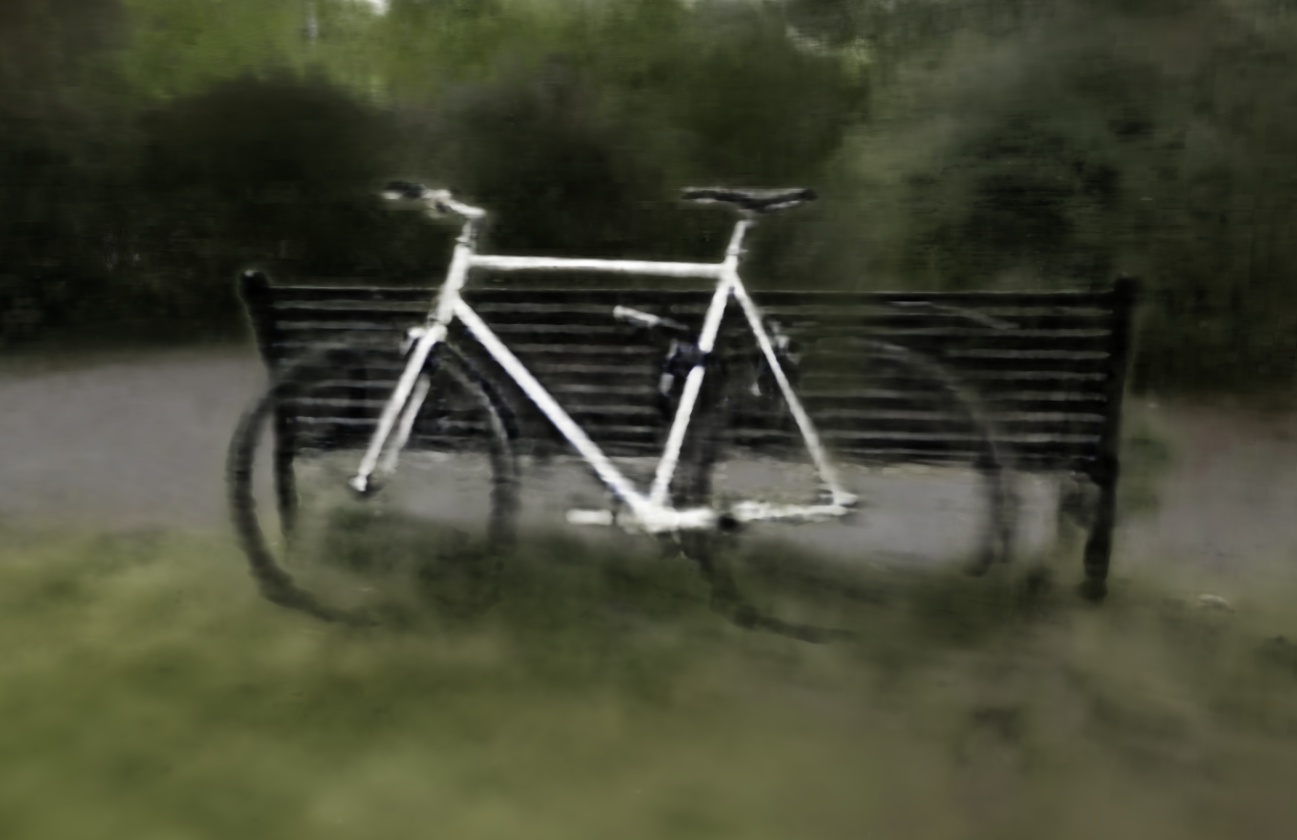} &
        \includegraphics[width=0.3\textwidth]{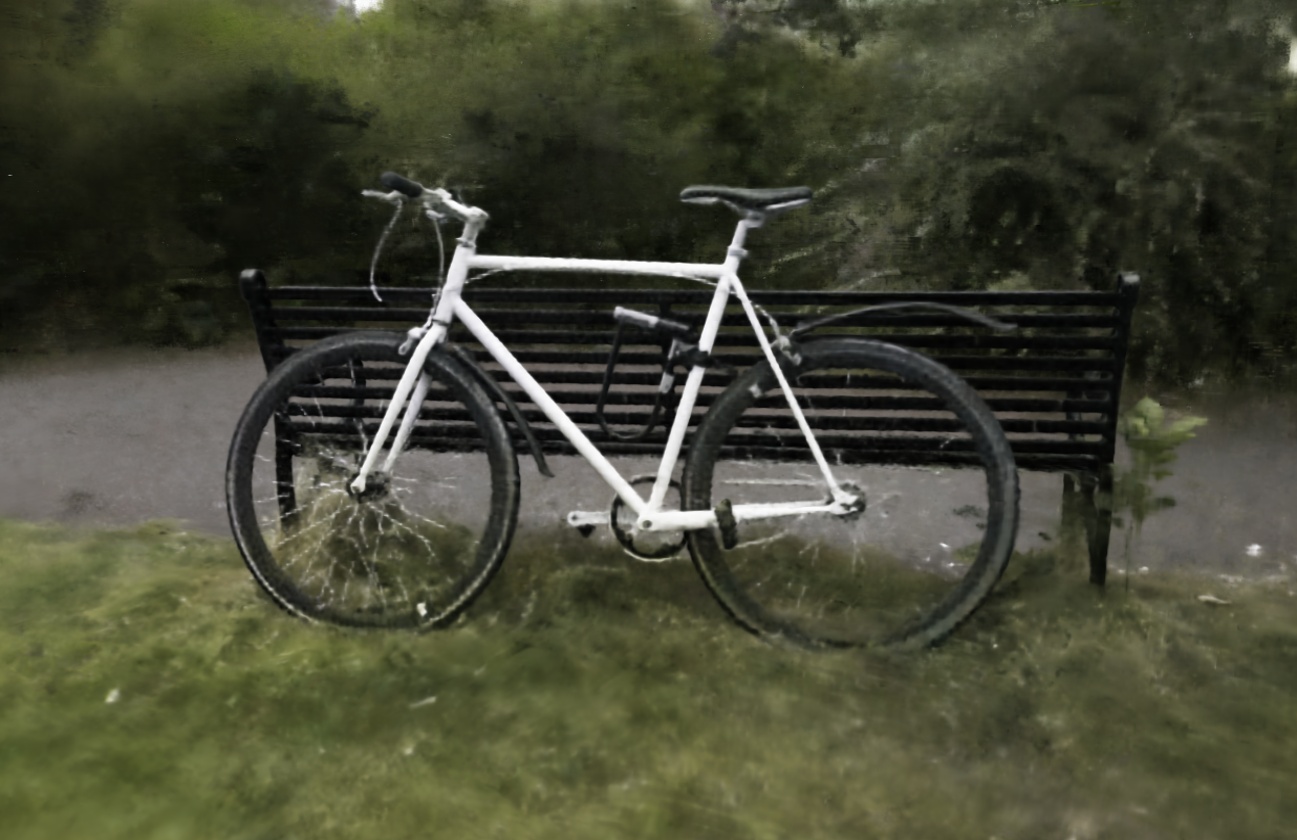} &
        \includegraphics[width=0.3\textwidth]{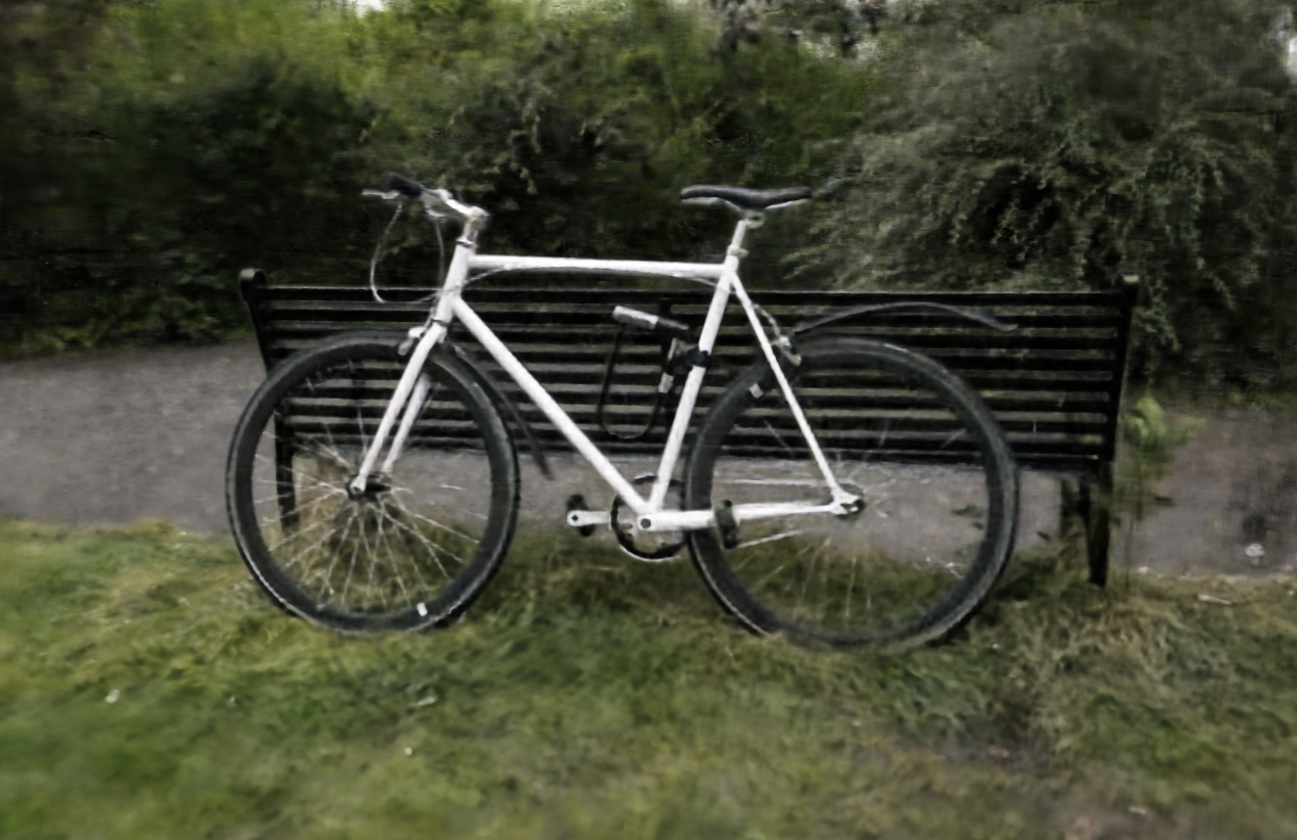}\\
        \includegraphics[width=0.3\textwidth]{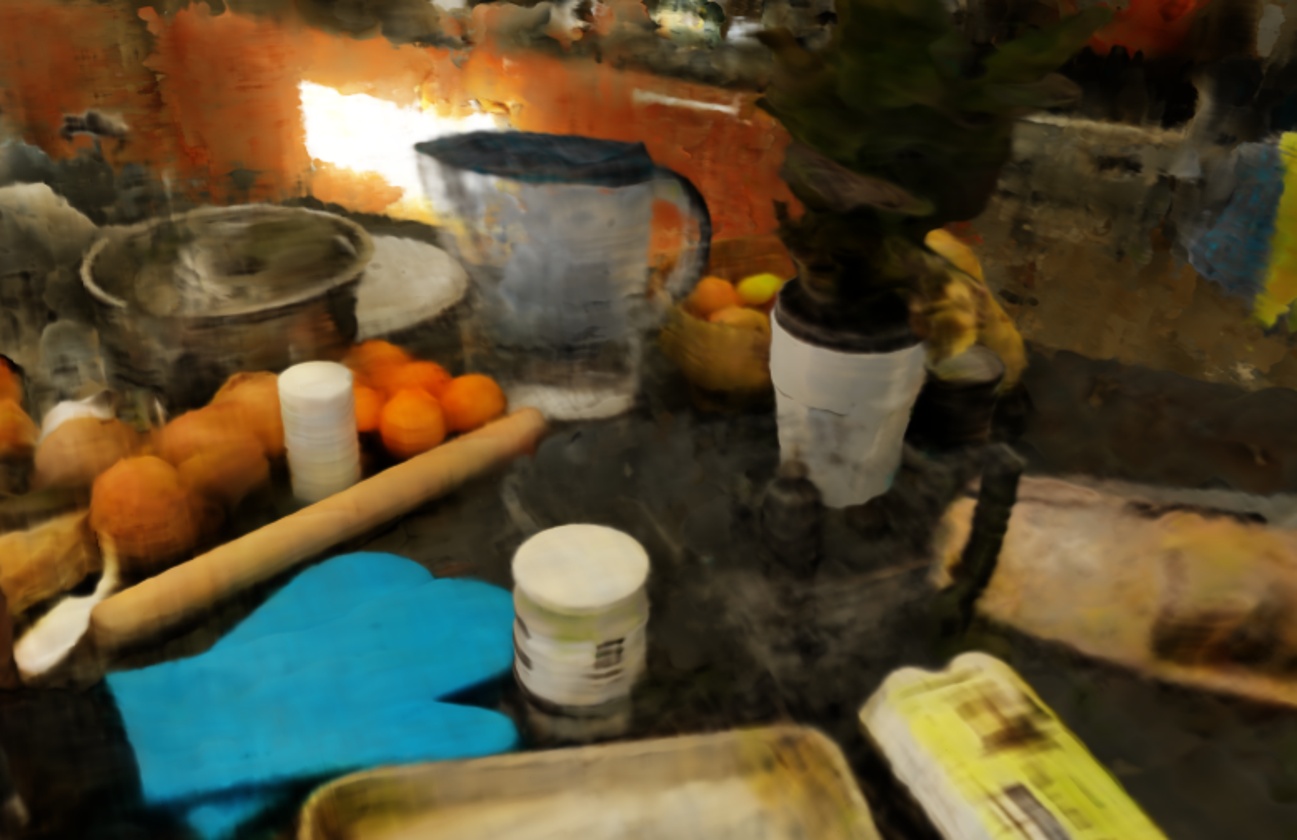} &
        \includegraphics[width=0.3\textwidth]{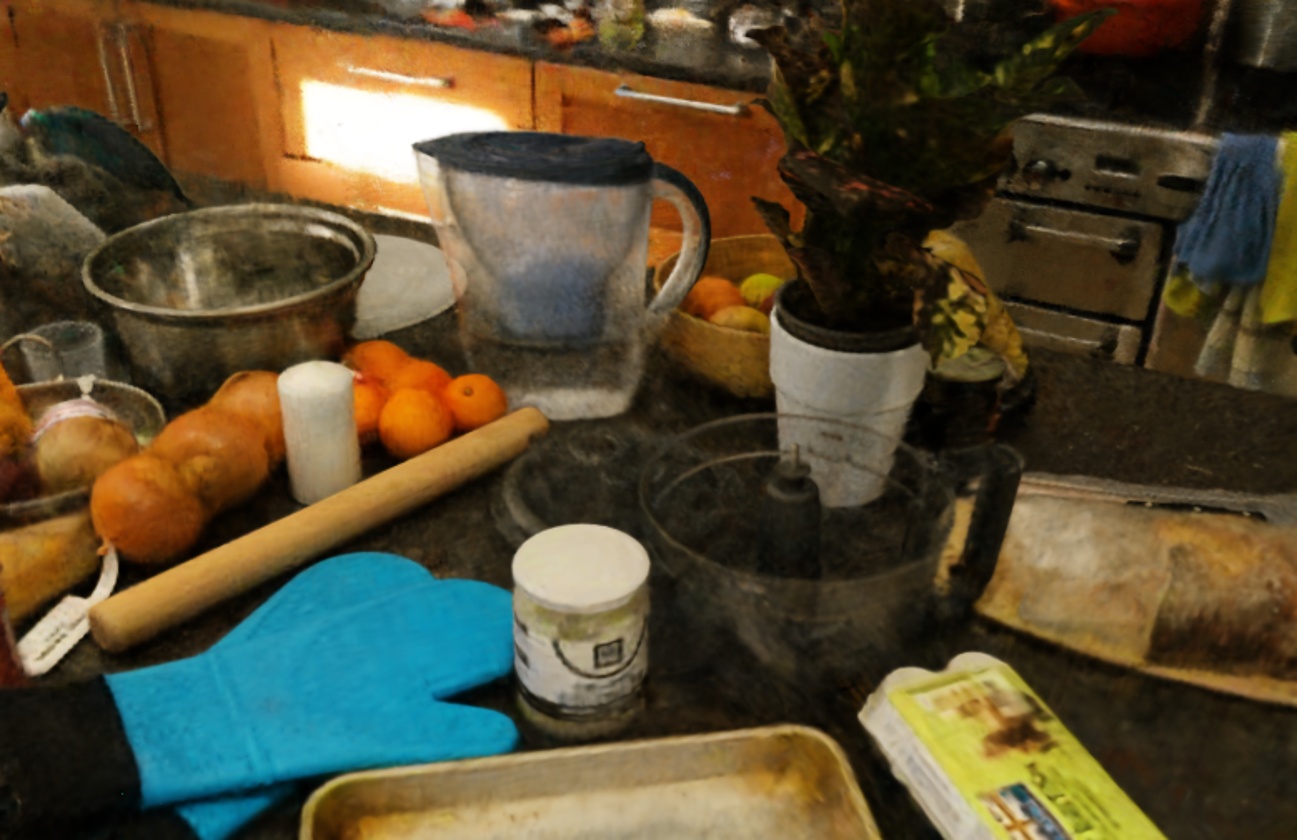} &
        \includegraphics[width=0.3\textwidth]{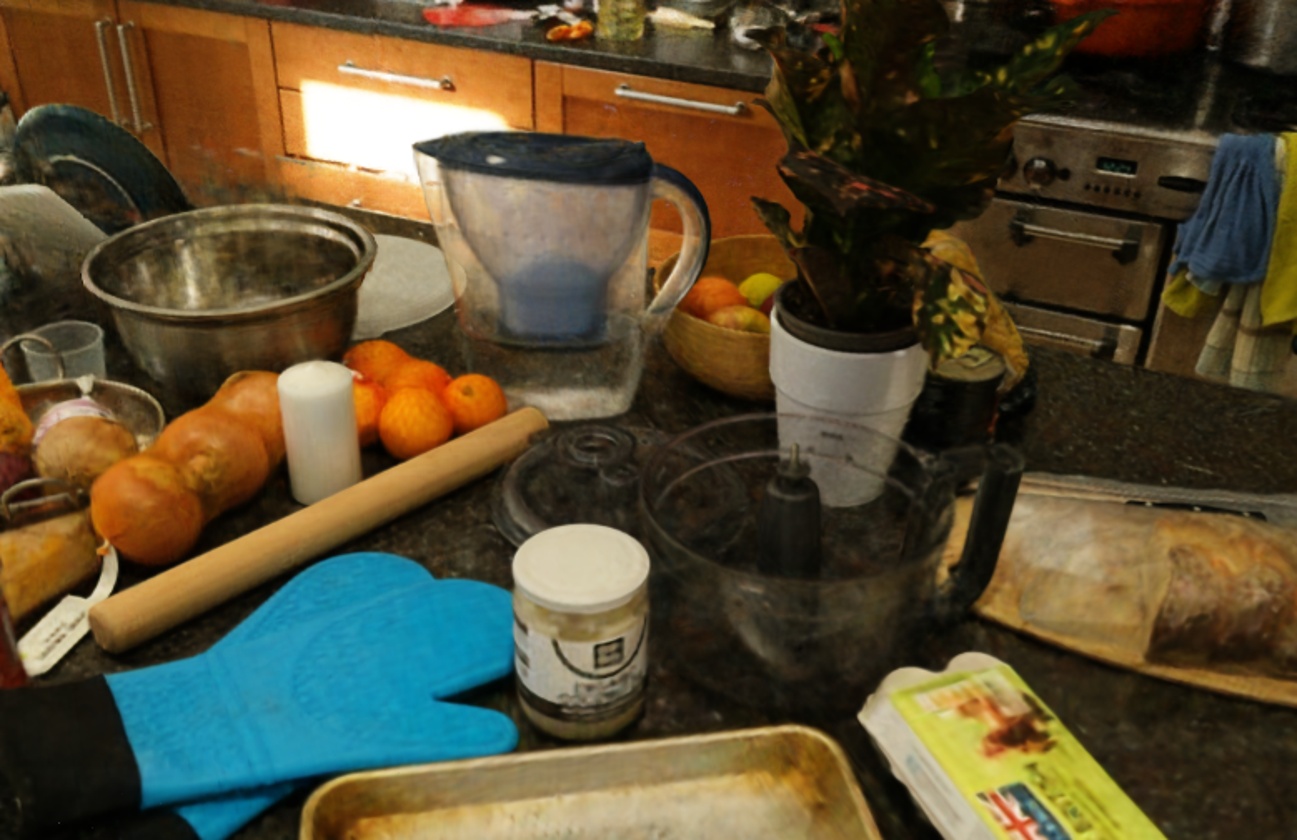}\\
        \includegraphics[width=0.3\textwidth]{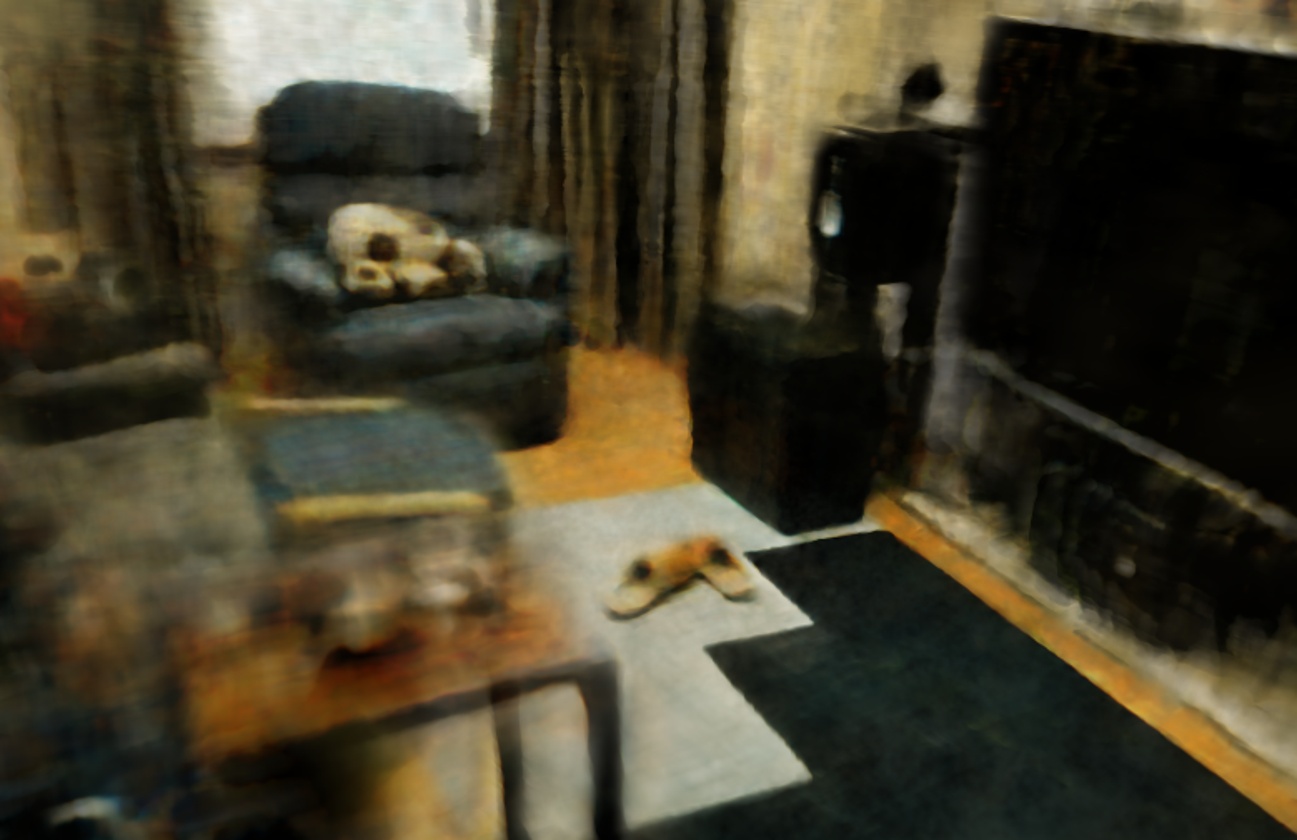} &
        \includegraphics[width=0.3\textwidth]{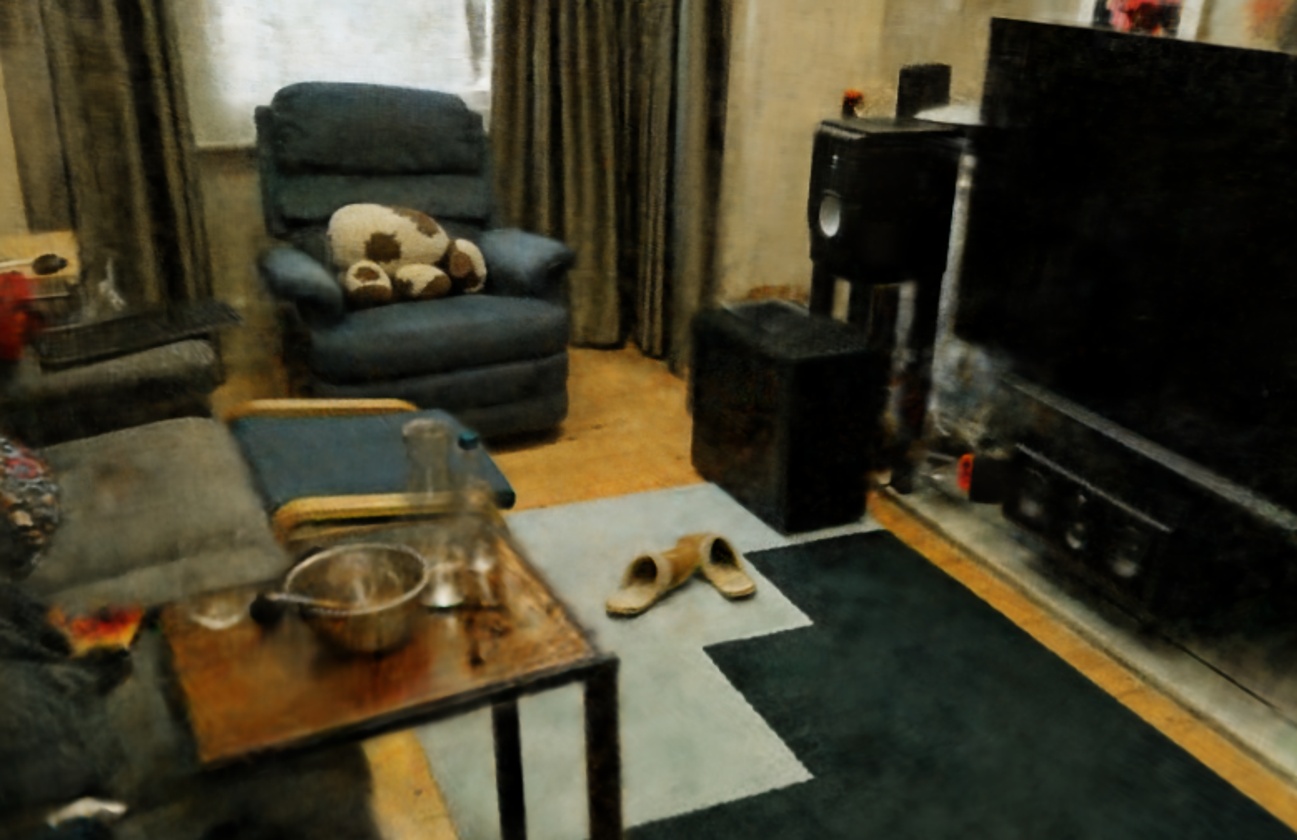} &
        \includegraphics[width=0.3\textwidth]{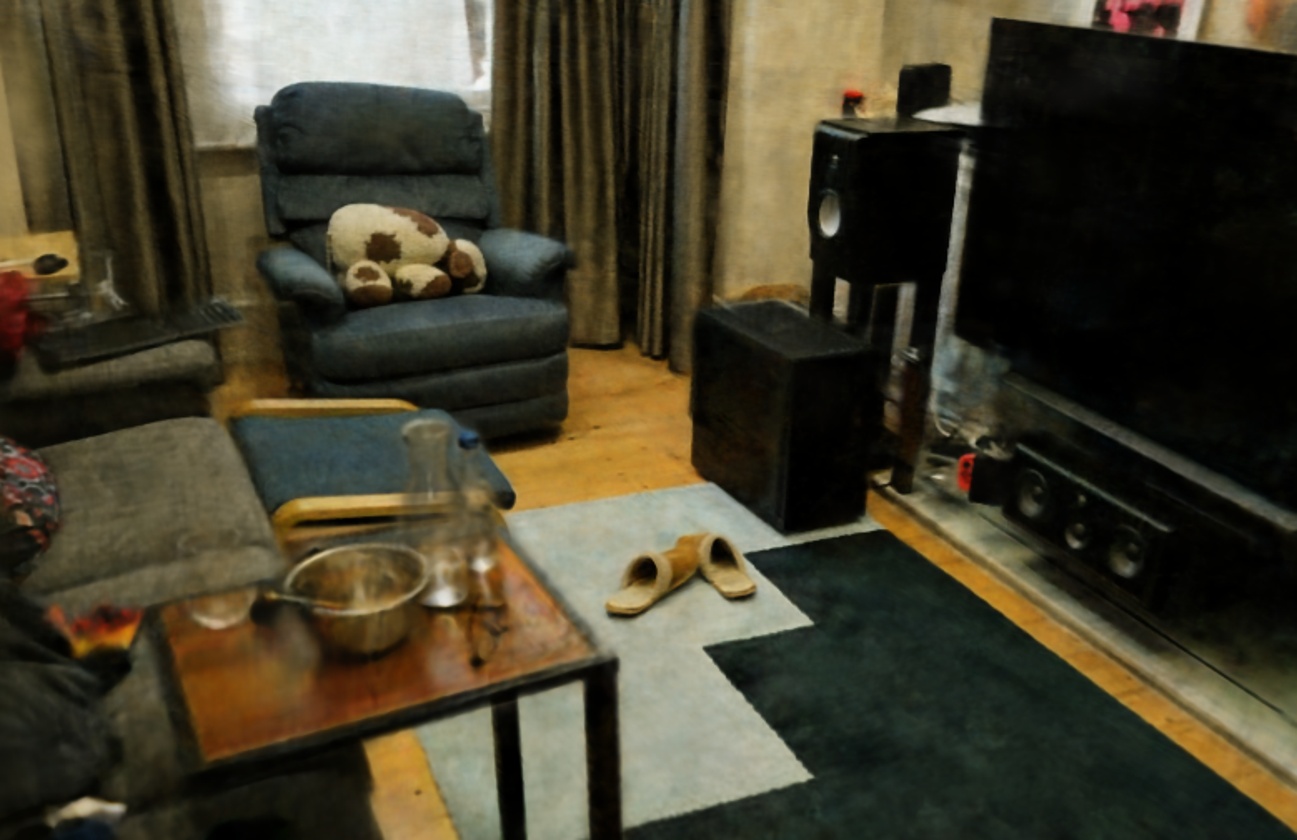}\\
        \includegraphics[width=0.3\textwidth]{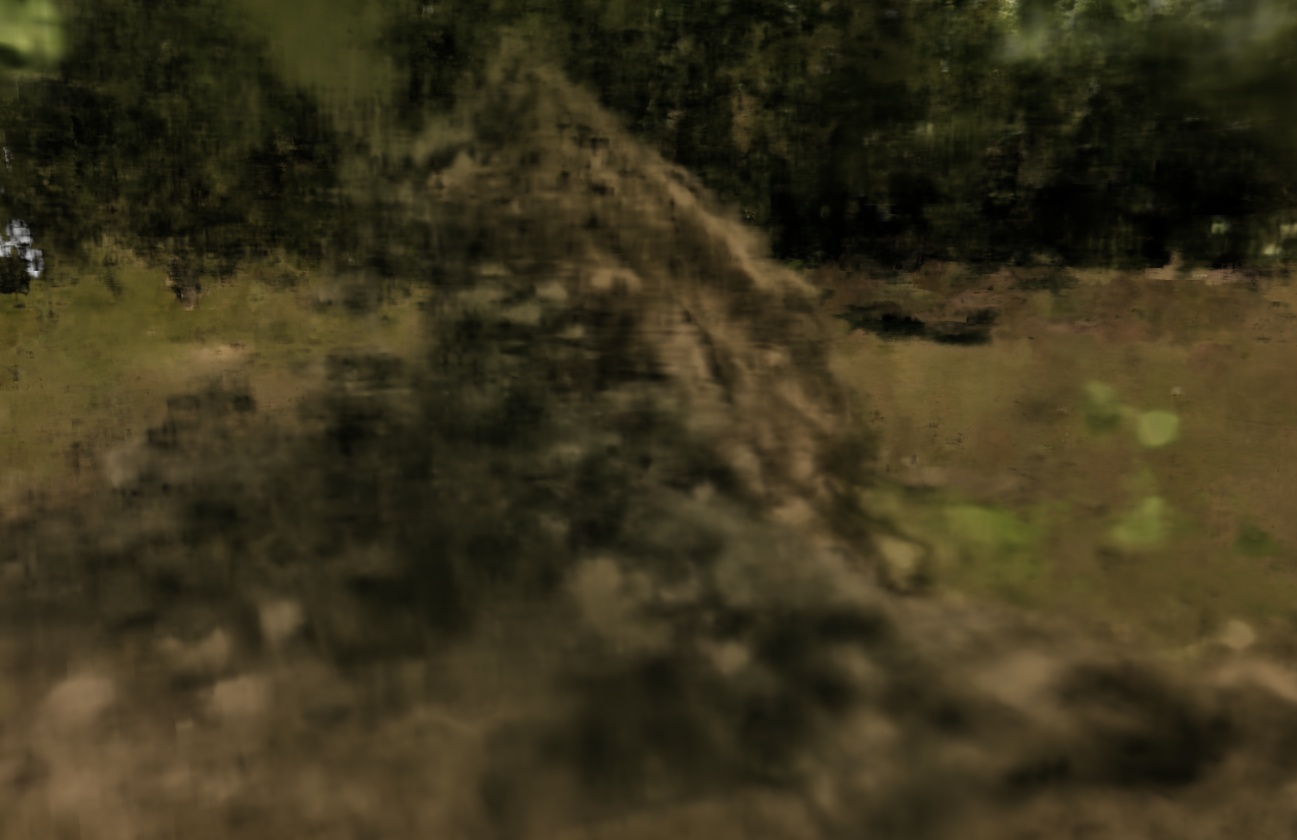} &
        \includegraphics[width=0.3\textwidth]{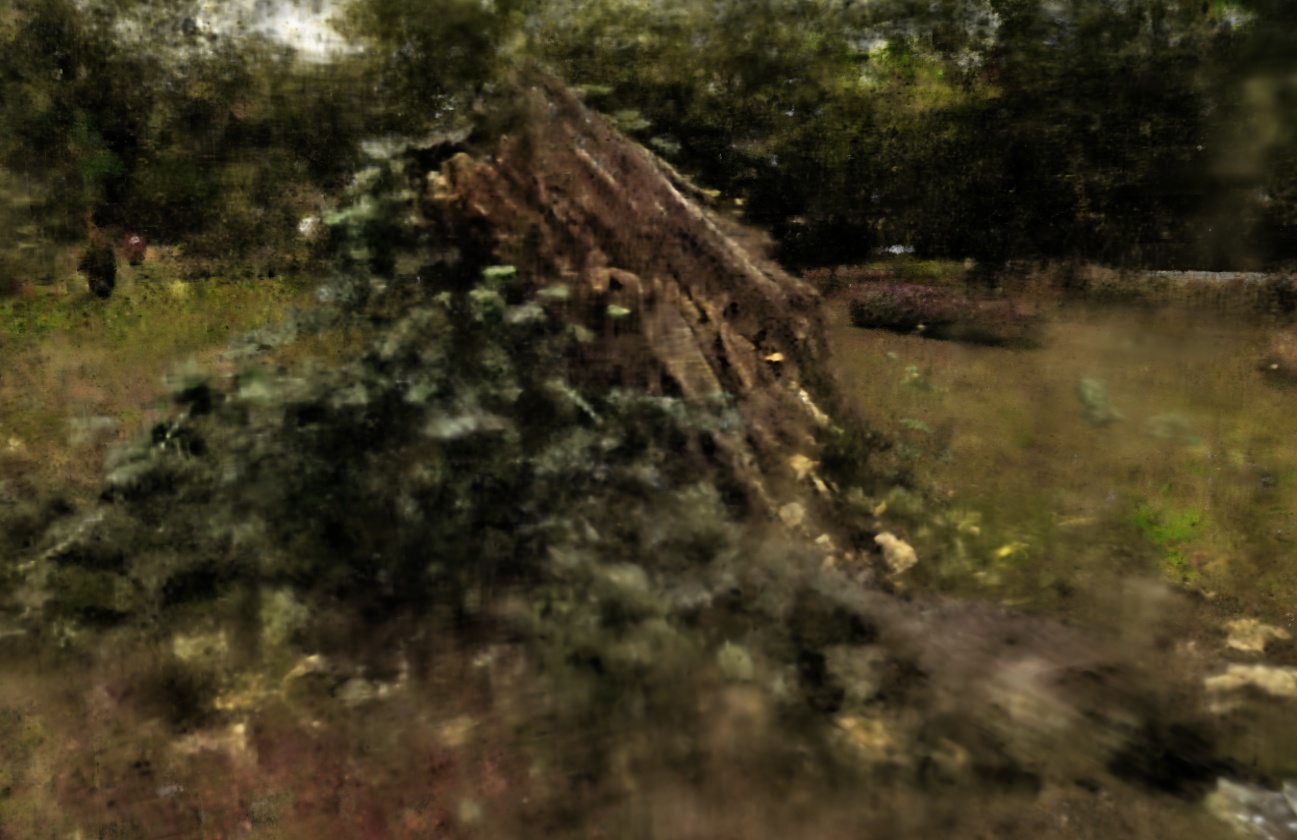} &
        \includegraphics[width=0.3\textwidth]{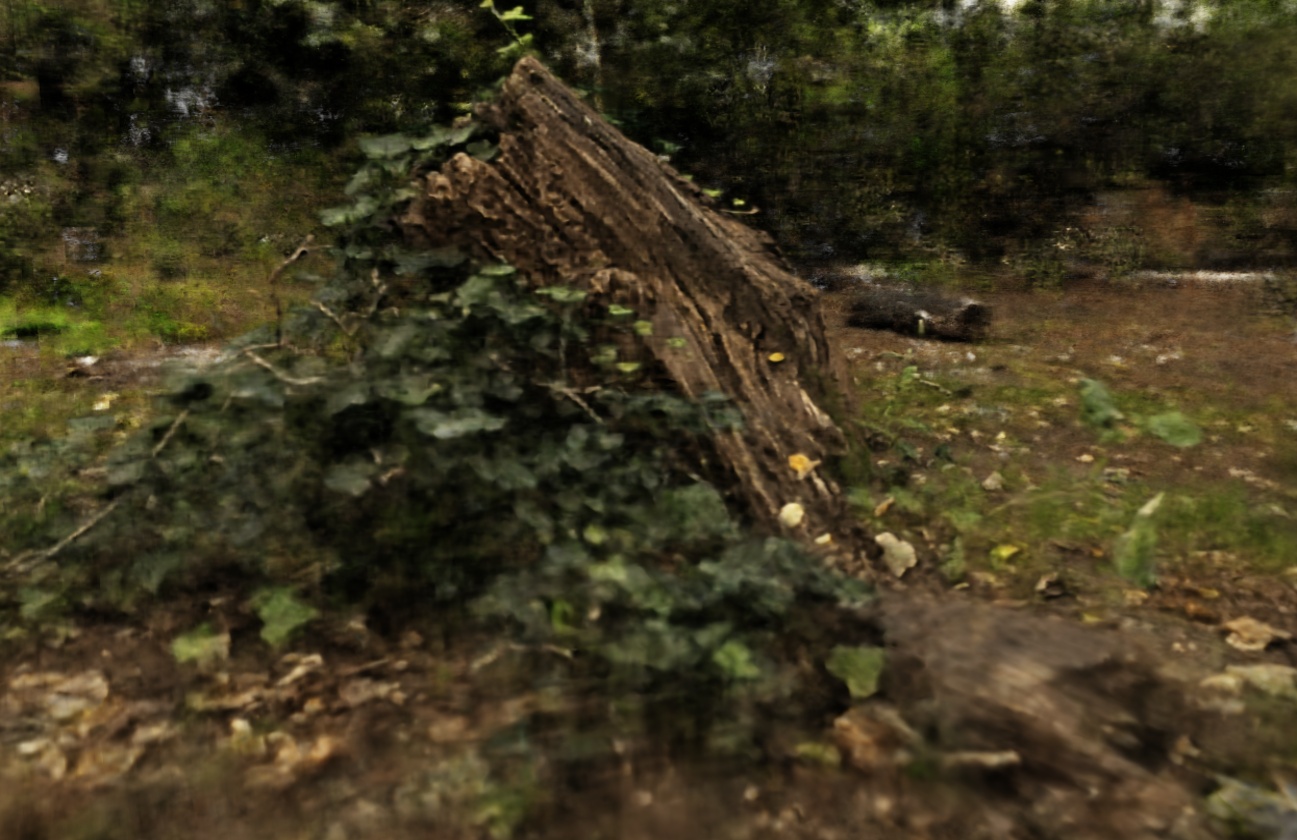}\\
        \includegraphics[width=0.3\textwidth]{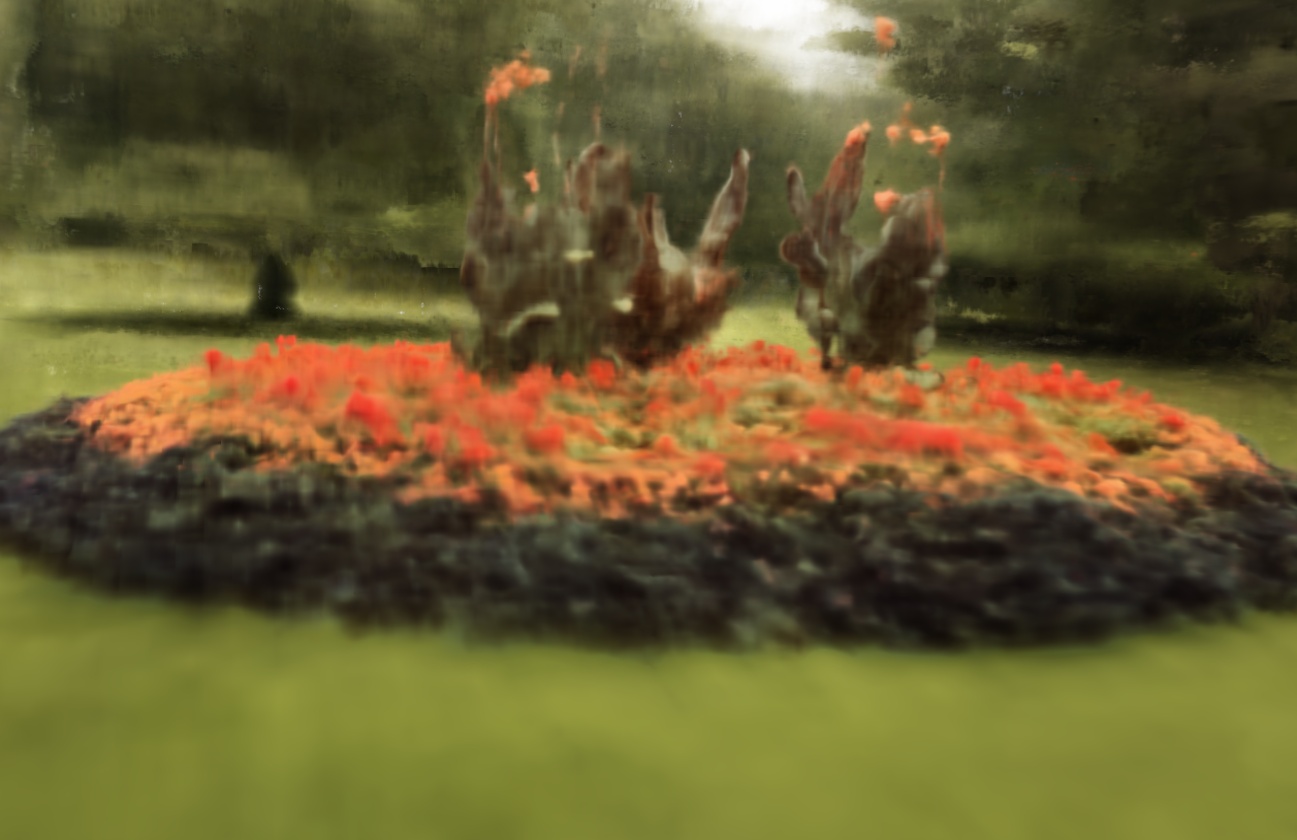} &
        \includegraphics[width=0.3\textwidth]{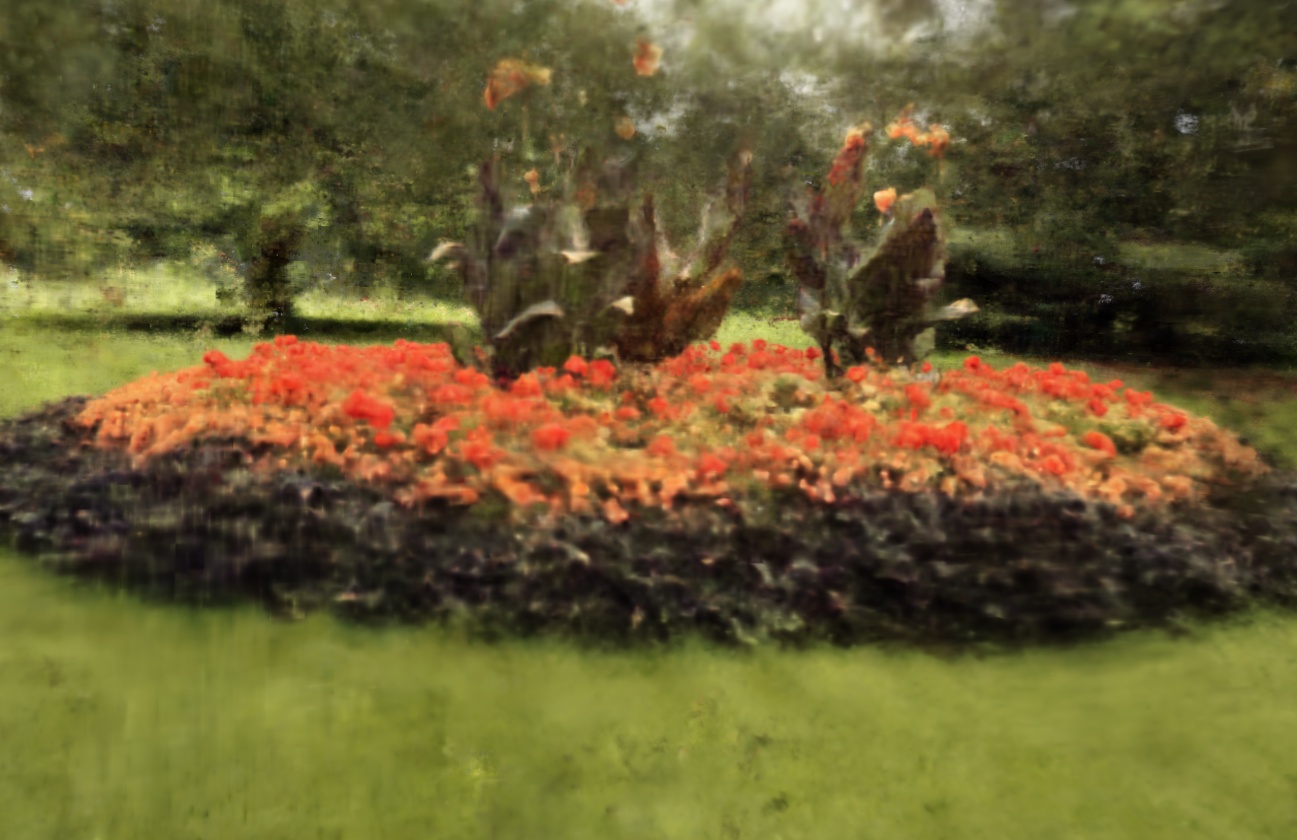} &
        \includegraphics[width=0.3\textwidth]{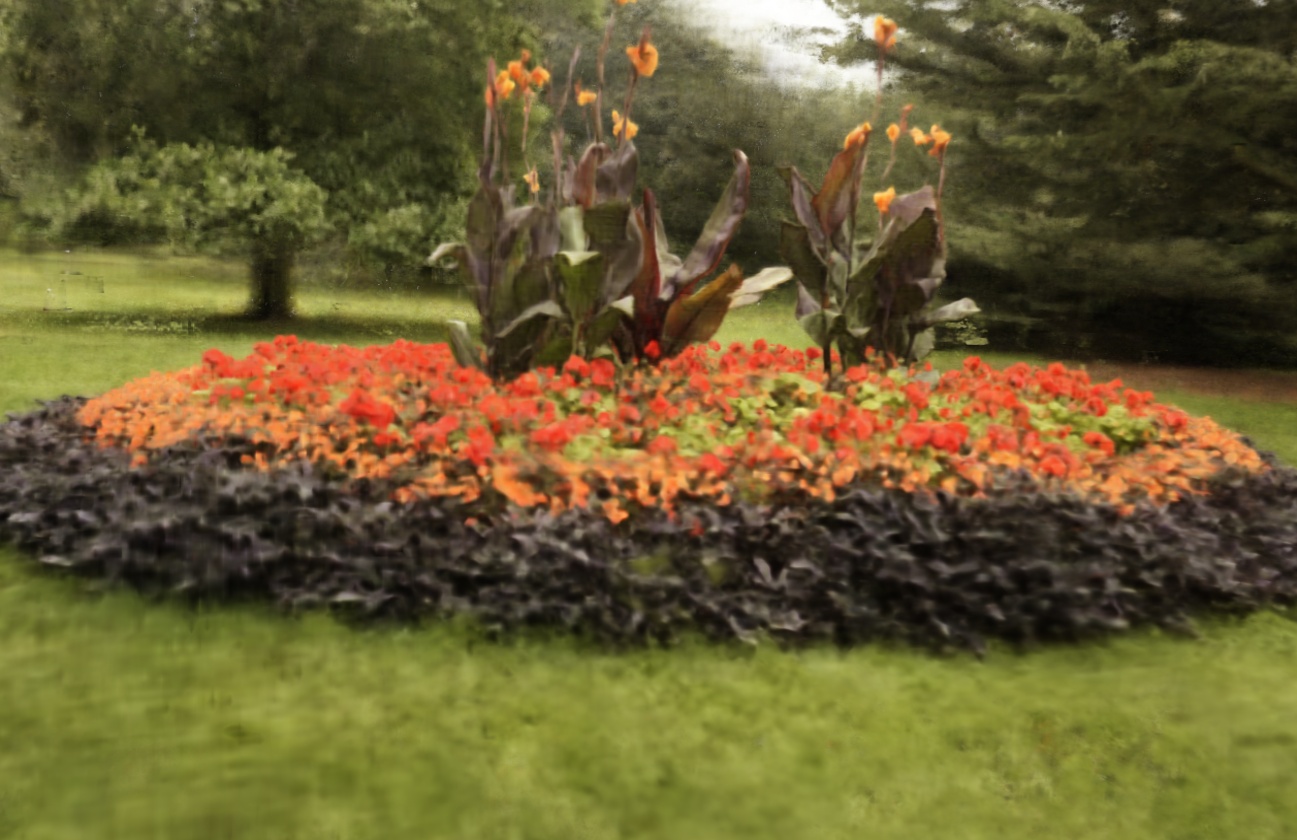}\\
         PPNG-1 & PPNG-2 & PPNG-3
    \end{tabular}
    }
    \caption{Qualitative results for unboudned 360$^\circ$ dataset.}
    \label{fig:appendix_360}
\end{figure*}

\end{document}